\documentclass[fleqn,final,5p,times,twocolumn,authoryear,10pt]{elsarticle}

\usepackage{hyperref}

\makeatletter
\providecommand{\doi}[1]{%
  \begingroup
    \let\bibinfo\@secondoftwo
    \urlstyle{rm}%
    \href{http://dx.doi.org/#1}{%
      doi:\discretionary{}{}{}%
      \nolinkurl{#1}%
    }%
  \endgroup
}
\makeatother

\usepackage{amssymb}
\usepackage{xspace}
\usepackage{lineno}
\usepackage[table]{xcolor}

\usepackage{tabularx}
\usepackage{booktabs}
\usepackage{nicematrix}
\usepackage{adjustbox}

\usepackage{array}
\newcolumntype{$}{>{\global\let\currentrowstyle\relax}}
\newcolumntype{^}{>{\currentrowstyle}}
\newcommand{\pixprob}{\ensuremath{P(\text{plume}|\text{pixel}) \in [0, 1]}\xspace}
\newcommand{\methane}{\ensuremath{\text{CH}_{4}}\xspace}
\newcommand{\cotwo}{\ensuremath{\text{CO}_{2}}\xspace}
\newcommand{\notwo}{\ensuremath{\text{NO}_{2}}\xspace}
\newcommand{\ammonia}{\ensuremath{\text{NH}_{3}}\xspace}
\newcommand{\water}{\ensuremath{\text{H}_{2}\text{O}}\xspace}

\newcommand{\msq}{\ensuremath{\text{m}^2}\xspace}
\newcommand{\kmsq}{\ensuremath{\text{km}^2}\xspace}
\newcommand{\gpmsq}{\ensuremath{\text{g/m}^2}\xspace}
\newcommand{\ppmm}{\ensuremath{\text{ppm-m}}\xspace}

\newcommand{\GHG}{\xspace{GHG}\xspace}

\newcommand{\ANG}{\xspace{AVIRIS-NG}\xspace}
\newcommand{\GAO}{\xspace{GAO}\xspace}
\newcommand{\EMIT}{\xspace{EMIT}\xspace}

\newcommand{\FOURC}{\xspace{Four} {Corners}\xspace}
\newcommand{\CALMETHANE}{\xspace{Cal}{\methane}\xspace}

\newcommand{\parsub}[1]{\noindent \textbf{#1}}
\newcommand{\itemsub}[1]{\item \textbf{#1}}

\newcommand{\first}{\ensuremath{1^\text{st}}}
\newcommand{\second}{\ensuremath{2^\text{nd}}}
\newcommand{\third}{\ensuremath{3^\text{rd}}}
\newcommand{\nth}[1]{\ensuremath{{#1}^\text{th}}}

\newcommand{\jake}[1]{\textcolor{black}{#1}}

\usepackage{xr-hyper}
\usepackage{hyperref}

\journal{Remote Sensing of Environment}

\begin{document}

\begin{frontmatter}

\title{Towards Operational Automated Greenhouse Gas Plume Detection \jake{and Delineation}}

\author[jpl]{Brian~D~Bue}
\author[jpl]{Jake~H~Lee}
\author[jpl]{Andrew~K~Thorpe}
\author[jpl]{Philip~G~Brodrick}
\author[cm]{Daniel~Cusworth}
\author[cm]{Alana~Ayasse}
\author[jpl,pri]{Vassiliki~Mancoridis}
\author[jpl,cit]{Anagha~Satish}
\author[jpl,col]{Shujun~Xiong}
\author[cm]{Riley Duren}

\affiliation[jpl]{organization={Jet~Propulsion~Laboratory, California~Institute~of~Technology}, 
%            addressline={4800~Oak~Grove~Drive}, 
            city={Pasadena},
            state={CA},
            postcode={91101}, 
            country={USA}}

\affiliation[cm]{organization={Carbon~Mapper~Inc.},
%            addressline={680~E~Colorado~Blvd~Suite~180}, 
            city={Pasadena},
            state={CA},
            postcode={91101}, 
            country={USA}}

\affiliation[pri]{organization={Princeton University},
            city={Princeton},
            state={NJ},
            postcode={08544},
            country={USA}}
          
\affiliation[cit]{organization={California~Institute~of~Technology}, 
%            addressline={1200~E~California~Blvd}, 
            city={Pasadena},
            state={CA},
            postcode={91125}, 
            country={USA}}
          
\affiliation[col]{organization={Columbia~University},
%            addressline={116th~and~Broadway}, 
            city={New~York},
            state={NY},
            postcode={10027}, 
            country={USA}}

\begin{abstract}
Operational deployment of a fully automated \jake{facility-scale} greenhouse gas (GHG) plume detection system remains \jake{challenging for fine spatial resolution imaging spectrometers,} despite recent advances in deep learning approaches. With the dramatic increase in data availability, however, automation continues to increase in importance for emissions monitoring. This work reviews and addresses several key obstacles in the field: data and label quality control, prevention of spatiotemporal biases, and correctly aligned modeling objectives. We demonstrate through rigorous experiments using multicampaign data from airborne and spaceborne instruments that convolutional neural networks (CNNs) are able to achieve operational detection performance when these obstacles are alleviated. We demonstrate that a multitask model that learns both instance detection and pixelwise segmentation simultaneously can successfully lead towards an operational pathway. We evaluate the model's plume detectability across emission source types and regions, identifying thresholds for operational deployment. Finally, we provide analysis-ready data, models, and source code for reproducibility, and work to define a set of best practices and validation standards to facilitate future contributions to the field.
\end{abstract}

\begin{keyword}
  Plume Detection
  \sep Greenhouse Gas
  \sep Machine Learning
  \sep Imaging Spectroscopy
  \sep AVIRIS
  \sep EMIT

\end{keyword}

\end{frontmatter}



\section{Introduction} \label{sec:introduction}

Modern airborne \& spaceborne imaging spectrometers such as the Airborne Visible-Infrared Imaging Spectrometer-Next Generation (\ANG) \citep{hamlin_imaging_2011,chapman_spectral_2019}, the Global Airborne Observatory (\GAO) \citep{asner_carnegie_2007,asner_carnegie_2012}, the Earth surface Mineral dust source InvesTigation (\EMIT) \citep{green_earth_2020}, and Carbon Mapper Coalition's Tanager-1 \citep{zandbergen2023preliminary} capture finely-sampled observations with spectral resolution sufficient to detect GreenHouse Gas (GHG) emissions with discriminative absorptions in the Visible to ShortWave InfraRed (VSWIR) and spatial resolution sufficient to \jake{attribute emissions to specific infrastructure elements.}

While current sensor platforms equipped with modern GHG retrieval approaches yield products that are valuable for plume identification and site-specific emissions quantification tasks, they are also prone to generating false enhancements. The presence of these false enhancements has to date prevented the operational deployment of fully automated plume detection and localization tasks.

Spatially-informed supervised machine learning models such as Convolutional Neural Networks (CNNs) provide a potential solution to automatically detect, localize, and delineate plumes from GHG retrievals. CNN-based approaches have proven successful in a variety of complex image classification, object detection, and pixelwise segmentation benchmark data sets \citep{russakovsky_imagenet_2014,pascal_voc_2010,krizhevsky_learning_2009}. Recent studies have claimed impressive results using CNN-based plume detection systems driven by GHG image products derived from data captured by airborne and spaceborne instruments \citep{joyce_using_2022,schuit_automated_2023,bruno_u-plume_2023, vaughan_ch4net_2023, kumar2023methanemapper, Rouet-Leduc2024}.

Advancing ML-driven plume detection capabilities to global scales demands informed optimization and validation strategies to ensure deployed models generalize operationally. Assessing the generalization capabilities of a supervised ML model involves distinguishing settings where we can trust the outputs produced by the model from those where we do not. While this is indeed a fundamentally difficult task, evaluating the capabilities of a proposed plume detector relative to its associated training and validation data with respect to the following criteria has proven instructive in our work:

\begin{itemize}\addtolength{\itemsep}{-0.5\baselineskip}
\item Do the validation procedures and performance metrics accurately and unambiguously capture the performance of the model on the available data? 
\item Are the available data sufficiently representative of the diversity of observations we expect to encounter when deployed operationally? 
\end{itemize}

We demonstrate in this work that effectively constructing and deploying an operational plume detector is an inherently interdisciplinary task, and that collaborative analysis performed by both ML practitioners and domain experts is crucial to operational plume detection capabilities. We show that systematic, informed analysis of model predictions, their associated GHG retrieval products and the settings in which they were captured can effectively reveal unexpected model or data-driven biases that impact generalization performance and methodological issues that corrupt training and validation procedures that produce misleading or uninterpretable results.

We address these issues in this work and in our previous efforts \jake{\citep{bue_opsch4_2023, bue_challenges_2024, lee_future_2021, lee_robust_2022, lee_robust_2023}} guided by the following research questions:

\begin{itemize}\addtolength{\itemsep}{-0.5\baselineskip}
\item What science objectives must an automated plume detection system address to be operationally effective? 

\item How should we frame plume detection as a machine learning problem to best progress towards achieving operational objectives? 

\item How do we rigorously and unambiguously characterize the capabilities of a plume detection system driven by machine learning?  
\end{itemize}

We aim to provide a basis towards advancing operational capabilities and identifying challenges involved in applying machine learning methods for automated plume detection. By collectively addressing these issues we hope to achieve more rigorous validation leading to an operational system for this critical climate problem.

This paper is organized as follows: Section \ref{sec:background} provides background information on the Columnwise Matched Filter (CMF) retrieval product, GHG plumes and their labels, and definitions of the plume detection task. Section \ref{sec:methods} describes the plume detection system, including quality control, sampling, and validation. Section \ref{sec:datamodels} describes this work's data sets and models, and Section \ref{sec:results} presents the corresponding plume detection results. Finally, Section \ref{sec:discussion} puts these results in context of limitations in existing literature, and we conclude in Section \ref{sec:conclusions}.

\section{Background} \label{sec:background}
To define operational requirements for a plume detection system, we first must understand the data and labels that will be used to construct and validate the system. Equipped with this knowledge, we can effectively construct protocols to collect representative samples to train and test the model, define optimization routines that account for task-specific sources of bias and uncertainty, and select capable validation metrics that provide realistic estimates of generalization performance. While the data and results described in this work are focused on matched filter-based retrievals of methane derived from imaging spectrometer observations, the challenges we describe are generally relevant to similar plume detection systems, including those based on image data captured by analogous remote sensing platforms such as PRISMA and EnMAP \citep{guanter_mapping_2021, roger2024enmap}, targeting other greenhouse gases such as \cotwo, \jake{\notwo \citep{borger2025high} or \ammonia \citep{balasus2026mapping}}, and retrievals derived using alternative algorithms such as mag1c \citep{foote_fast_2020}. 

\subsection{GHG Retrieval Image Products} \label{sub:retrievals}
We derive pixelwise GHG concentrations from calibrated imaging spectrometer radiance observations using the Columnwise Matched Filter (CMF) \citep{thompson_real_2015} retrieval approach. Based largely on the methods developed and refined by scientists at Los Alamos National Labs starting in the late 1990s \citep{villeneuve_improved_1999,theiler_incredible_2012}, the CMF is conceptually simple, computationally efficient, and produces retrievals that are both numerically stable across repeat observations of the same sites and emission sources and consistent with coincident measurements by complementary sensors (e.g., comparisons with \jake{Dynamic Aviation in situ observations} collected during Four Corners campaign \citep{thompson_real_2015}) and with measurements from controlled release experiments  \citep{thorpe_airborne_2017,thorpe_methane_2023,thorpe_mapping_2016,duren_californias_2019,el2024technological,sherwin_single-blind_2023,ayasse_performance_2023}. The CMF retrieval computes the pixelwise enhancement of VSWIR absorption features associated with a MODTRAN-simulated GHG transmittance spectrum in \ppmm units relative to multivariate Gaussian backgrounds, each estimated columnwise from the observations of the independent detector elements in the sensor's focal plane array. CMF \methane products derived from \ANG and \GAO data are highly analogous---both are based on the approach described in \cite{thompson_real_2015} that retrieves pixelwise \methane concentrations according to the depth of the SWIR2 \methane absorption feature of airborne spectrometer radiance observations captured at GSD $\in$ [1,10] m. Any differences between the instruments are dwarfed by atmospheric, geometric, and radiative transfer assumptions within the retrieval. EMIT observations are observed at 60 m GSD and EMIT CMF products are derived using the same matched-filter-based algorithm used with the airborne data. The operational EMIT L2 \methane product uses all VSWIR wavelengths rather than just the SWIR2, is more stable due to its environment in orbit, and leverages a Look-Up Table (LUT) to select an appropriate \methane target spectrum for the retrieval that accounts for varying \water concentration levels and solar zenith angles \citep{foote_impact_2021,thorpe_attribution_2023}. Filtering of cloud, water, and thermal anomalies from the calculation of the covariance also avoids artifacts in the retrieval.

\subsection{GHG Plume Types} \label{sub:cmfplume}
\begin{figure*}
 \centering \includegraphics[width=0.8\linewidth,clip,keepaspectratio]{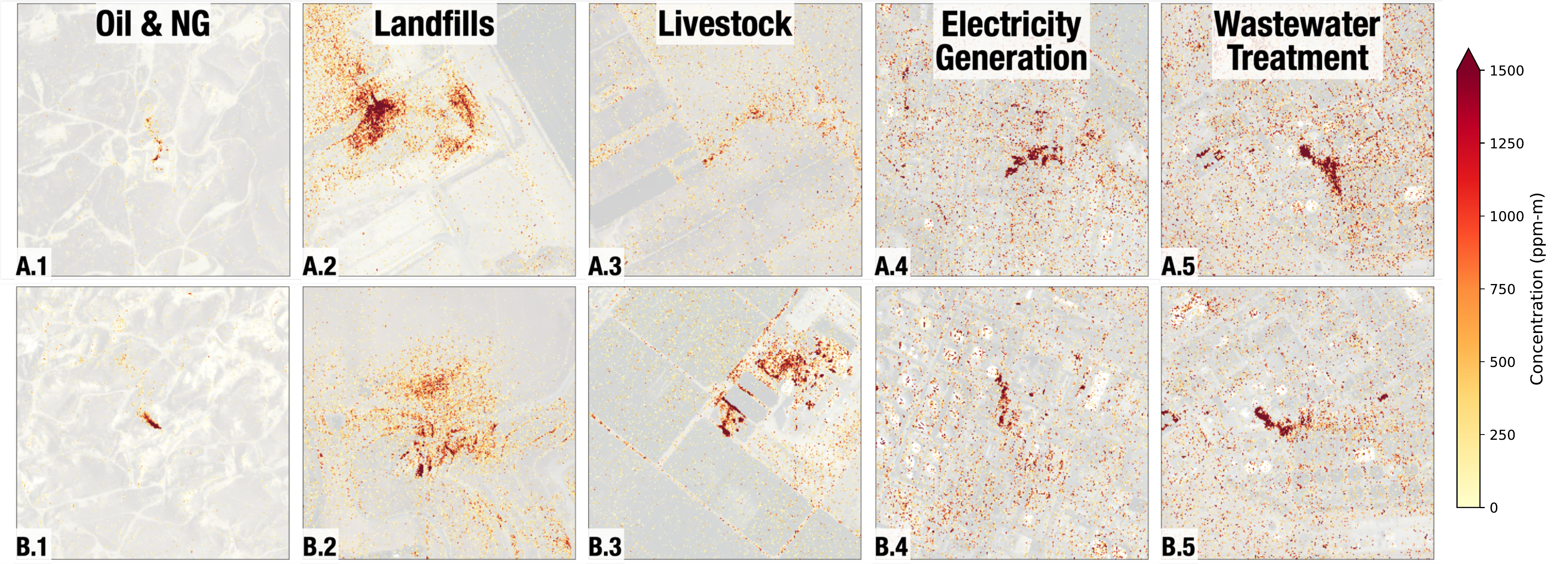} 
 \caption{Example \methane plumes with characteristic features of Oil \& Natural Gas (\first column) infrastructure, Landfills (\second column). Livestock/Manure Management (\third column), Electricity Generation (\nth{4} column) and Wastewater Treatment (last column) facilities. \methane concentration is overlaid as a yellow-red colormap over a desaturated RGB quicklook.}
 \label{fig:source_plumes}
\end{figure*}

GHG plumes manifest as spatially contiguous clusters of pixels distinguishable by higher concentration levels than their surrounding background pixels. Figure~\ref{fig:source_plumes} shows several example \methane plumes from distinct source types observed in airborne \methane imaging campaigns. The type of emission source and the emission rate, adjusted by dispersion from local winds, are the primary factors determining the concentration level and morphology of a given plume. Although distinct infrastructure types from various emission sources produce plumes with significant variability, for the purposes of detection we can broadly group source emitters into two categories based on the characteristics of the plumes they emit.

GHG plumes from oil \& natural gas infrastructure and industrial facilities tend to produce \textbf{``point source''} plumes with localized, concentrated emissions that can be traced back to a single specific location. The shape of point source plumes is largely governed by local wind conditions. In contrast, \textbf{``diffuse sources''} from agricultural infrastructure (e.g., manure, enteric fermentation, rice cultivation) and solid waste management facilities (e.g., landfills) produce plumes representing the collective emissions of multiple sources in close proximity of one another, or plumes emitted by area sources lacking distinct points of origin. 

\subsection{Plume Labels} \label{sub:plumelabels}
This work leverages \methane plumes identified by domain experts as described in Suppl. Sections~\ref{apx:qc} and \ref{apx:airborne}. Constructing and validating a ML-driven GHG plume detector also requires pixelwise labels \jake{(binary masks)} to teach the model to distinguish plume pixels from background pixels. However, plumes are not discrete objects with well-defined boundaries, and manually labeling individual plume pixels is a difficult and subjective task, particularly near plume boundaries. One alternative to requiring users to manually label individual plume pixels is to use the CMF enhancements to deterministically generate ``CMF-guided'' plume labels consisting of highly concentrated pixels near the origin point of each user identified plume. While the CMF-guided labels may be less subjective than manually defined labels, their quality can vary significantly by scene according to the local contrast in \ppmm values between plumes and background enhancements or the presence of false enhancements adjacent to real plumes, and consequently require comprehensive manual inspection to refine or reject uninformative instances.

\begin{figure*}
 \centering
 \includegraphics[width=0.8\linewidth,clip,keepaspectratio]{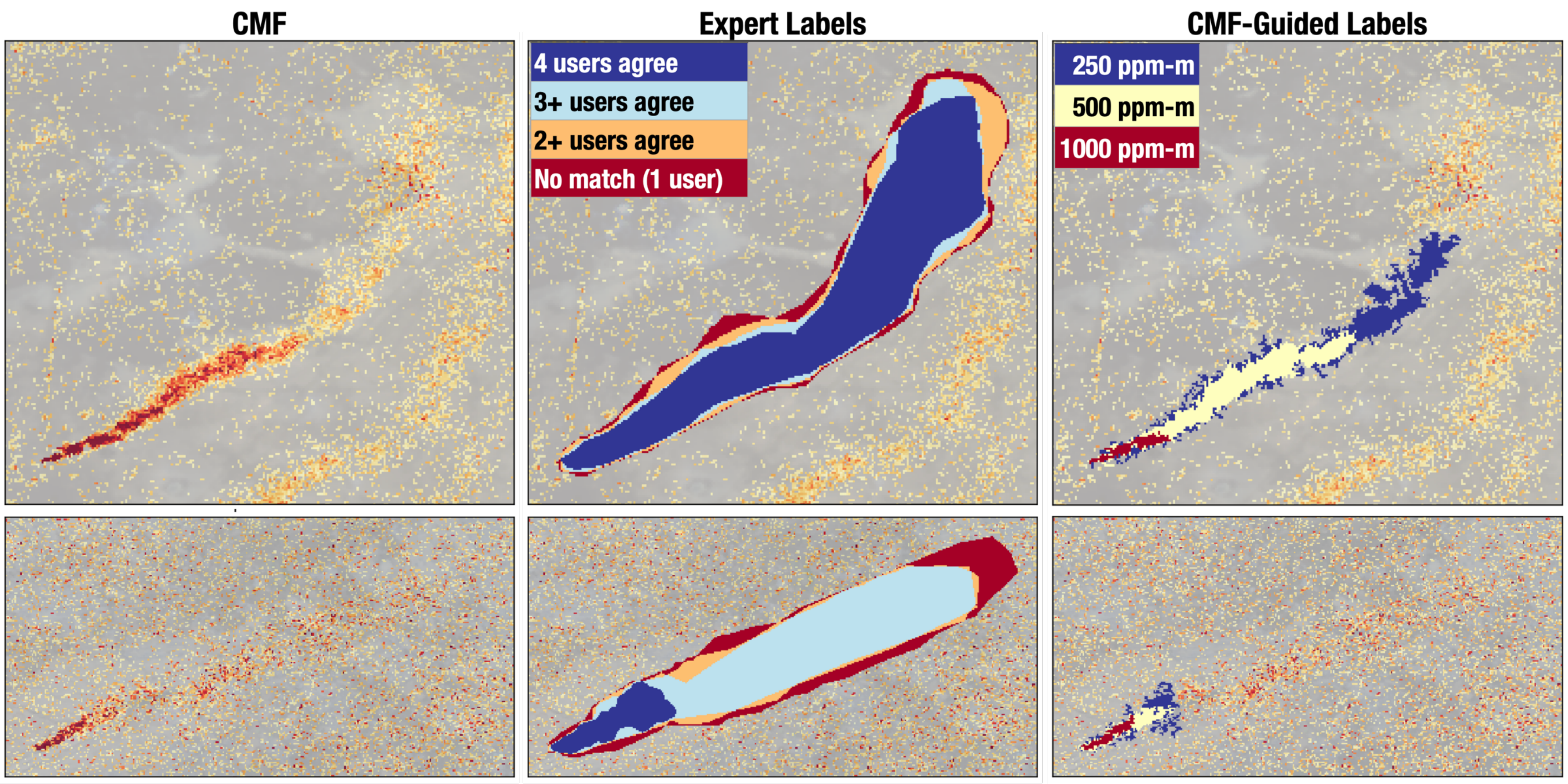} 
 \caption{Left column: example \methane plumes observed in CMF retrieval images. Center column: Pixelwise agreement between user-provided plume label masks provided by four independent experts. Right column: CMF-guided plume labels with thresholds $\in$ [250,500,1000]\ppmm. Labels were generated using the Multi-Mission Geographic Information System \citep{soliman_2025}.}
 \label{fig:userweaklab}
\end{figure*}

Figure~\ref{fig:userweaklab} shows two example plume candidates (left column) with corresponding labels provided by four independent experts (center column), along with the deterministic CMF-guided plume labels for each plume according to increasing \ppmm thresholds (right column). \jake{The experts' labels are smoother and coarser compared to the CMF-guided plume labels, as they are defining polygons by clicking on vertices. Experts were also instructed to include a small buffer region beyond the diffused edges of the plume to capture the entire extent.} All four users agree that each enhancement represents a real \methane plume, but the pixelwise masks they each provide disagree with respect to the boundaries where plume pixels end and background pixels begin (center column). In contrast, the CMF-guided labels are more conservative, capturing only the highly concentrated pixels near the origin point associated with each plume. \jake{The CMF-guided labeling process is described in detail in Suppl. Section \ref{sub:cmflab}.}

\subsection{Background and False Enhancements} \label{sub:cmfbg}
Even in targeted GHG monitoring campaigns involving repeated measurements of known emission sources, pixels representing GHG plumes are extremely rare. For instance, labeled \methane plume pixels account for less than 1\% of the science pixels captured in the CMFs from the 2018-2020 airborne \ANG and \GAO \methane imaging campaigns. The remaining pixels represent enhancements where the target \GHG is absent or concentrated at a level too low to be detectable by the retrieval algorithm or measurable by the sensor. These enhancements generally fall into one of two categories: {\em Background Enhancements} and {\em False Enhancements}, typified by the spatial distribution of their enhancements, how often they occur in each scene, and the factors (real or artificial) that produce them. 

\parsub{Background Enhancements}: The majority of pixels in a typical scene are common background enhancements representing retrievals of surface materials lacking characteristic GHG absorption features varying with scenewise imaging geometry, albedo, illumination conditions, and atmospheric state. Most background enhancements exhibit small values that fall within the range where GHG concentrations are too low to distinguish them from instrument noise (e.g., $<$ 250 \ppmm in typical airborne \methane CMFs). By definition, the CMF column concentration should follow a zero-mean distribution, with the diffuse background emerging as a consequence of instrument noise characteristics and environmental conditions. Though higher concentration retrievals occasionally occur in regions dominated by common background enhancements, they tend to be spatially sparse. Consequently, regions consisting primarily of common background enhancements consist of mostly low concentration pixels with sporadic high enhancement pixels that lack cohesive spatial structure, making such regions easily distinguishable from real GHG plumes. 

\parsub{False Enhancements}: False enhancements are a nuisance class of background enhancements with concentration levels within the range of real GHG plumes (exceeding 500-1000\ppmm) caused by factors unrelated to the presence of GHG emissions. One common driver involves \textit{confusers}, including surface spectroscopy that mimics methane absorption features at the spectral sampling interval of the instrument and anomalous observations, such as high albedo materials and thermal anomalies. Another involves \textit{artifacts} caused by limitations of the instrument or retrieval method, such as corrupt background covariance estimates due to flares and sample size limitations from short flightlines. Although false enhancements are relatively rare in terms of pixelwise coverage (we estimate false enhancements represent 1-5\% of the retrieved pixels in a typical CMF scene) false enhancements that manifest at spatial scales and with spatial structure similar to real plumes constitute the majority of false positive plume detections observed in this work, and can be difficult even for domain experts to distinguish from real GHG plumes. False enhancements can severely restrict plume detectability when they occur near real emission sources and merge with pixels representing observed plumes, thereby altering their spatial distribution and concentration levels. Additionally, in some rare, pathological cases, false enhancements can cover the spatial extent of an entire scene (see examples in Suppl. Section~\ref{apx:gao_penn} and Suppl. Figure~\ref{fig:penn_tiles}).

It is worth noting that, in cases of low sensitivity observations, it is also possible to identify false negative enhancements. While less problematic and even rarer than the false positive enhancements discussed above, it can cause challenging plume morphologies; for example, a plume blowing over a river may appear to be cut in half where methane was not retrieved due to the low albedo.

\begin{figure*}
  \centering
  \includegraphics[width=1.0\linewidth,clip,keepaspectratio]{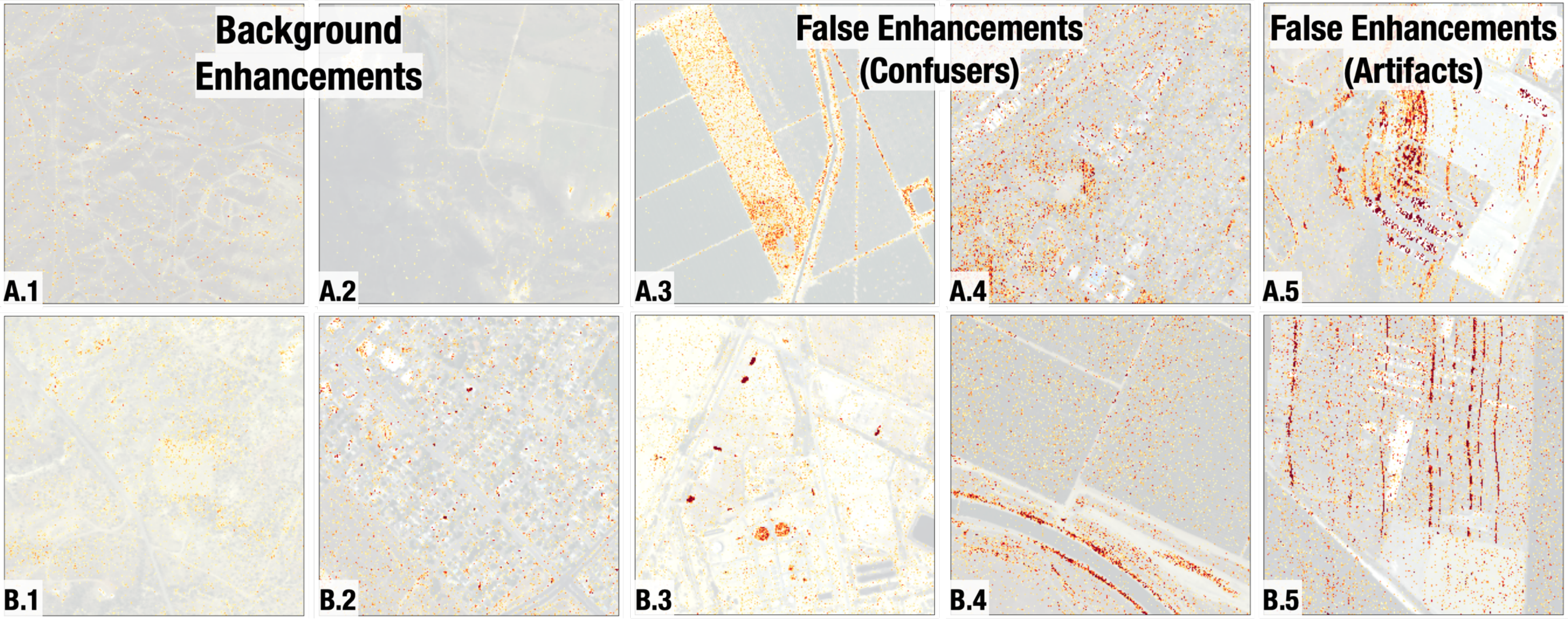} 
 \caption{Example common background (\first\ two columns) and false (last three columns) \methane enhancements. Common background enhancements lack cohesive spatial structure, making them easy to distinguish from plumes.``False'' enhancements from confuser materials and retrieval artifacts exhibit concentration levels in the range of real plumes, and can be difficult to distinguish from real plumes.}
 \label{fig:bge_false}
\end{figure*}

Figure~\ref{fig:bge_false} shows example background and false enhancements. Common background enhancements (columns 1 and 2) consist of low concentration pixels and sporadic, sparsely distributed high enhancement pixels. False enhancements from confuser materials or albedo variation (columns 3-4) are typically correlated with visible surface features, while false enhancements from artifacts (column 5) tend to be uncorrelated with surface features, typically following downtrack imaging geometry (e.g., columnwise retrieval artifacts) or with both downtrack \& crosstrack imaging geometry simultaneously (e.g., Fourier ringing from thermal saturation in SWIR channels caused by oil \& natural gas flares \citep{foote2018artifact}).

\subsection{Operational Plume Detection} \label{sub:science_objectives}

An automated plume detection system is {\em operationally effective} for image products measuring emissions of a target GHG (e.g., CMF images of \methane or \cotwo) observed by a target remote sensing platform (e.g., AVIRIS-NG, EMIT) if, when applied to arbitrary target images observed in nominal imaging conditions, the plume detection system performs comparably to domain experts with respect to the following objectives:

\begin{enumerate}
\itemsub{Instance Detection:} Detect and localize identifiable target plumes and reject identifiable false enhancements.
\itemsub{Pixelwise Segmentation:} Delineate pixelwise boundaries of detected plumes.
\end{enumerate}

Both of these objectives are achievable using current GHG image retrieval approaches applied to data captured by modern airborne and spaceborne imaging spectrometer platforms. \jake{Prior manual plume identification efforts have shown that human experts can consistently identify, localize, and delineate most \methane plumes above the minimum detection limit of EMIT through visual inspection of the CMF retrieval images we consider in this work \citep{duren_californias_2019, thorpe_attribution_2023}}.

Comprehensive manual analysis of retrieval images captured in airborne imaging campaigns has also informed improvements to GHG retrieval algorithms, suppressing common classes of false enhancements, and providing diagnostic capabilities to quantify retrieval sensitivity and uncertainty.

Requiring a plume detector to perform comparably to human experts makes for a challenging machine learning problem. However, it also provides constraints that simplify validation efforts. Specifically, we do {\em not} include scenes in model training where domain experts are unable to exhaustively identify all candidate plume enhancements in the scene with adequate confidence. We also require all plume candidates be identifiable using {\em only} the GHG retrieval images we use to train and validate our models. Ambiguous plume candidates that can only be resolved by cross-referencing external data sets (e.g., high spatial resolution basemap images) or prior measurements of the same site are excluded from our experiments. Evaluating pixelwise plume delineation accuracy when human experts disagree regarding which pixels distinguish plumes from background pixels is difficult to interpret, so we avoid it. Details regarding the quality control criteria we use to exclude low SnR scenes and unidentifiable plumes are provided in Section~\ref{sec:data}, and Supplemental Section~\ref{sub:angqc}.

While a fully-operational system must be able to flag data quality issues such as low SnR scenes or ambiguous detections for manual review or reprocessing, relying on the plume detector for such tasks is often a waste of resources, as simple triage protocols to account for and handle problematic scenes could be more effective at identifying known issues. We provide a summary of several data triage routines in Supplemental Section~\ref{sub:angqc}.

\section{Methods} \label{sec:methods}

\begin{figure*}
 \centering
 \includegraphics[width=1.0\linewidth,clip,keepaspectratio]{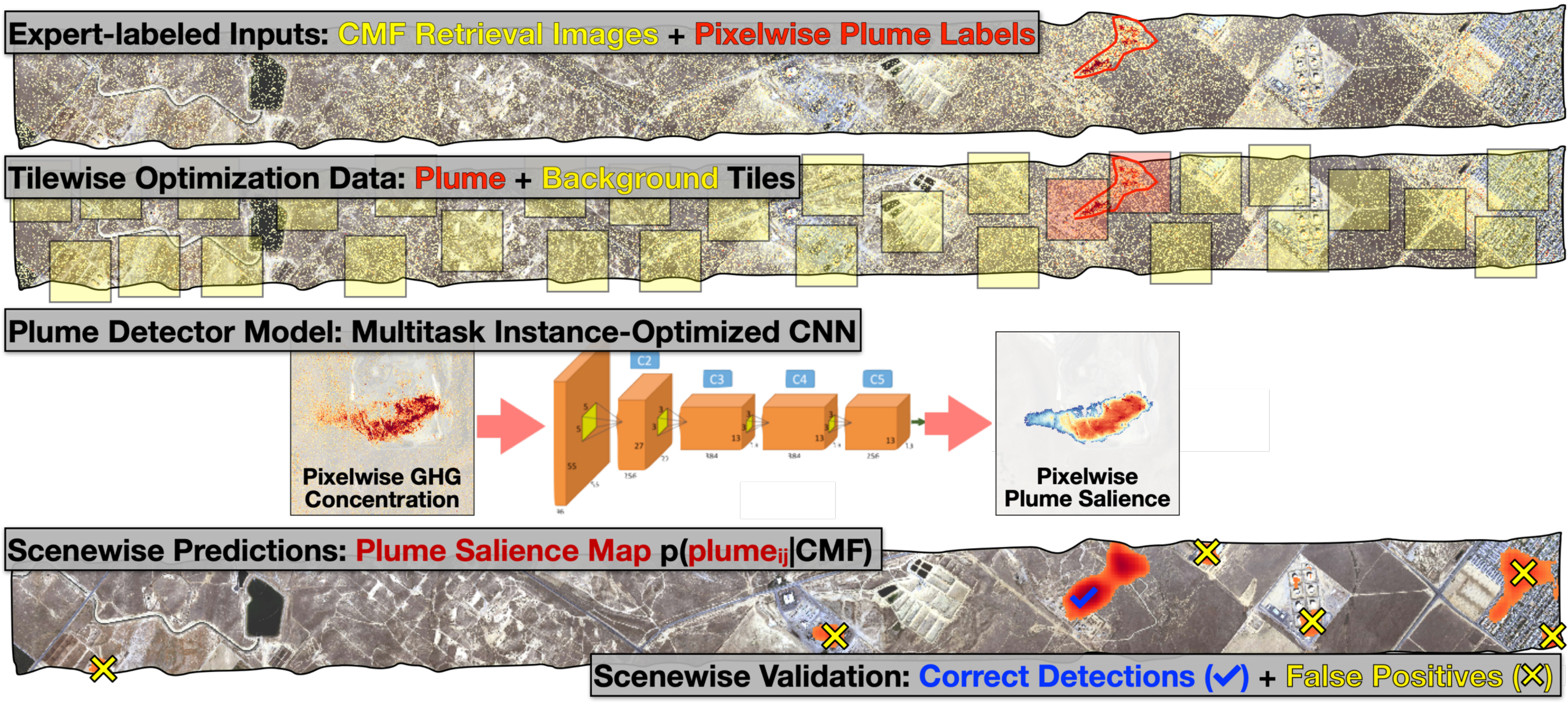} 
 \caption{Components of an operational CNN-based plume detection system driven by GHG retrieval images. {\bf A:} We are provided a set of GHG retrieval image scenes with pixelwise labels indicating the boundaries of plumes within each scene. {\bf B:} We employ a spatially stratified sampling and validation strategy to extract tiles from our labeled GHG scenes and partition the extracted tiles by scene into spatially disjoint training and test sets. {\bf C:} Provided our training/test tiles, our objective is to construct a plume detector that predicts the probability that each pixel in an input GHG image represents a plume. {\bf D:} Once trained, an operational model will be deployed to detect plumes in full-scene GHG retrieval image products, and we validate candidate plume detectors with respect to {\em scenewise} plume instance detection and segmentation metrics accordingly.}
 \label{fig:system}
\end{figure*}

\parsub{System Overview}: We consider plume detection systems of the form shown in Figure~\ref{fig:system} in this work. Provided a set of labeled scenes consisting of GHG retrieval image products and their associated pixelwise plume labels, our objective is to train a plume detector model to estimate the probability (or ``plume salience'') \jake{\pixprob} that each pixel in the image represents a plume. Supervised machine learning models achieve this by learning to map labeled plume image pixels to high salience values and unlabeled ``background'' image pixels to low salience values. In this work, we focus on CNN-based plume detectors which learn a hierarchy of spatial filters that capture \jake{discriminative morphological} features of labeled plumes and unlabeled background enhancements using fixed-size image tiles extracted from our labeled CMF scenes. Once trained, CNNs are capable of efficiently generating spatially-informed pixelwise predictions (or ``plume salience maps'') based on CMF images with arbitrary input dimensions. 

\parsub{Quality Control}: We rely on comprehensive quality control procedures (see Suppl. Sections~\ref{sub:angqc} and \ref{sub:emitqc}) to ensure that our models are trained and validated on high quality, exhaustively labeled CMF image products whose constituent plume labels are readily identifiable and distinguishable from (unlabeled) background enhancements by domain experts without additional context.

\parsub{Tilewise Training \& Scenewise Validation}: Once deployed, we assume an operational plume detector will be used to detect and delineate plumes in ``full scene'' CMF image products, and the performance of candidate plume detector models should be validated on full scene CMF products to replicate operational conditions. While the CNN models described in this work are functionally capable of generating full scene predictions, scenewise validation of plume predictions is not standard practice. This is due to the fact that CNNs are trained ``tilewise''  using collections of sub-image tiles with fixed dimensions extracted from full CMF scenes. Once trained, such models are typically also evaluated ``tilewise'' by applying the model to test tiles with the same image dimensions.     

We apply our tilewise-trained models to generate salience maps of the subset of our test scenes that meet strict ``scenewise'' quality-control requirements (see Suppl. Section~\ref{sub:angqc} for details regarding quality control criteria). For these scenes, we measure prediction accuracy on all valid pixels in each scene with respect to pixelwise plume segmentation and plume instance detection (see Suppl. Section~\ref{sub:instance} for details regarding measuring plume instance detection accuracy).

To ensure our models and results are not spatiotemporally biased, we use the bounding boxes of the georeferenced CMF images to stratify our data into spatially disjoint training and test scenes. For airborne campaigns, we compute the graph of pairwise Intersection-over-Union (IoU) scores with respect to the scenes observed in each campaign. Each connected component of the resulting IoU graph identifies a group of overlapping flightlines that are spatially disjoint from all the remaining flightlines. For the spaceborne EMIT data, we group scenes according to their respective \jake{Universal Transverse Mercator (UTM)} zone IDs. We generate spatially disjoint, class stratified training and test partitions by assigning the labeled plume and background tiles associated with each group of flightlines to {\em either} the training set or the test set via the GroupShuffleSplit function provided in the scikit-learn package \citep{scikit-learn}. For each campaign, we designate roughly 75\% of the flightlines (and their associated tiles) as training data, and the remaining 25\% of the flightlines as test data.

\parsub{Spatially-stratified Tile Sampling}: Similar to the scene-wise stratification, we need to ensure that the sampled distribution of CMF tiles is well distributed and non-overlapping within a given scene. We use a spatially-stratified tile sampling strategy that  prioritizes sampling background tiles in regions with high concentration levels, and incorporates controls that explicitly prevent inter-class tile overlap (i.e., plume and background tiles may never overlap) while permitting nominal within class overlap (i.e., background tiles may overlap other background tiles). Our procedure begins by extracting candidate plume tiles at the center of mass of each labeled plume ROI. We then exclude any pixelwise plume labels that lie within the current plume tile boundaries, which partitions any large plume ROIs into a new set of sub-plume ROIs. We repeat this process until we cannot sample any more spatially disjoint plume tiles centered on distinct sub-plume ROIs. After sampling all available plume tiles, we use the CMF-guided background labels to randomly sample background tiles that roughly cover the remainder of the scene. We ensure that background tiles do not overlap any existing plume tiles or any pixelwise plume labels outside of the selected plume tiles. However, we do allow background tiles to overlap other background tiles in the scene by at most 10\%, and we also allow background tiles to overlap nodata regions (most commonly, scene boundaries) by at most 50\%. We found that allowing some amount of background-background overlap allows for a more comprehensive sampling of the background enhancements within each scene and exposes the model to observations near scene boundaries that are inevitably present in full CMF images. 

\parsub{Evaluating Plume Detection and Segmentation Metrics}
In addition to tilewise performance metrics, we report plume detection and segmentation metrics calculated over entire scenes/flightlines. Pixelwise segmentation metrics measure how closely pixelwise predictions align with labeled plume boundaries, while instance-level detection metrics measure how many plume instances were successfully detected and how many false plume detections occurred. High pixelwise accuracy does not imply high instance detection accuracy (and vice-versa), so both instance-level plume detection and pixelwise segmentation metrics are essential to unambiguously assess plume detector performance. Pixelwise segmentation metrics are reported simply as precision, recall, and F1-Scores over all pixels, while instance-level detections are reported based on whether a detection mask overlaps a label mask; this is defined in detail in Suppl. Section~\ref{sub:instance}.

\section{Data \& Models} \label{sec:datamodels}

\subsection{Data} \label{sec:data}

\begin{table*}
  \centering
  \includegraphics[width=1.0\linewidth,clip,keepaspectratio]{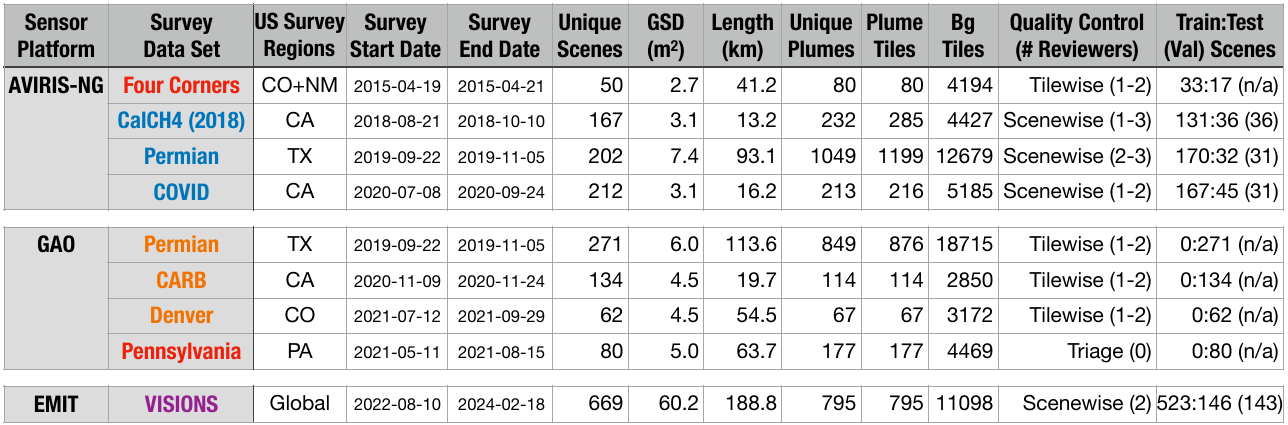}
  \caption{AVIRIS-NG, GAO and EMIT methane plume data sets considered in this work. Data sets IDs used to train/test the airborne (AVIRIS-NG/GAO) methane plume detector model are indicated by blue text, and purple text indicates the data used to train/test the spaceborne (EMIT) methane plume detector. Airborne campaigns used only as hold-out test data for the airborne plume detector are shown in orange text. Red text indicates the data sets with unique characteristics described in detail in Suppl. Sections~\ref{apx:bias_region} and \ref{apx:gao_penn}. These were also only used as hold-out test data.}
  \label{tab:datatab}
\end{table*}

Table~\ref{tab:datatab} provides a summary of all of the \methane CMF data sets we consider in this work. We summarize relevant details regarding the data sets used for training and validating our airborne (\ANG and \GAO) and spaceborne (EMIT) plume detectors here. Detailed column definitions for Table~\ref{tab:datatab} are provided in Section~\ref{apx:datacols}, and quality control levels are described in Suppl. Section~\ref{apx:qc}. The distribution of source sectors for plumes observed in airborne imaging campaigns are provided in Section~\ref{apx:campaign_sectors}.

\subsection{Airborne Plume Detector Data} \label{sub:airdata}

\parsub{\ANG Multicampaign \methane Data Set}: Except when otherwise specified, all plume detection experiments described in this work involving airborne imaging campaigns are based on  the curated, quality controlled set of labeled CMF images derived from data captured by the AVIRIS-NG airborne imaging spectrometer in the 2018 CalCH4 campaign \cite{duren_californias_2019}, 2019 Permian Campaign \cite{cusworth_intermittency_2021}, and 2020 COVID \cite{thorpe_methane_2023} \methane imaging campaigns. 

\parsub{GAO Multicampaign \methane Data Set}: We evaluate the generalization capabilities of the airborne plume detectors on test data consisting of labeled CMF images derived from data captured by the GAO airborne imaging spectrometer in the 2019 Permian, 2020 CARB, and 2021 Denver \methane imaging campaigns \citep{cusworth2022strong}. 

\parsub{Unique Campaigns}: CMF image products derived from data captured in the 2015 \ANG Four Corners campaign and the 2021 GAO \jake{Pennsylvania (Penn)} campaign exhibit characteristics that are distinct from CMF images derived from data captured in other airborne campaigns. We summarize how these distinctions impact plume detection capabilities and the interpretation of plume detection result in Suppl. Section~\ref{apx:bias_region}, but we exclude the observations from these anomalous campaigns from all of our remaining experiments. 

\subsection{EMIT Plume Detector Data} \label{sub:spacedata}

\parsub{EMIT VISIONS \methane Data Set}: All plume detection experiments involving EMIT observations were trained and tested on observations from the EMIT VISIONS \methane Data Set. The EMIT VISIONS data set consists of the expert curated, high SnR \methane CMF image products derived from data captured by the spaceborne imaging spectrometer EMIT between 2022-2024 that contain high-confidence methane plumes delivered to the public-facing EMIT VISIONS portal \footnote{\url{https://earth.jpl.nasa.gov/emit/data/data-portal/Greenhouse-Gases/}} that were formally-reviewed \& approved by multiple reviewers using the Multi-Mission Geographic Information System \citep{soliman_2025}. \jake{For experiments with EMIT data, we use the readily available expert plume extent delineation products as segmentation mask labels instead of CMF-guided labels.}

\subsection{Models} \label{sec:models}

\parsub{CNN Architectures}

Our experiments leverage two widely-used and well-established CNN architectures: GoogLeNet and U-Net. Model architecture, training procedures, and loss functions referenced are described in full in Suppl. Section~\ref{apx:ml}.

\begin{itemize}

\itemsub{GoogLeNet}: The GoogLeNet architecture \citep{szegedy_going_2015} incorporates several ``inception blocks'' designed to learn convolutional filters capturing discriminative image patterns at multiple spatial scales. Designed for image classification tasks, we train GoogLeNet using tiles extracted from CMF image scenes with tilewise plume or background labels determined by the presence of plume pixels in each tile. We compare the capabilities of GoogLeNet-based plume image classifier models to those of the recently proposed MethaNet \citep{jongaramrungruang_methanet_2022} plume image classifier model on a variety of plume image classification experiments (see Sections~\ref{apx:bias_region} and \ref{apx:bias_distrib}). To perform pixelwise plume segmentation, we follow the procedure described by \cite{long_fully_2015} to convert our trained GoogLeNet image classifier to a \jake{Fully Convolutional Network (FCN) by replacing the fully connected layers with $1 \times 1$ convolutional filters that retain the learned parameters. To obtain a dense, high-resolution output, we then employ the ``shift and stitch'' procedure to account for the spatial downsampling inherent in the pooling layers. }This allows us to efficiently apply GoogLeNet image classifiers to full-scene, pixelwise plume segmentation and instance detection tasks. We also apply model anti-aliasing as described in \cite{zhang2019makinginvar} to avoid shift invariance artifacts in produced masks \citep{lee2020shiftinvar}.

\itemsub{U-Net}: The U-net architecture \cite{ronneberger_u-net_2015} is among the most widely-used CNN architectures to date. Originally designed for medical applications involving pixelwise "semantic segmentation" tasks \cite{siddique_u-net_2021}, U-net is the base plume detector model used in several recent papers \citep{bruno_u-plume_2023, ruzicka_semantic_2023}. The U-net architecture learns detailed image patterns directly from pixelwise labeled training images by connecting each convolution/downsampling layer to a matching deconvolution/upsampling layer. These `skip connections'' allow the model to learn discriminative image filters at multiple spatial scales (via convolution \& pooling) while simultaneously learning to generate accurate pixelwise predictions at native spatial resolution by combining the outputs of learned multiscale filters. 

\end{itemize}

\parsub{Scenewise Plume Detector Models}

For our experiments involving airborne \methane plume detection, we evaluated the capabilities of the following scenewise plume detector models with respect to plume instance detection and pixelwise segmentation objectives.

\begin{itemize}

\itemsub{Tilewise Model}: \jake{The GoogLeNet FCN model architecture with ``shift-and-stitch'' initially trained for the plume classification task using tilewise (rather than pixelwise) labels. Due to its optimization objective, it offers advantages for plume instance detection but is ill-suited for pixelwise segmentation as it produces relatively coarse and imprecise masks. Complete details about this model and its optimization objective is provided in Suppl. Section \ref{sub:tilewise}.}

\itemsub{Pixelwise Model}: \jake{The U-Net model architecture trained for the plume segmentation task optimized with purely pixelwise optimization objective. It generates detailed pixelwise saliency map predictions, but it is ill-suited for instance detection due to a large number of false positives. Complete details about this model and its optimization objective is provided in Suppl. Section \ref{sub:pixelwise}.}

\itemsub{Multitask Model}: \jake{The U-net model architecture trained for the plume segmentation task optimized with both segmentation and classification task objectives (hence, multitask). Provides a compromise with good instance detection performance (reduced false positives) and detailed saliency map predictions. Complete details about this model and its optimization objective is provided in Suppl. Section \ref{sub:multitask}.}

\end{itemize}

Based on our airborne plume detection results with the models defined above, we trained and validated a separate multitask spaceborne plume detector model using the EMIT visions dataset and provide a comparative analysis of our airborne versus spaceborne plume detection/segmentation results.

\section{Results} \label{sec:results}

\subsection{Quantitative Scenewise Detection and Segmentation Performance} \label{sub:scenewise_eval}

\begin{figure}
  \centering
  \includegraphics[width=1.0\linewidth,clip,keepaspectratio]{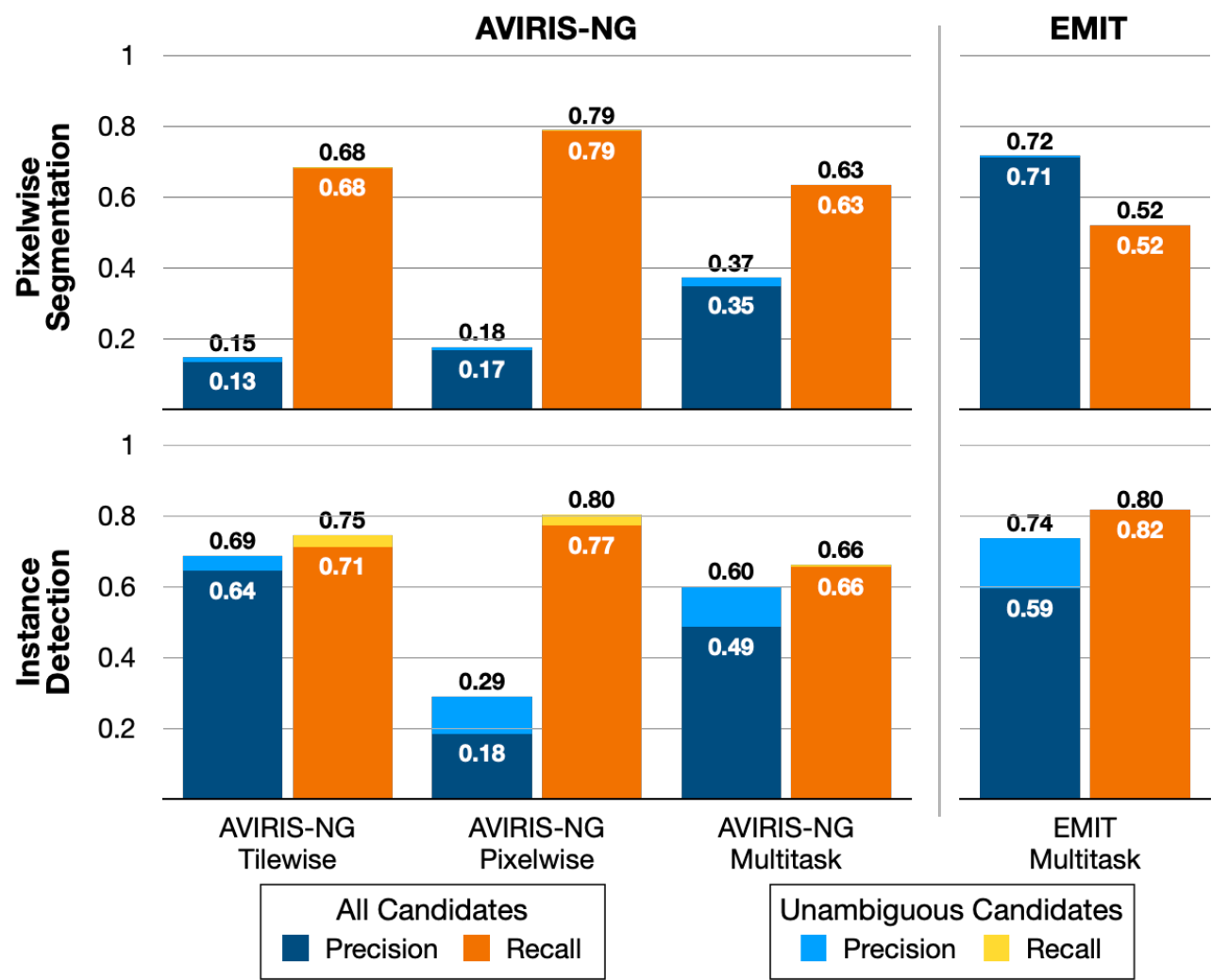} 
 \caption{Pixelwise segmentation (top) and Instance detection (bottom) precision and recall scores on \ANG validation scenes with respect to the \ANG Tilewise, Pixelwise and Multitask models (left), and EMIT validation scenes with respect to the EMIT Multitask model (right). Dark and light blue bars give the per-pixel scores on all detections versus unambiguous detections, respectively. Orange and yellow bars give the per-instance scores on all detections versus unambiguous detections, respectively.}
 \label{fig:scenewise_pre_rec}
\end{figure}

\jake{In this section, we describe the scenewise detection and segmentation performances of our airborne and spaceborne models in detail. As described in Section~\ref{sec:methods} and further in Suppl. Section~\ref{sub:instance}, we calculate our metrics on entire flightlines and scenes, not only the chipped tiles that the models were trained on. By evaluating both detection and segmentation metrics, we aim to characterize the models' performances in contexts relevant to operational use.}

\subsubsection{Tradeoffs Between AVIRIS-NG Models}

\jake{Figure~\ref{fig:scenewise_pre_rec} compares the pixelwise segmentation and instance detection performances of the three \ANG models, and Figure~\ref{fig:salcmp} visualizes examples of detected plumes and their segmentations. The tilewise model, only optimized for classification, yields the highest instance detection performance (0.64 precision and 0.71 recall). However, partially because it was not trained on any label masks, it generates relatively coarse masks, as shown in the second column of Figure~\ref{fig:salcmp} (0.13 precision and 0.68 recall).} Consequently, excluding ambiguous small/weak detection candidates (indicated in light blue and yellow) produces the smallest increases in \jake{both instance-level (+0.05 precision and +0.04 recall) and pixelwise accuracy (+0.02 precision and +0 recall)} of the three models.

The pixelwise-trained model gives a slight improvement in pixelwise accuracy \jake{(0.17 precision and 0.79 recall)}, but produces the worst overall instance detection accuracy due to the large quantity of small false positive detections it generates \jake{(0.18 precision and 0.77 recall)}. Consequently, excluding ambiguous small/weak candidates produced by the pixelwise model provides the most substantial increase in detection accuracy of the three models \jake{by filtering out spurious false positive detections (+0.11 precision and +0.03 recall)}. \jake{However, it is not enough to sufficiently improve the instance detection performance for operational use.}

The multitask model\jake{, optimized for both classification and segmentation, }is less prone to overfitting to ambiguous/uncertain plume boundary pixels in the training data, which accounts for its increase in pixelwise accuracy \jake{(0.35 precision, 0.63 recall)} over the pixelwise-trained model. It also performs competitively with the tilewise model in terms of instance detection \jake{(0.49 precision, 0.66 recall)}, particularly when we exclude small/weak ambiguous detections \jake{(+0.11 precision, +0 recall)}. \jake{By accounting for both detection and segmentation in its optimization, the multitask model provides the best tradeoff between detecting the presence of plumes and producing detailed masks.}

\subsubsection{Comparing \ANG and EMIT Results}

\jake{Figure~\ref{fig:scenewise_pre_rec} also compares the pixelwise segmentation and instance detection performances of the multitask EMIT model.} With respect to pixelwise segmentation performance, \ANG pixelwise mispredictions are dominated by false positives \jake{(0.35 precision $<$ 0.63 recall)} while false negatives dominate the EMIT pixelwise mispredictions \jake{(0.71 precision $>$ 0.52 recall)}. Several factors contribute to the mismatched pixelwise performance of the \ANG and EMIT models on their respective validation scenes.

The primary difference lies in the methodology used to define pixelwise plume labels for the airborne platforms versus the EMIT platform. \jake{As described in Section~\ref{sub:plumelabels},} EMIT plume labels are inclusive by design, while \ANG plume labels are intentionally conservative. EMIT plume labelers are instructed to include a small pixel buffer around any plume they annotate, so the labels for each EMIT plume instance will capture some background pixels outside of each plume's boundary. These low-concentration pixels account for the lower recall \jake{(0.52; higher false negative rate)} on the EMIT plumes. In contrast, the \jake{CMF}-guided \ANG plume labels only include high-concentration pixels near each plume's origin point that exceed our 500\ppmm background enhancement threshold. As a result, the pixelwise labels of \ANG plume instances often do not capture the full spatial extent of each plume, often excluding boundary pixels and ``disconnected'' plume components. These unlabeled pixels account for the \jake{lower pixelwise precision (0.35; higher false positive rate)} on the \ANG validation scenes. Additionally, the EMIT VISIONS data set only includes high SnR CMFs approved by multiple reviewers, while the \ANG validation scenes were inspected and reviewed by fewer reviewers overall.

While differences in plume labeling methodologies broadly account for the mismatched distributions of pixelwise \jake{error} between \ANG and EMIT results, sensor and retrieval-specific differences \jake{also} help explain the gap in pixelwise precision between the \ANG and EMIT results \jake{(0.35 and 0.71, respectively)}. EMIT is a more efficient sensor than \ANG, and CMFs derived from EMIT data use an improved, full-VSWIR-based retrieval compared to the original SWIR2-only retrieval approach applied to \ANG data from early airborne imaging campaigns. Compared to the \ANG CMF products, plume and background pixels tend to be better separated in terms of local pixelwise contrast, and fewer false enhancements are present in EMIT CMFs. This is also reflected in the EMIT instance detection performance, where the multitask model performs better in both \jake{recall (0.59) and precision (0.82).}

\subsection{Salience Map Comparison Between Models}

\begin{figure}
  \centering
  \includegraphics[width=1.0\linewidth,clip,keepaspectratio]{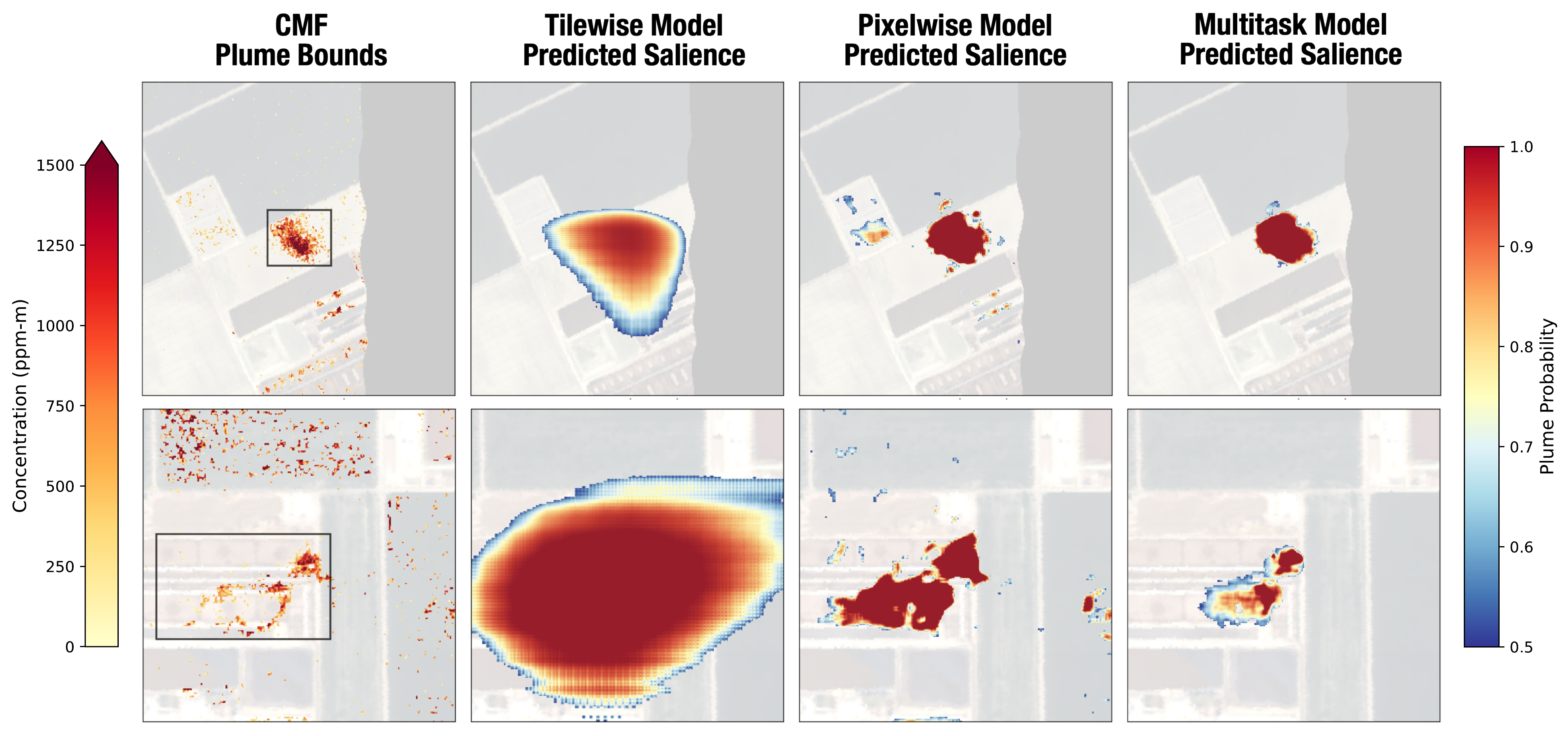} 
 \caption{\jake{\textbf{CMF Plume Bounds:} Example \methane plumes and their label mask bounding boxes. False enhancements outside of the bounding boxes are caused by surface materials. \textbf{Predicted Saliences:} Each column shows predicted plume salience maps generated by each model. The salience map produced by the multitask model is able to accurately delineate each plume extent while ignoring the high concentrations of the background false enhancements.}}
 \label{fig:salcmp}
\end{figure}

Figure~\ref{fig:salcmp} shows two example \methane plumes and the corresponding plume salience maps produced by the tilewise, pixelwise, and multitask models.  The tilewise model produces detections that are typically much larger than the extent of the each plume, and tilewise detections occasionally merge plumes and neighboring false enhancements. However, the tilewise model generates fewer false positive instance detections than the pixelwise and multitask models, and also comes with the distinct advantage of not requiring pixelwise labels during training. 

The pixelwise model generates detections that fit the boundaries of high concentration plume enhancements well, but also generates more spurious false detections than the other models. Most of these false detections correspond to small, ambiguous enhancements or image artifacts that are more effectively rejected by the other two models. However, the pixelwise model is more effective than the other models for plume bodies split into several adjacent disconnected components.

The multitask model provides a compromise between instance detection and pixelwise segmentation, and generates compact plume detections that fit high contrast plume boundaries more accurately than the other models, while also generating fewer false positive instance detections than the pixelwise model. However, because the multitask model generates more conservative detections than the other models, these improvements come at the cost of a higher false negative rate.

\subsection{Evaluating Regional Generalization} \label{sub:tilewise_region}
\begin{figure}
 \centering \includegraphics[width=1.0\linewidth,clip,keepaspectratio]{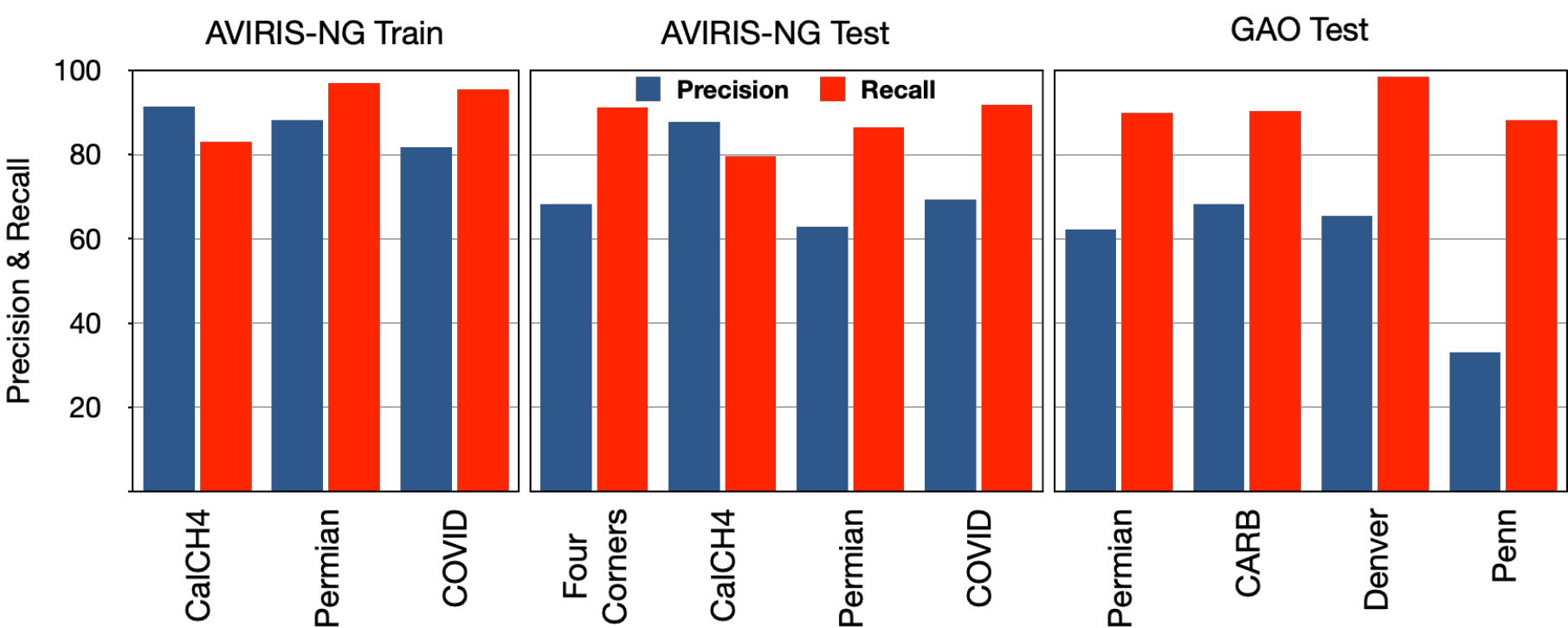}
 \caption{Tilewise prediction accuracy of \ANG mulitcampaign model on select \ANG and \GAO imaging campaigns. Blue bars give the Precision (FDR=1-precision). Red bars give the Recall (FNR=1-recall).}
 \label{fig:tilewise_test}
\end{figure}

We evaluate \jake{the} generalization performance \jake{of the \ANG multicampaign model} across distinct regions observed in \ANG and \GAO airborne imaging campaigns. However, since we do not have scenewise quality-controlled data for the \GAO campaigns, we restrict our evaluation to tilewise instance classification performance results. Figure~\ref{fig:tilewise_test} summarizes the tilewise classification accuracy of the \ANG multicampaign model applied to the \ANG train/test tiles and the \GAO test tiles detailed in Table~\ref{tab:datatab}. 

\jake{In all cases, the lower precision indicates that the false positives constitute most of the prediction errors. Still, the multicampaign model yields similar results on the \ANG \CALMETHANE, Permian, and COVID train/test datasets, as expected. On the \ANG Four Corners test dataset, the multicampaign model performs similarly to the other \ANG datasets, as the Four Corners region is geographically similar to California and the Texas Permian Basin. This is also reflected in the results for the \GAO Permian and CARB (California), and Denver (Colorado) datasets.}

\jake{However, the model has a significantly reduced precision on the \GAO Penn (Pennsylvania) dataset, which is geographically distinct from the rest of the regions. This confirms that the regional generalization of the \ANG multicampaign model is limited to regions that share the geographical characteristics of California and Texas. While not demonstrated here, we would expect the \EMIT model trained on globally sampled plumes and background scenes to have further improved global generalizability. We also present experiments where we train and evaluate models trained on individual campaigns in Suppl. Section~\ref{apx:bias_region}, demonstrating that training on multiple campaigns is necessary for regional generalization.}

\begin{figure*}
 \centering \includegraphics[width=1.0\linewidth,clip,keepaspectratio]{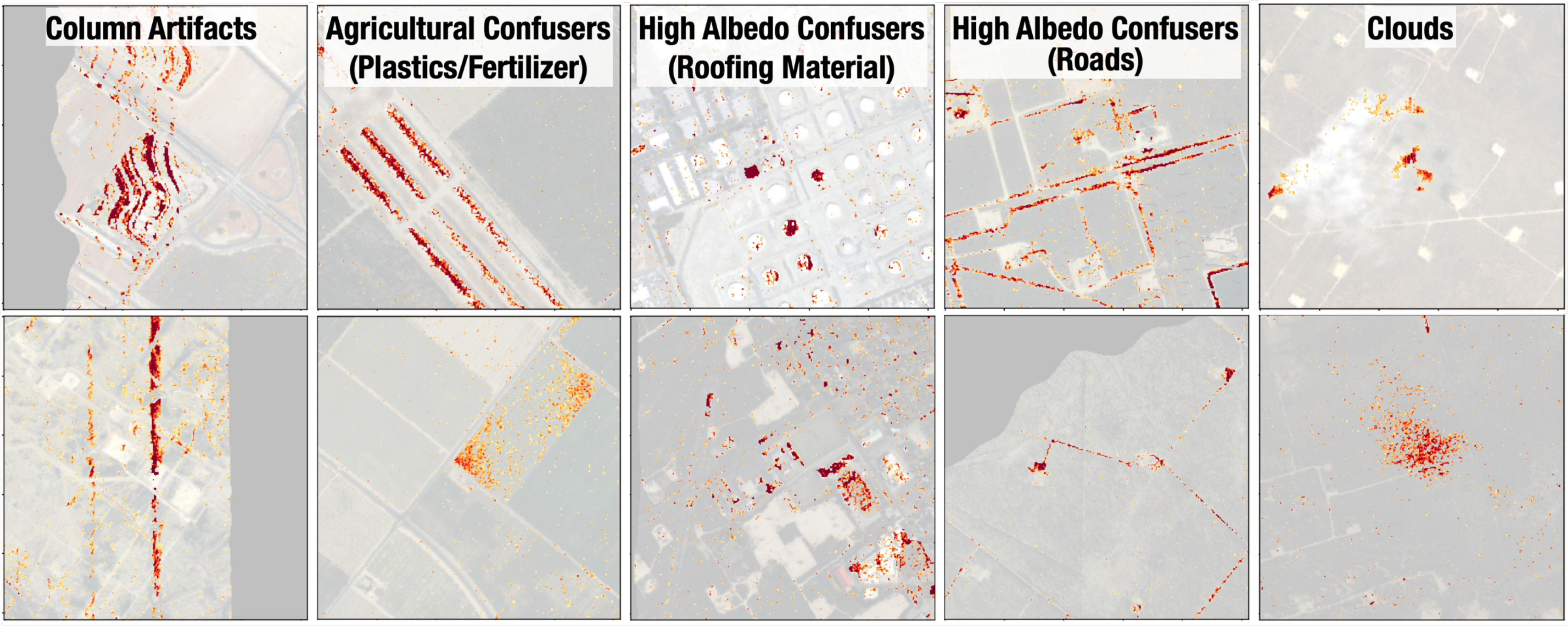}
 \caption{\jake{Example false positive predictions by the \ANG multicampaign model due to false enhancements, including columnwise artifacts (column 1), spectral confusers from agricultural materials (column 2), high-albedo spectral confusers from rooftop (column 3) and road (column 4) materials, and clouds (column 5).}}
 \label{fig:mispred_falsepos}
\end{figure*}

Figure~\ref{fig:mispred_falsepos} shows example false positive mispredictions produced by the \ANG multicampaign model. All are false enhancements with concentration levels consistent with real \methane plumes, though their morphologies vary according to the the phenomena that produce them. For instance, linear false enhancements that follow the downtrack imaging direction are CMF artifacts produced when outliers corrupt the background covariance matrix used by the retrieval algorithm, while other linear false enhancements are caused by confuser materials (e.g., agricultural plastics, fertilizer, rooftop, or road surfaces with absorptions consistent with some hydrocarbons) coupled with sharp local changes in albedo. 

We note that our comprehensive tile sampling strategy which selects tiles that cover the spatial extent of each scene plays a crucial role to ensure validation metrics capture realistic estimates of generalization performance. Due to the rarity of false enhancements, without comprehensive tile sampling, such phenomena would not be discovered. 

\begin{figure*}
 \centering \includegraphics[width=1.0\linewidth,clip,keepaspectratio]{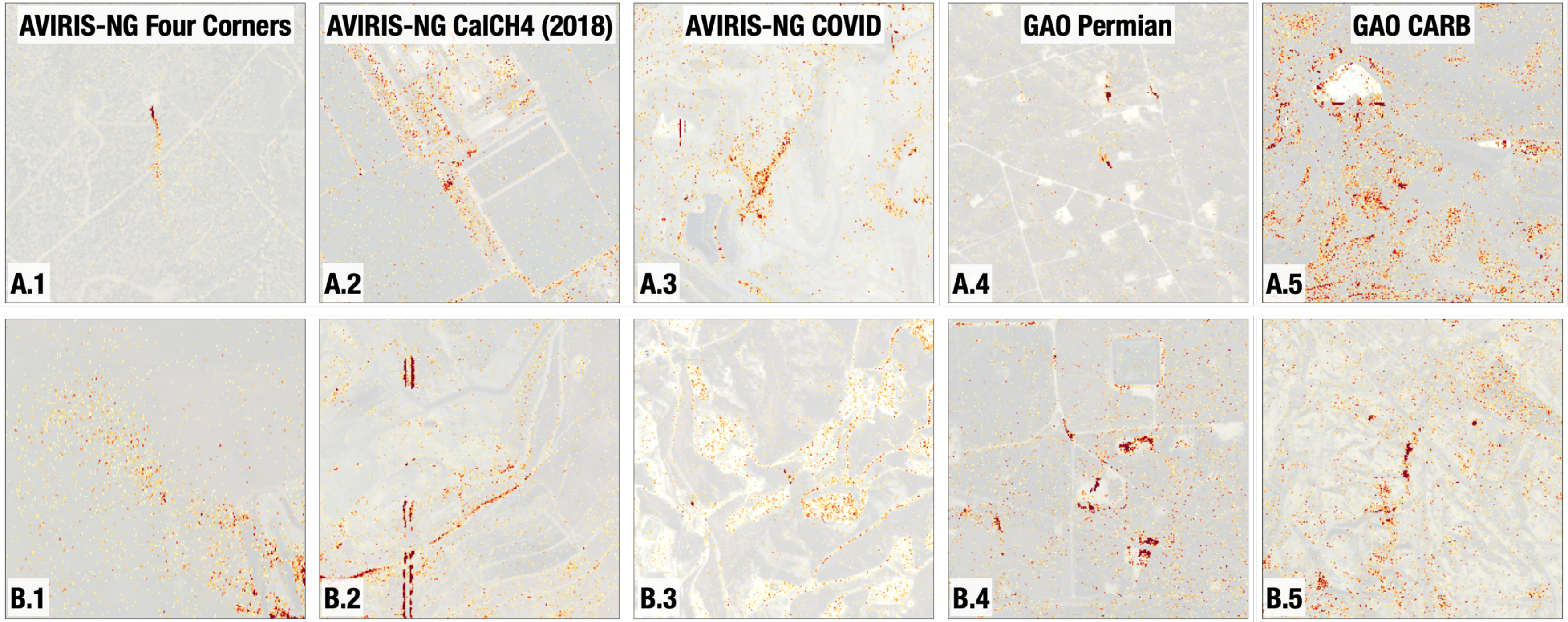}
 \caption{Selected false negatives mispredictions from \ANG~\FOURC (column 1), \CALMETHANE (column 2) and Permian (column 3) campaigns; and \GAO Permian (column 4) and CARB (column 5) campaigns.}
 \label{fig:mispred_falseneg}
\end{figure*}

Investigating the characteristics of plume mispredictions (false negatives) also yields informative diagnostics to assess the capabilities of the plume detector. Figure~\ref{fig:mispred_falseneg} shows randomly-selected false negatives from the \ANG~\FOURC, \CALMETHANE and Permian campaigns, and the \GAO Permian and CARB campaigns. Some mispredictions are due to small (A.3, B.4-B.5) or indistinct (A.4-5,B.2-3) plumes or spatially diffuse plumes surrounded by comparably high background enhancements (A.1, A.2). These likely approach the detection limits of the current plume detector model or retrieval algorithm. Others are likely driven by the presence of adjacent surface confusers (A.4-5, B.4-6) and/or adjacent columnwise artifacts (B.2) that the plume detector is trained to reject.

\subsection{Measuring Airborne Plume Detectability}

\jake{We evaluate the limits of our models' plume detectability based on the physical characteristics and source sectors of the labeled plumes in our test dataset. As discussed in Section~\ref{sub:science_objectives}, we do not consider instances where enhancements fall below the range of instrument detectability. Of course, detectability limits are intrinsically tied to the spatial resolution and sensitivity of the instrument, platform, and retrieval, not only the model performance. In this section, we focus on plumes observed by the \ANG and \GAO airborne campaigns.}

\subsubsection{Impact of Physical Characteristics on Detectability}

\jake{We begin by assessing the positive and negative detections of plumes from the \ANG and \GAO test datasets by the \ANG multicampaign multitask model. We exclude the \ANG Four Corners and GAO Penn datasets due to anomalies in data quality further described in Suppl. Section~\ref{apx:bias_region}. We characterize the physical characteristics of each plume by determining its area of delineation and the average \methane concentration within that area. Sorting the plumes by these two properties allows us to highlight correlations with model performance. For the purposes of this analysis, we use the tilewise False Negative Rate (FNR), as each tile is cropped around each plume.}

\begin{figure*}
 \centering \includegraphics[width=1.0\linewidth,clip,keepaspectratio]{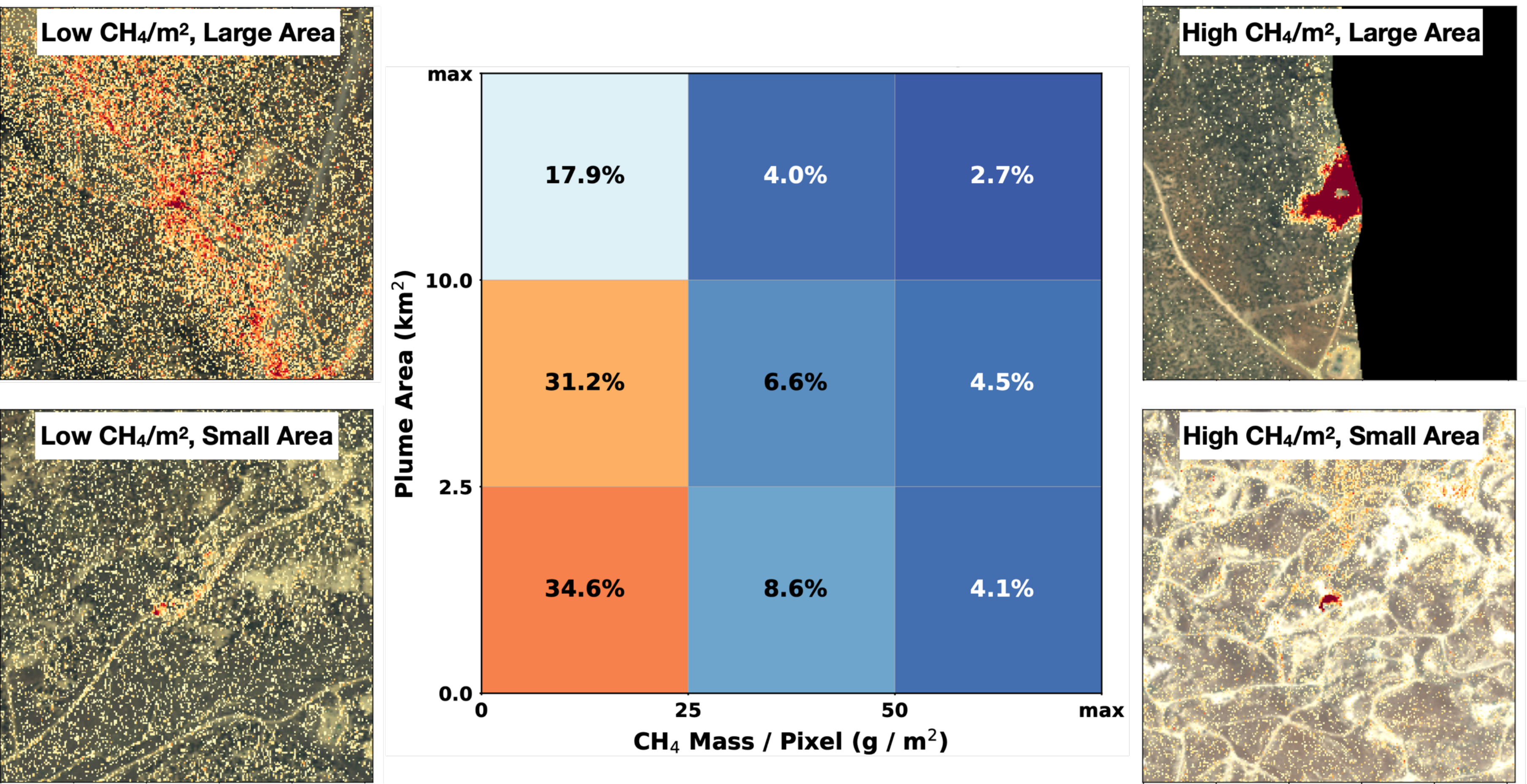}
 \caption{Tilewise \jake{false negative detection rate} of test set \ANG and \GAO plumes with respect to \methane concentration (x-axis) and plume area (y-axis). Characteristic tiles for low and high concentration plumes are shown on the left and right, respectively, and characteristic tiles for small and large plumes are shown on the bottom and top, respectively.}
 \label{fig:ch4vsarea}
\end{figure*}

\jake{Figure~\ref{fig:ch4vsarea} shows these tilewise FNR scores calculated for binned ranges of plume area (y-axis) and average \methane concentration (x-axis). As expected, most mispredictions occur at low average pixelwise concentrations ($<25$ \gpmsq) where plume enhancements are less distinguishable from background enhancements. Further, such weak concentration plumes with also small areas ($<2.5$ \kmsq) are the most commonly mispredicted, with a false negative rate of 34.6\%. These are especially difficult to distinguish from false enhancements of similar size, as there are less distinct differences in spatial morphology---such as the high albedo false enhancements shown in Figure~\ref{fig:mispred_falsepos}. By contrast, plumes with similarly small area but higher concentration (25 to 50 \gpmsq) are detected far more reliably (8.6\% FNR), whereas plumes with similarly small concentration but larger area (2.5 to 10.0 \kmsq) are still missed often (31.2\% FNR). Despite the larger plume area, a plume with concentrations that cannot be reliabily distinguished from the background cannot be detected.}

\subsubsection{Impact of Source Sector Imbalance}

\begin{figure}
 \centering
 \includegraphics[width=0.475\linewidth,clip,keepaspectratio]{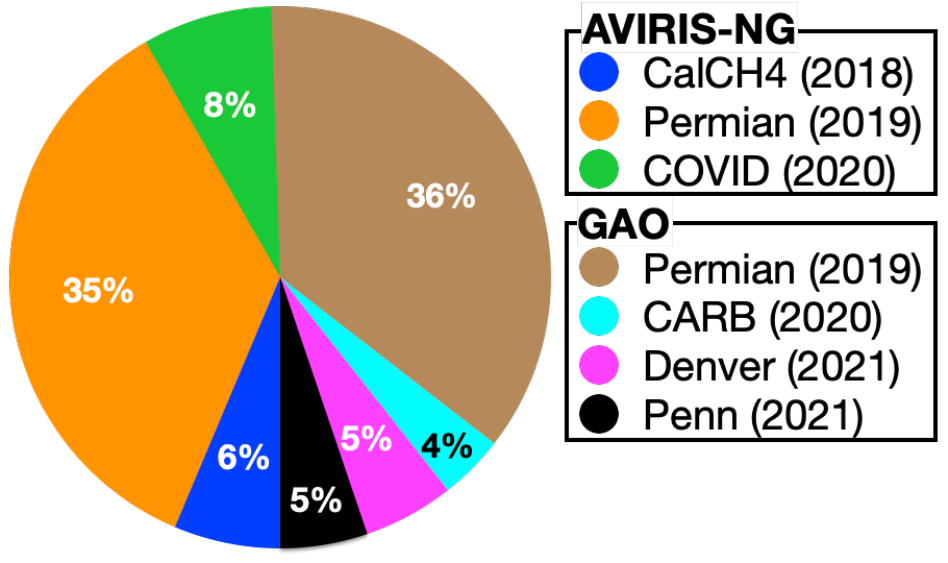} 
 \includegraphics[width=0.475\linewidth,clip,keepaspectratio]{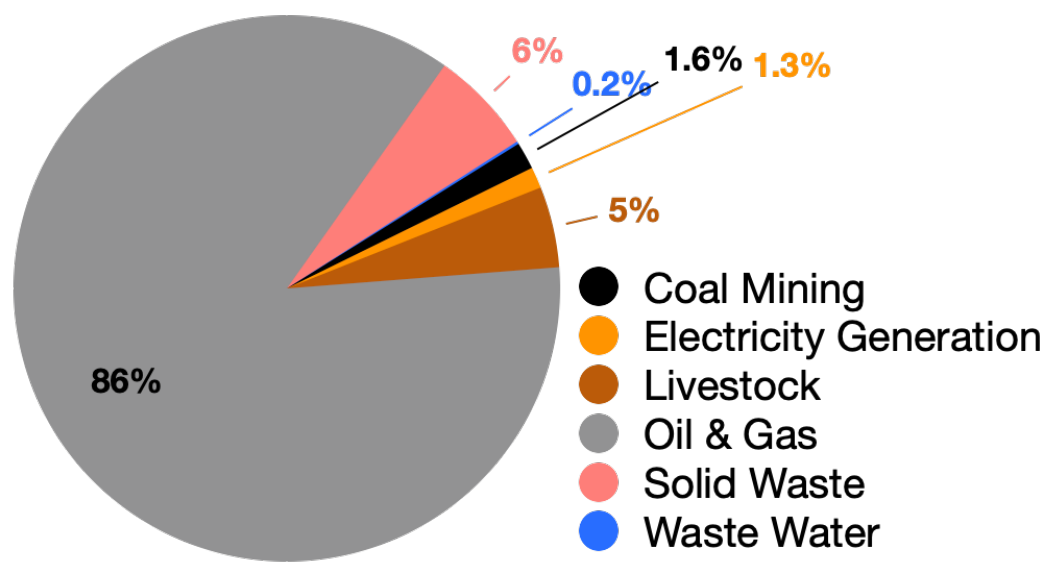} 
 \caption{Left: distribution of identified emission sources observed in select airborne campaigns. Right: distribution of airborne sources by sector.}
 \label{fig:percent_campaign_sector}
\end{figure}

\jake{In addition to identifying \methane plumes in the \ANG and \GAO campaign datasets, human experts also determined the IPCC sectors of each plume based on identified ground infrastructure. The distribution of source sectors (coal mining, electricity generation, livestock, oil \& gas, solid waste, and waste water) for sources observed across these campaigns is shown in Figure~\ref{fig:percent_campaign_sector}. As expected from the US regions observed, oil \& gas sources dominate the dataset (86\%), with few sources attributed to solid waste (6\%) and livestock (5\%), and minimal representation of coal mining (1.6\%), electricity generation (1.3\%), and waste water (0.2\%). In such cases of high sector imbalance, we expect data-driven machine learning models to fail on severely underrepresented classes, as there are not enough examples to learn from. We also base this on the fact that plumes from different sectors have different observed characteristics; for example, oil \& gas plumes from flare stacks are at higher pressure and temperature than livestock plumes from manure management systems.}

\jake{We investigate these effects by calculating the average False Negative Rate across campaigns for each source sector. We again exclude the \ANG Four Corners and GAO Penn datasets, which also excludes the coal mining sector (mostly represented by the \GAO Penn campaign). In this analysis, we calculate the FNR for both the training and test sets. A perfect machine learning model would have a 0\% FNR on any plumes in the training dataset (as it will have learned from them), but in practice single-digit FNRs are expected.}

\begin{figure}
 \centering
 \includegraphics[width=0.85\linewidth,clip,keepaspectratio]{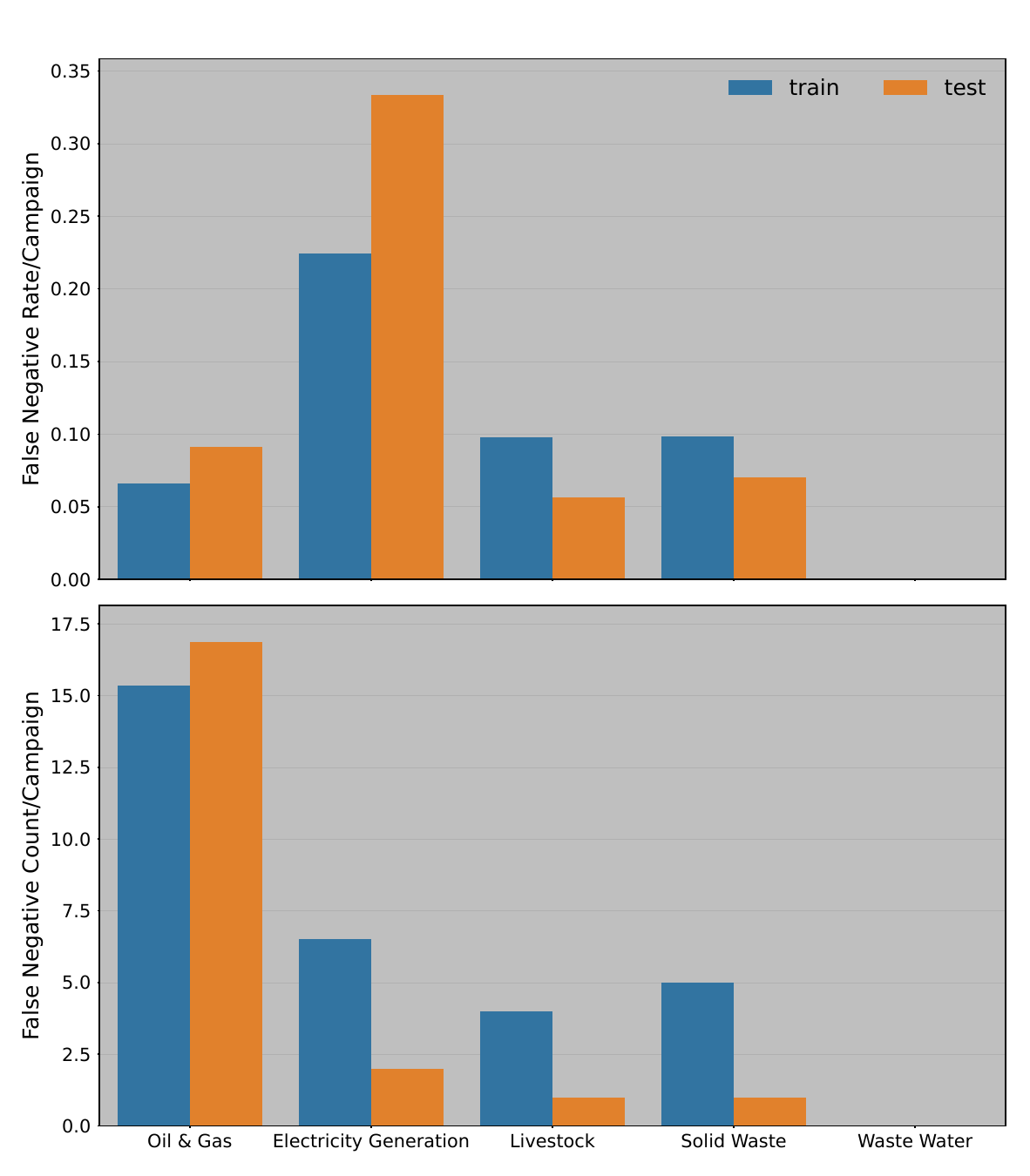} 
 \caption{Top: Average train and test plume false negative rate across campaigns for plumes attributed to source sectors observed in selected \ANG \& \GAO \methane campaigns. Bottom: Average number of false negatives across campaigns with respect to the same source sectors.}
 \label{fig:ipccfnr}
\end{figure}

\jake{Figure~\ref{fig:ipccfnr} shows the average FNR (top) and the average number of false negatives (bottom) across campaigns with respect to the source emission sectors of the identified plumes. For oil \& gas, a well-represented sector (86\%), the training set FNR is very low and the test set FNR is slightly higher, as expected. The average number of false negative detections dominates that of other sectors, only because it already represents the vast majority of all identified sources. Further analysis of these false negative detections reveals that many of these plumes have very small areas, which are difficult to detect as discussed previously. For lesser-represented sectors, livestock (5\%) and solid waste (6\%), both training set and test set FNRs remain below 10\%, indicating there was sufficient representation in the training set for the model to learn how to detect these plumes. It is also possible that these plumes shared similar morphological features as plumes attributed to other sectors.}

We observe a significantly higher false negative rate for plumes associated with Electricity Generation (EG) sources---\jake{0.22 on the training set and 0.33 on the test set}---relative to the other sectors. \jake{This indicates that 1.3\% of the dataset distribution was not sufficient for the model to learn to detect plumes attributed to EG. This effect is further exacerbated by the fact that} all of the plumes we observed from EG sources occur in urban areas with diverse surface materials and substantial albedo variation. As shown in Figure~\ref{fig:source_plumes}, CMF images representing complex and diverse urban areas often contain similarly complex false enhancements, with many occurring in the relative vicinity of observed plumes. \jake{A more representative sample of EG plumes were necessary for the model to learn to detect them robustly.}

Interestingly, we observe the lowest FNR (0\%) of all sectors on wastewater treatment facilities (0.2\% of all observed airborne sources) which also occur in similarly diverse urban areas, but are also likely driven by the lack of representative samples. Rather than assigning a systematic justification, we conclude these plumes may have been sufficiently high concentration to be detected regardless of sector-correlated factors, or the model may have been lucky on the few plumes included in the evaluation.

\section{Discussion} \label{sec:discussion}

In this work, we present a GHG plume detection system developed with the fundamentals necessary for operationalization. A state-of-the-art deep learning plume detection model optimized for both pixelwise segmentation and instance detection stands on a strong foundation of multicampaign, quality controlled, and appropriately sampled methane plume data. Here, we discuss literature gaps we address, open issues that remain, and future research directions to address current challenges.

\subsection{Addressing Common Limitations}

\subsubsection{Biases in Data Sampling}

\parsub{Background Sampling:}
Background samples play a crucial role in constructing and validating a robust plume detector. Collecting a representative negative class dataset (i.e. the background) is as important as collecting a representative positive class dataset (i.e. plumes), and failure to do so will result in poor generalization. \jake{Our work addresses this potential issue by sampling as many background tiles from scenes as possible. As described in Section~\ref{sec:methods} and Table~\ref{tab:datatab}, the plume-to-background ratio of the AVIRIS-NG dataset used to train our models was less than 1:13, an unbalanced sampling that more accurately reflects the observed rarity of high \methane concentrations, which allows our models to perform well when evaluated on entire scenes without tile sampling in Section~\ref{sub:scenewise_eval}.}

\jake{By contrast, sampling an equal number of plume and background tiles for model development, as in \cite{ruzicka_semantic_2023}, results in a biased representation of the true observed distribution, in which plume pixels are a distinct minority.} Sampling such few background tiles also misses many of the rare false enhancements that drive most false positive detections. Therefore, a model trained and evaluated on datasets with equal quantities of the plume and background will appear to perform well, but it will generalize poorly operationally when encountering the true observed distribution and the most challenging false enhancements. We illustrate this in detail in Suppl. Section~\ref{apx:bias_sample} and Figure~\ref{fig:instance_tilebias}.

\parsub{Avoiding Spatiotemporal Bias and Data Leakage:}
CNN-based plume detectors are prone to memorizing, or ``overfitting'' to, site-specific features that occur in spatially overlapping observations of the same site, or from overlapping image tiles extracted from the same retrieval image scene. When overlapping images are present in {\em both} training and test sets, data leakage occurs. Validation procedures corrupted by data leakage generally produce highly optimistic validation metrics, as the model is able to ``cheat'' on test samples that overlap training samples with spatially correlated features. \jake{Our work makes a significant effort to control for spatiotemporal bias and prevent data leakage in our tile sampling and cross validation procedures (described in Section~\ref{sec:methods}) by ensuring that spatially overlapping observations are assigned to {\em either} the training or the test dataset, but not both. As a result, a plume nor its source location (whether emitting or not) will never appear in both the training and test datasets. }

\jake{By contrast,} \cite{schuit_automated_2023} reports precision and recall scores over 0.97 using a CNN to distinguish $32\times32$ tiles of labeled plume and non-plume tiles from TROPOMI XCH4 products between 2018-2020. However, the $\sim850$ plume samples are derived from observations of only 60 regions with persistent emission sources, making it likely that many of the plumes represent repeat observations of the same emission sources. When tiles with 50\% overlap are sampled from each TROPOMI image, then randomly split into training and test sets without stratification, data leakage inevitably occurs, likely accounting for the abnormally high precision and recall detection performances. \jake{\cite{schuit_automated_2023} reports a more realistic precision score of 0.61 for confident plume detections and 0.76 for both confident and ambiguous detections on TROPOMI XCH4 products from 2021.} However, these results are reported after an undefined filtering process and calculated from a \textit{post hoc} analysis of detection results instead of a previously labeled dataset, therefore omitting false negative analysis.

We \jake{further} demonstrate the potential impacts of data leakage in Suppl. Section~\ref{apx:bias_stratify} by inducing data leakage with a random non-stratified train-test split and showing its results. \jake{One such experiment with data from the Four Corners campaign showed that a random train-test split resulted in an inflated F1 Score of 0.75, whereas the stratified train-test split where data leakage was mitigated resulted in the actual F1 Score of 0.41. Other experiments with the CalCH4 (random 0.92; stratified 0.80) and Permian (random 0.86; stratified 0.81) datasets showed similar results, whereas the COVID (random 0.79; stratified 0.81) did not show this effect.}

\parsub{Replicable Data Collection \& Sampling}: Assessing the operational effectiveness of a ML-driven GHG plume detection approach requires detailed accounts that describe the assumptions and practices involved in data collection, tile sampling and cross-validation. Specifics describing how samples were collected and the scope and diversity of both the plume class and the background class are necessary to determine if they are sufficiently representative to construct the model and assess its effectiveness and the extent which known sources of bias might impact reproducibility on similar data, and potential replicability in broader operational settings. For instance, if a proposed approach does not specify whether sampled tiles are allowed to overlap within scenes, across distinct scenes, or between plume and background samples, we cannot assess the impact of spatiotemporal bias on the associated model's predictions, nor can we replicate results on similar data. To-date, detailed data collection and model validation procedures have been sorely lacking in the literature; we aimed to address this by thoroughly describing data sampling strategies in Section~\ref{sec:methods}, supported by Suppl. Sections~\ref{apx:datainfo} and \ref{apx:qc}. \jake{We also release our dataset of sampled tiles for \ANG and EMIT, as described in the Data and Code Availability statement.}

\subsubsection{Informative Performance Metrics}

Even with proper sampling, tilewise plume detection performance metrics (Figure~\ref{fig:tilewise_test}) tend to be optimistic estimates of scenewise plume detection accuracy (Figure~\ref{fig:scenewise_pre_rec}). This gap between tilewise and scenewise performance is exacerbated by any undersampling of the background class, as discussed previously. Since operational plume detection systems are deployed on entire scenes and not sampled tiles, it is crucial to report scenewise \jake{performance metrics} (as in Section~\ref{sub:scenewise_eval}) in addition to any tilewise \jake{performance metrics}. \jake{While tilewise metrics remain useful for comparing model performance (as in Section~\ref{sub:tilewise_region}), they are insufficient for the practical evaluation of operational applicability.}

\jake{Measuring the operational performance of a proposed plume detector requires measuring both instance detection and pixelwise segmentation performance as defined in Section~\ref{sub:science_objectives}.} As Figure~\ref{fig:scenewise_pre_rec} illustrates, a model which performs poorly in pixelwise plume segmentation metrics (e.g., our tilewise-trained model) can significantly outperform comparable models in plume instance detection metrics. Recognizing that certain use cases may prioritize either segmentation (e.g. plume masking for emission rate estimation) or detection (e.g. identifying sources for review), reporting both metrics is necessary for complete operational evaluation. \jake{This comprehensive approach addresses a common limitation in the literature, where most prior work only report the tilewise metrics for which a model was optimized.}

Finally, capturing detection and segmentation accuracy with respect to false positive and false negative prediction errors is also important to provide an unbiased view of a proposed model's capabilities. ``Coupled'' metrics such as traditional classification accuracy can provide a misleading view of performance, especially when samples are imbalanced. In cases where a single metric is necessary for comparison, F-$\beta$ scores, where $\beta$ weighs recall relative to precision, may be appropriate in addition to precision and recall.

\subsubsection{The Simulation Gap of WRF-LES Datasets}

A common theme among proposed plume detection approaches is to construct synthetic GHG plume observations by injecting plumes with known characteristics generated by the WRF-LES system into plume-free ``background/noise'' retrieval products derived from data captured by the target platform. \jake{This approach may be attractive because ``perfect'' label masks can be generated for these simulated plume based on known concentrations. However, all such works in the literature are limited in evaluating their WRF-LES-based detection systems on observed data captured by the target remote sensing platforms for operational deployment.}

Prior works that only evaluate their WRF-LES-trained models on other WRF-LES plumes \citep{jongaramrungruang_methanet_2022, bruno_u-plume_2023} are difficult to discern for operational feasibility due to the significant differences between plumes generated by WRF-LES and real-world plumes. Other prior works do discuss results of their WRF-LES-trained models on observed methane plumes \citep{joyce_using_2022, radman_s2metnet_2023, si_unlocking_2024}; however, they only report a couple dozen true positive detections given small scenes containing previously identified plumes, or perform emission rate quantification on manually detected plumes. The omission of false positive or false negative detections on a large number of instrument-observed scenes makes it difficult to determine the capabilities of these methods in an operational setting.

In order to determine the capabilities of WRF-LES trained models when applied to observational data, we reproduced the plume detection results of \cite{jongaramrungruang_methanet_2022} and compared the capabilities of the WRF-LES-trained models to \ANG-trained models in Suppl. Section~\ref{apx:bias_distrib}. The results demonstrate poor generalization, with the WRF-LES-trained models failing to perform beyond a 0.3 F1-Score in tilewise instance detection on \ANG test datasets despite reporting a nearly 1.0 F1-Score on the WRF-LES test dataset. This demonstrates that discrepancies between WRF-LES simulations and CMF GHG retrievals pose a significant challenge to the operational deployment of models trained solely on WRF-LES plumes, despite being injected on real-world backgrounds. \jake{Discrepancies in this case include the lack of changes in wind speed and direction over time and as the plume rises in altitude, as well as the lack of diffuse emission simulations, which make up a significant portion of real world emissions and the \ANG dataset.}

Spatiotemporal bias and data leakage also likely contribute to the high overall detection accuracies reported in WRF-LES-based results, as most prior works inject multiple versions of WRF-LES generated plumes into the same observed ``background'' retrieval image, or worse, repeatedly inject a single WRF-LES plume instance into multiple distinct backgrounds. Allowing these samples to cross training and test boundaries will introduce severe validation bias, as the model will be tested on plumes and backgrounds identical to those in the training set. The former is especially common in prior works, with many training datasets composed of many WRF-LES simulated plumes injected into a smaller collection of background images \citep{joyce_using_2022, radman_s2metnet_2023, bruno_u-plume_2023, jongaramrungruang_methanet_2022}. If the same plume-background images appear in both the training and test datasets, it increases the likelihood that the CNN learns to make plume detections based on background surface features seen during training (e.g. roads, infrastructure, geographic features) instead of the plume itself (by ``overfitting''). Deliberate care must be taken to ensure background tiles do not leak between the training and test datasets.

\subsection{Open Issues}

One challenge we encountered in comparing our results to published work is the use of nonstandard retrievals as input to plume detector models. For instance, \cite{ruzicka_semantic_2023}, use mag1c \citep{foote_fast_2020} retrievals, while \cite{kumar2023methanemapper} used newly contributed linear regression model. As a result, we cannot easily attribute differences in detection performance to the plume detection model as opposed to deviations in products derived by the respective retrieval approaches. While alternative retrieval methods likely address some limitations of the CMF retrieval approach, the relative benefits and costs involved in replacing the CMF with an alternative retrieval have only been established a handful of curated scenes. 

Since false enhancements are the primary driver of prediction errors, retrieval methods that suppress or eliminate false enhancements simplify plume detection. For instance, while EMIT uses exactly the same matched-filter-based retrieval algorithm as the airborne CMF retrievals we consider in this work, the EMIT retrievals exploit the full VSWIR wavelength range of the sensor rather than just the SWIR2 \methane absorption used in the airborne retrievals. EMIT also leverage dynamically-generated \methane template spectra that account for scenewise characteristics \cite{foote_impact_2021}. These updates significantly reduce the number of false enhancements from surface confusers observed in typical EMIT scenes, and retroactive application to existing airborne observations has shown similar results. Additionally, recent work by \cite{fahlen_sensitivity_2024} has the potential to suppress false enhancements driven by varying albedo  or illumination conditions.

Some classes of false enhancements cannot be rejected using models driven by the CMF alone. Augmenting CMF-driven CNN with diagnostic input bands can suppress some classes of currently unidentifiable enhancements, as shown by  \cite{satish_altinput_2023}.

\subsection{Future Work}

Given our objective of constructing plume detector models that perform as well as domain experts, controlled user studies that assess which plumes are consistently identified by multiple domain experts and the variability among their provided plume labels would provide valuable insights. Prior works have demonstrated the utility of GHG plumes manually identified by domain experts, and the results of such studies would (i) inform operational accuracy requirements for deployed models, (ii) provide a basis to assess relative tradeoffs between plume detection and pixelwise segmentation objectives, and (iii) connect results from automated, statistical plume detection approaches with prior analysis estimating the probability of plume detection, which are currently driven by manual plume identification procedures. 

While advances in sensing capabilities and retrieval algorithms will provide commensurate improvements in plume detection capabilities, much work remains to address the scope and diversity of global observations captured by spaceborne platforms like EMIT and  the Carbon Mapper Coalition's Tanager-1. The improved plume detection and segmentation accuracy on the EMIT data relative to the \ANG results (Figure~\ref{fig:scenewise_pre_rec}) provide some supporting evidence that detection capabilities improve in tandem with advanced hardware and retrieval methods. Indeed, many of the problematic false enhancements we encountered in early airborne imaging campaigns were driven by limitations of the \ANG SWIR2-only CMF retrieval approach. These improvements also demonstrate that the quality control procedures applied to the publicly released EMIT \methane products are effective. 

Data triage routines will also play a crucial role in a fully operational plume detection system. Rather than applying a computationally expensive plume detector en-masse to all data products, efficient data triage routines can optimize compute resources by selectively applying the detector to scenes with adequate SnR. For instance, scenewise summary statistics are often sufficient to detect severe data quality issues such as high background enhancement scenes covered by false enhancements (e.g. Figure \ref{fig:mispred_falseneg} B.2). Cross-referencing high concentration enhancements with relevant external data products such as derived mineralogical maps, cloud masks, and geoprocessing metadata (e.g., pixelwise sensor-to-ground path lengths) can disambiguate otherwise unidentifiable false enhancements. Targeted data triage routines that leverage relevant external data sources can effectively reduce false positive detections driven by limitations of the GHG retrieval approach.

\section{Conclusions} \label{sec:conclusions}
Automating GHG plume detection from remotely sensed imagery has proven to be a deceptively difficult machine learning problem requiring high-quality, interpretable labeled image retrieval products that capture diverse GHG emission sources and their surrounding background enhancements from distinct regions under varying imaging conditions, along with systematic, rigorous validation procedures aligned with operational science objectives. Interdisciplinary expertise is essential to define protocols that unambiguously characterize generalization performance of a candidate plume detection system, and to identify and account for hidden model and data-driven biases that impact generalization capabilities. 

This work demonstrated the potential of ML-driven plume detection as a viable route to automate GHG plume detection in operational settings. Provided representative plume and background image retrieval products derived from data captured by current imaging spectrometers, we showed that established ML models equipped with rigorous optimization and validation procedures lead to results that are both \jake{reproducible and replicable on observations with analogous characteristics.} 

While previously published work claimed promising proof-of-concept results using ML-driven approaches for plume detection,
\jake{our review revealed that a variety of methodological issues discussed above, including undersampling the background, the simulation gap between WRF-LES and observed plumes, and spatiotemporal biases, likely explain overly optimistic performance metrics compared to real-world deployments.}
Furthermore, due to the absence of crucial details regarding data collection, model optimization, and/or validation procedures, the majority of published results are not reproducible experimentally. Prior efforts all lack rigorous validation procedures to ensure that trained models generalize on independent and identically distributed observational data that generally lead to optimistic results driven by data and model-driven biases. Consequently, even in cases where the published results are feasible to reproduce, successfully replicating published accuracies on distinct but analogous data is unlikely. Here, we take steps to address these gaps in the literature, with the goal of pushing towards operational deployments of automated plume detection models.

\section*{Data and Code Availability}

\ANG and EMIT datasets as described are available at \url{https://doi.org/10.5281/zenodo.19011045}. \GAO datasets are not available for public access due to data controls, and only evaluation results are presented as reference. Dataset sampling and preprocessing code, as well as model training, inference, and evaluation code, is open source and available at \url{https://github.com/JPLMLIA/operational-ghg}.

\section*{Acknowledgments}

The Earth surface Mineral dust source InvesTigation (EMIT) plume delineations were performed using a tool built with the Jet Propulsion Laboratory Multi-Mission Geographical Information System. Carbon Mapper acknowledges the generous support of its philanthropic donors. The High Performance Computing resources used in this investigation were provided by funding from the JPL Information and Technology Solutions Directorate. EMIT is supported by the National Aeronautics and Space Administration Earth Venture Instrument program, under the Earth Science Division of the Science Mission Directorate. The research was carried out at the Jet Propulsion Laboratory, California Institute of Technology, under a contract with the National Aeronautics and Space Administration (80NM0018D0004). © 2025. California Institute of Technology. Government sponsorship acknowledged.

\appendix

\section{Data Table Column Definitions} \label{apx:datacols}

\begin{itemize}
\item Sensor Platform: imaging spectrometer sensor platform used to collect survey data set.
\item Survey Data Set: survey / data set identifier.
\item Survey Regions: regions indicated by US state name or 'global' measured in data set.
\item Survey Start/End (Date): begin / end dates of image acquisitions captured in data set.
\item Unique Scenes (\#): Total number of labeled CMF scenes in data set.
\item GSD (\msq): Average Ground Sampling Distance (GSD) in \msq of all scenes in data set.
\item Length (km): Average length in km of all scenes in data set.
\item Unique Plumes (\#): Total number of manually identified methane plume candidates in labeled scenes in data set.
\item Plume Tiles (\#): Total count of plume tiles in data set.
\item Bg Tiles (\#): Total count of background tiles in data set. 
\item Quality Control (\# Reviewers): Quality control level and (number of reviewers / scene). 
\item Train:Test (Val) Scenes (\#): Total training and test scenes used in tilewise optimization and (total test scenes used in scenewise validation).
\end{itemize}

\section{Airborne Plume Source Sectors} \label{apx:campaign_sectors}

\begin{figure}
 \centering
 \includegraphics[width=0.95\linewidth,clip,keepaspectratio]{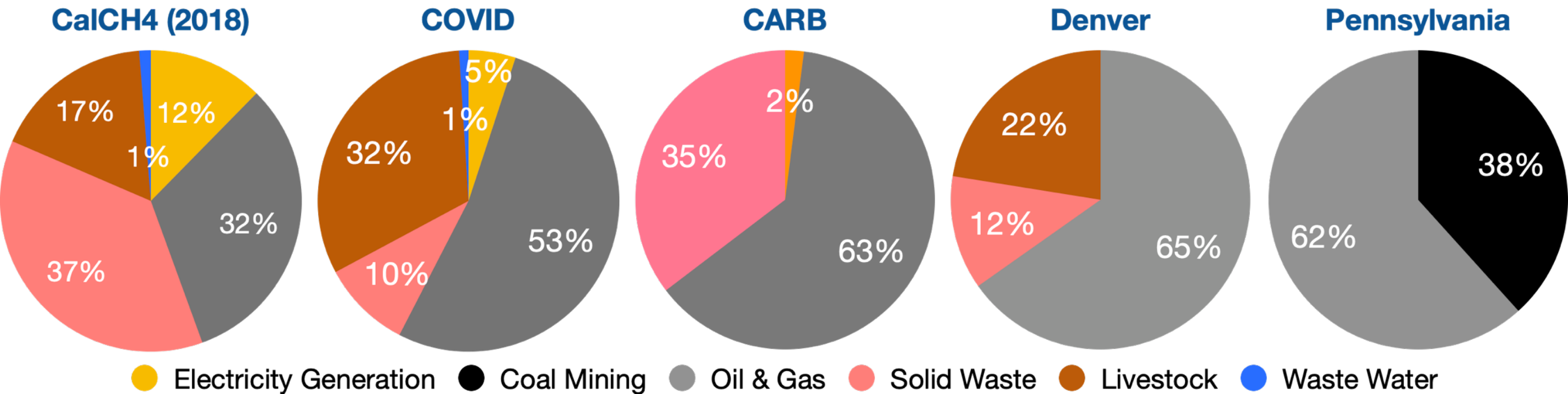} 
 \caption{Distribution of identified emission sources sectors observed in each airborne campaign. All sources from the \ANG Four Corners campaign, and both the \ANG and GAO Permian campaigns represent Oil \& Natural Gas sources, and are not shown here.}
 \label{fig:campaign_sectors}
\end{figure}

The distribution of source sectors for plumes observed in airborne imaging campaigns are provided in Figure~\ref{fig:campaign_sectors}.


\bibliographystyle{elsarticle-num-names-nourl} 
\bibliography{opsghgml}




% \bibitem[()]{}

% \end{thebibliography}
\end{document}

\begin{frontmatter}



\title{Supplemental Information: Towards Operational Automated Greenhouse Gas Plume Detection}

\author[jpl]{Brian~D~Bue}
\author[jpl]{Jake~H~Lee}
\author[jpl]{Andrew~K~Thorpe}
\author[jpl]{Philip~G~Brodrick}
\author[cm]{Daniel~Cusworth}
\author[cm]{Alana~Ayasse}
\author[jpl,pri]{Vassiliki~Mancoridis}
\author[jpl,cit]{Anagha~Satish}
\author[jpl,col]{Shujun~Xiong}
\author[cm]{Riley~Duren}

\affiliation[jpl]{organization={Jet~Propulsion~Laboratory, California~Institute~of~Technology}, 
%            addressline={4800~Oak~Grove~Drive}, 
            city={Pasadena},
            state={CA},
            postcode={91101}, 
            country={USA}}

\affiliation[cm]{organization={Carbon~Mapper~Inc.},
%            addressline={680~E~Colorado~Blvd~Suite~180}, 
            city={Pasadena},
            state={CA},
            postcode={91101}, 
            country={USA}}

\affiliation[pri]{organization={Princeton University},
            city={Princeton},
            state={NJ},
            postcode={08544},
            country={USA}}
          
\affiliation[cit]{organization={California~Institute~of~Technology}, 
%            addressline={1200~E~California~Blvd}, 
            city={Pasadena},
            state={CA},
            postcode={91125}, 
            country={USA}}
          
\affiliation[col]{organization={Columbia~University},
%            addressline={116th~and~Broadway}, 
            city={New~York},
            state={NY},
            postcode={10027}, 
            country={USA}}          

\end{frontmatter}


\section{Data Description} \label{apx:datainfo}

\subsection{GHG Retrieval Data \& Processing}\label{apx:ghg_retrieval}
Here we provide a summary of the GHG retrieval process, along with the criteria domain experts leverage to identify plumes using data collected in \ANG and \GAO airborne imaging campaigns, and from spaceborne data observed by EMIT. \ANG and \GAO are high-performing operational imaging spectrometers that share equivalent specifications and employ analogous radiometric calibration and geometric processing procedures. EMIT uses an optically efficient Dyson spectrometer design capable of capturing high SNR observations that approach the fidelity of field spectroscopy \citep{coleman_accuracy_2024}. All three sensors were designed and built by the Jet Propulsion Laboratory, California Institute of Technology. \cite{chapman_spectral_2019}, \cite{asner_carnegie_2012} and \cite{thompson_orbit_2024} detail the radiometric calibration and data processing pipelines for \ANG, \GAO, and EMIT, respectively.

In this work, we derive pixelwise GHG concentrations from calibrated imaging spectrometer radiance observations using the Columnwise Matched Filter (CMF) \citep{thompson_real_2015} retrieval approach. Analogous to \cite{villeneuve_improved_1999}, the CMF is conceptually simple, computationally efficient, and produces retrievals that are numerically stable\footnote{i.e., similar phenomena observed in similar conditions yield similar retrievals} across repeat observations of the same site and consistent \footnote{i.e., prior measurements by the same instrument and coincident measurements by other instruments agree in their interpretation} with coincident measurements by complementary sensors in multiplatform campaigns and ground measurements in controlled release experiments (see e.g., \cite{thompson_real_2015,thorpe_airborne_2017,thorpe_methane_2023,thorpe_mapping_2016,duren_californias_2019,cusworth_intermittency_2021,sherwin_single-blind_2023,ayasse_performance_2023}). The CMF retrieval computes the pixelwise enhancement of VSWIR absorption features associated with a lab-measured GHG transmittance spectrum in \ppmm units relative to multivariate Gaussian backgrounds, each estimated columnwise from the observations from the independent detector elements in the sensor's focal plane array. CMF \methane products derived from \ANG and \GAO data are functionally identical -- both are based on the approach described in \cite{thompson_real_2015} that retreives pixelwise \methane concentrations from airborne spectrometer radiance observations captured at Ground Sample Distance (GSD) $\in$ [1,10]\msq according to the depth of the SWIR2 \methane absorption feature. EMIT CMF products are derived using the same method, but are computed at GSD $\approx$ 60\msq using all VSWIR wavelengths, and also account for scaling factors driven by \water concentration and varying solar illumination \citep{thorpe_attribution_2023}.

When the surface coordinates of GHG point sources are known, CMF retrievals have been effective for estimating point source emission rates from plumes observed in typical surface wind conditions, as shown by several independent multisensor methane controlled release experiments \citep{duren_californias_2019,thompson_real_2015,thorpe_mapping_2016,sherwin_single-blind_2023}. CMF products have played a crucial role in the field and in real time onboard aircraft during imaging campaigns, where the operator-identified plumes efforts inform follow-up flight plans and enable mitigation efforts.

\section{Data and Label Quality Control} \label{apx:qc}
Pixelwise CMF image products have been the primary product leveraged by domain experts to manually locate \methane emission sources in dozens of airborne imaging spectrometer campaigns spanning nearly 10 years. Since 2022, the spaceborne EMIT mission has leveraged CMF images in a formalized, multi-user identification, delineation and review proccess to identify \methane plumes observed globally. Dedicated efforts by domain experts to comprehensively inspect CMF images post-campaign have yielded catalogs of plumes and their corresponding sources and sectors observed in each campaign. Insights derived from analysis of these plume catalogs enabled scientists to compute regional/sector-scale emissions, informed statewide GHG policy (e.g. \CALMETHANE and CARB campaigns), provided deeper understanding of the relative impacts of superemitters and persistent sources \citep{duren_californias_2019, cusworth_intermittency_2021}, and informed advanced retrieval approaches \citep{fahlen_sensitivity_2024} and sensor designs for follow-up missions \citep{zandbergen2023preliminary}.

In this section, we summarize the lessons learned from exploratory analysis of data from early airborne \methane imaging campaigns, along with data and label quality control procedures for \methane CMF image products derived from said analysis. Given our ultimate objective of achieving plume detection performance comparable to human experts, high-quality observations that human experts can label with confidence are necessary to assess progress. Even in targeted GHG imaging campaigns, not all CMF products are informative for---or relevant to---plume detection tasks, and not all expert-identified plume candidates are readily identified based solely on their corresponding CMF images. Consequently, much time and effort was devoted to data and label quality control to ensure that only high-quality data with informative plume labels are used to train and validate our models. 

% Comprehensive manual inspection and quality control of CMF products from airborne GHG imaging spectrometer campaigns has played a invaluable role in our efforts to construct and validate ML-driven plume detection systems. Training and validating a ML-driven plume detector on ambiguous data / labels can only lead to ambiguous results. 

\subsection{Airborne GHG Imaging Campaigns and Plume Identification} \label{apx:airborne}

Airborne GHG imaging spectrometer campaigns involve the targeted collection of image data in regions where known GHG emission sources are present. The objectives and scope of these GHG imaging campaigns began as small scale exploratory surveys (e.g., Four Corners) aimed to comprehensively measure and localize specific types of emission sources (e.g., Oil \& Natural Gas) within a relatively small survey area (Four Corners covered a $\sim50$\kmsq bounding box) and have evolved to regional surveys (e.g., \CALMETHANE, Permian) tracking diverse source sectors and/or persistent sources via repeat measurements in targeted regions of much larger survey areas ($\sim500$-750\kmsq bounding box). Between 50-300 images are collected to measure GHG plumes within the campaign survey bounds over the course of 1-2 months as conditions permit. Most airborne surveys so far captured data at relatively low altitudes, yielding orthorectified image products with GSD between 2-4\msq / pixel, though some surveys (e.g., \ANG Permian) captured data at higher altitudes, yielding image retrieval products with GSD between 5-8\msq / pixel.

After each airborne imaging campaign is complete and the required CMF retrieval products have been generated, each scene is visually inspected by scientists to ensure the data are of sufficient quality to identify plumes. Low quality scenes are excluded from GHG-related follow-up analysis and optionally flagged for review if any quality control issue(s) demand further investigation. For the remaining high quality scenes, candidate plume enhancements are cross-referenced with the coregistered true color images captured by the imaging spectrometer along with fine spatial resolution base map imagery (provided via e.g., Google Earth) to exclude potential false positives. The ``origin points'' and IPCC sectors of the emission sources associated with each high confidence plume candidate are then determined through visual identification of characteristic surface infrastructure in fine resolution base map images, guided by each plume's location, shape, and local wind conditions.

\subsection{Profiling CMF Enhancements} \label{sub:cmfprofile}

Large scale plume detection requires GHG retrievals that are numerically stable and consistent with prior retrievals of similar phenomena (i.e., GHG plumes versus background substrate) observed by the same sensor in comparable imaging conditions. To ensure the CMF retrieval images form the early airborne \methane imaging campaigns satisfied these conditions, we measured the CMF-retrieved background enhancements in each scene observed by each of the 598 independent detector elements in the \ANG and \GAO focal plane arrays. Each of these "CMF BackGround Enhancement (BGE) profiles" is a 598-dimensional vector representing the downtrack median of the positive CMF enhancements per detector in a single $N \times 598$-dimensional CMF image prior to orthorectification. Comparing the CMF BGE profiles averaged over all scenes captured in each airborne imaging campaign allowed us to assess the numerical stability of the CMF retrieval approach and its variability across distinct regions and imaging conditions. 

\begin{figure*}
 \centering
 \includegraphics[width=0.9\linewidth,clip,keepaspectratio]{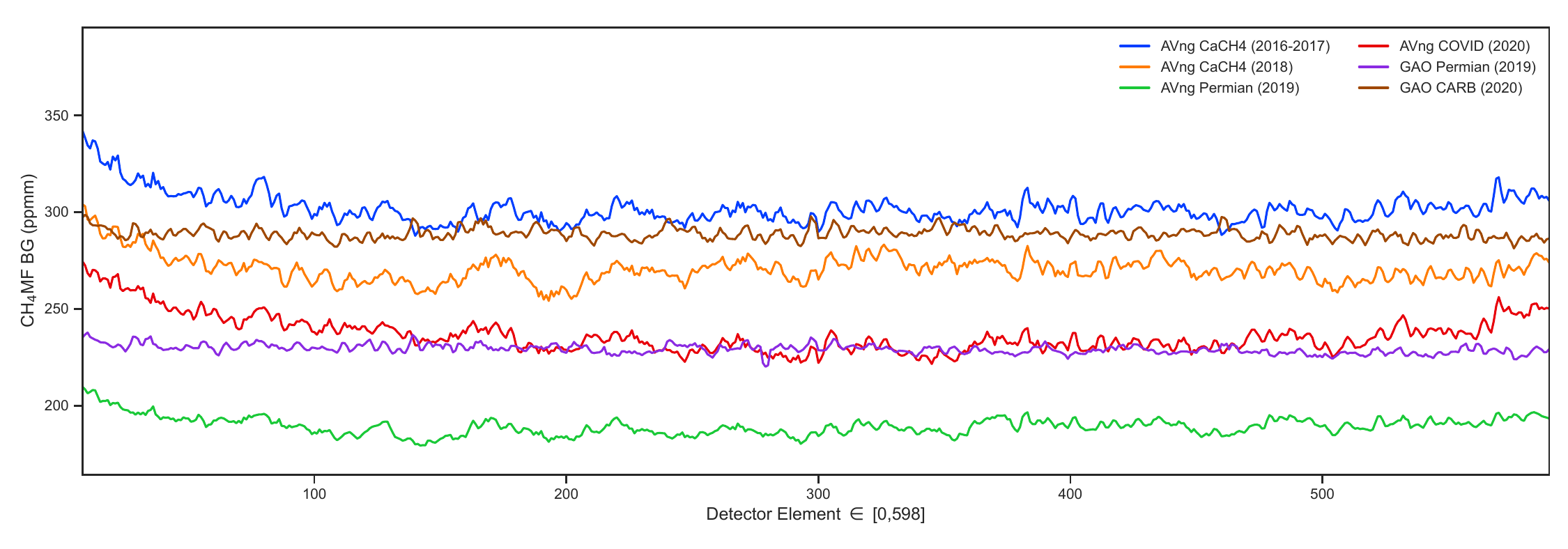} 
 \caption{Median CMF \methane column enhancements averaged across all flightlines observed in \ANG and \GAO airborne imaging campaigns.}
 \label{fig:cmfbge}
\end{figure*}

Figure~\ref{fig:cmfbge} shows the median background enhancements per CMF column averaged across flightlines observed in each airborne \ANG and \GAO imaging campaign. We discuss the distinct characteristics of the GAO Penn campaign in detail in Section~\ref{apx:gao_penn} that contribute to its higher BGE (median=524 \ppmm) relative to prior airborne campaigns. For the remaining \ANG and \GAO imaging campaigns, we observe relatively small variations among the columnwise BGE estimates across distinct campaigns. The Median and Median Absolute Deviation (MAD) of the columnwise BGE estimates from the \ANG campaigns are $\hat{\mu_{ANG}}=243.3$, $\hat{\sigma_{ANG}}=78.7$, respectively. The Median and MAD BGE for the \GAO CARB and Permian campaigns are $\hat{\mu_{GAO}}=280.1$ and $\hat{\sigma_{GAO}}=29.9$, respectively.

We arrived at several conclusions from our campaign-level BGE analysis: 

\begin{enumerate}[I]
\item On average, BGE levels estimated from most \methane CMF observations captured in the selected airborne imaging campaigns are numerically stable and consistent across campaigns that observe similar regions in similar climate zones. 
\item As a consequence of (I), we can assume that on average, \methane plumes whose maximum enhancements fall within the BGE range are undetectable using the associated CMF retrieval products. 
\item As consequences of (I) and (II), our plume detection experiments should focus on enhancements exceeding the BGE range where plumes are feasibly detectable.
\item As consequences of (I-III), campaigns and/or individual CMF products whose average BGE levels exceed the upper limit of the BGE range are outliers and should be manually reviewed to assess data quality \& plume detectability.
\end{enumerate}

Based on our preliminary analysis of the 2018-2020 \ANG campaigns, we selected a BGE threshold of 500\ppmm ($\hat{\mu_{ANG}} + 3\hat{\sigma_{ANG}}=479.4$). Subsequent analysis of CMF plume summary statistics from \ANG and \GAO campaigns confirmed that plumes with max enhancements below 500\ppmm were very rare (less than one per campaign on average), and were erroneous (incorrect origin coordinates) or mislabeled (no observable plume present at the origin point of a controlled release observation) in some instances. Additional follow-up analysis of EMIT CMF products detailed in Section 6.1 of the EMIT GHG ATBD and in \cite{mancoridis_adaptation_2023} suggests the 500 \ppmm BGE threshold is satisfactory for use with the EMIT CMF products. 

\subsection{CMF-Guided Plume Labels} \label{sub:cmflab}

We compute CMF-Guided plume label images for all CMF scenes we consider in this work for the following reasons:

\begin{enumerate}[i]
\item to address the lack of pixelwise plume labels from early airborne imaging campaigns prior to 2021
\item to sanity check expert-provided plume instances for errors and detectability
  \item to guide background tile sampling towards unlabeled CMF pixels in regions with highly concentrated enhancements
\end{enumerate}
	
Provided a set of plume instances consisting of either a set of origin point coordinates indexed by scene or a set of plume label masks indicating the pixelwise boundaries of plumes instances in each scene, we generate ``plume candidate vs. background'' masks by applying the BGE threshold to each relevant CMF image. We assume pixels with enhancements exceeding a fixed BGE concentration threshold potentially represent plumes or false enhancements, while pixels below the BGE threshold can be trivially assigned to the background class. As described in the previous section, we currently use a BGE threshold of 500 \ppmm for all \methane CMFs we consider in this work. 

\begin{figure*}
 \centering
 \includegraphics[width=1.0\linewidth,clip,keepaspectratio]{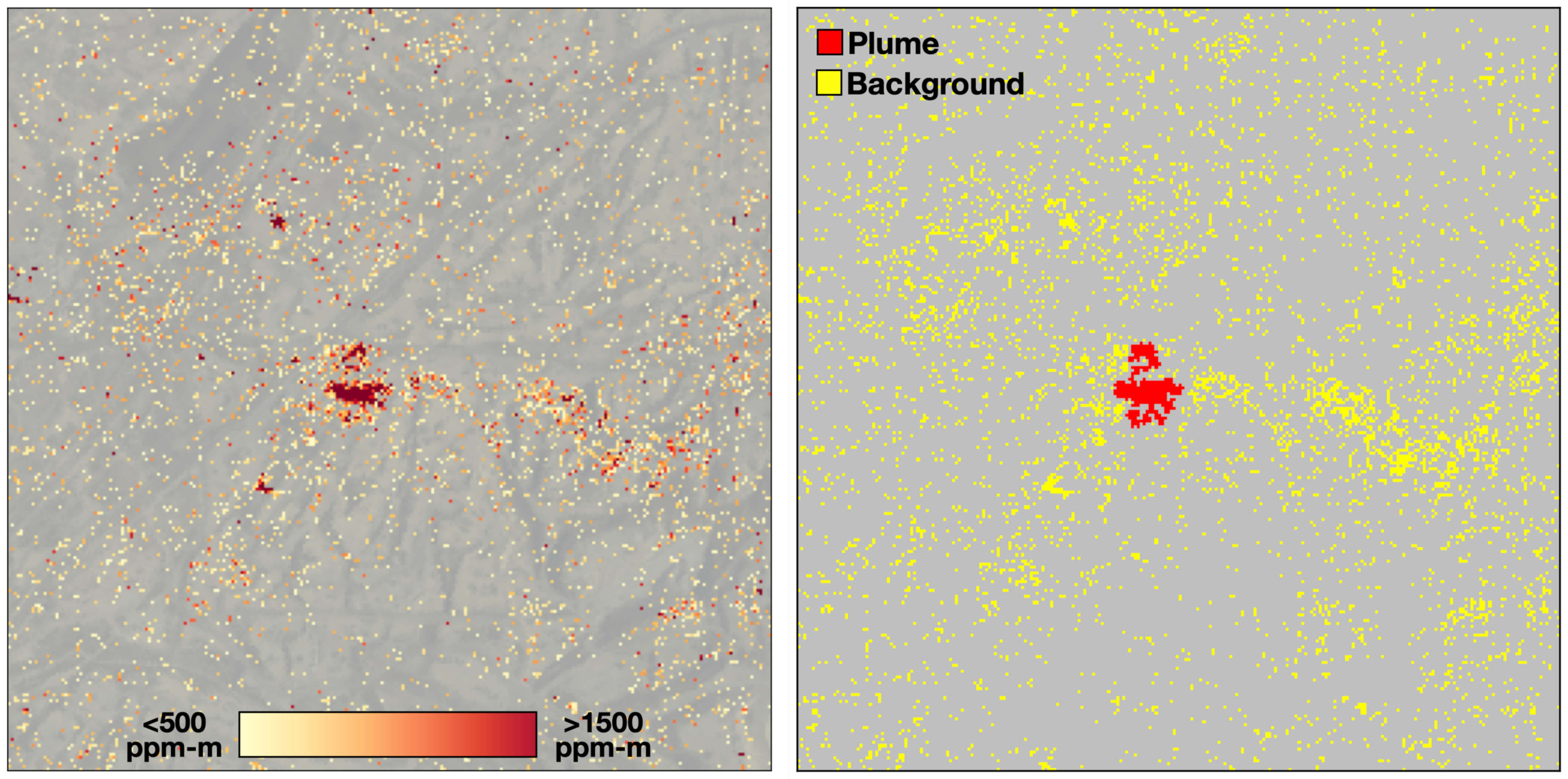} 
 \caption{Left: CMF \methane retrieval image. Center: CMF-guided plume pixel labels (red) derived from CMF enhancements $>$ 500\ppmm within 10px of the connected component that intersects the plume origin point (center pixel). High concentration non-plume pixels (yellow) pixels are prioritized when background tiles are sampled from each scene.}
 \label{fig:cmflab_zoom}
\end{figure*}

An example CMF-guided label image is shown in Figure~\ref{fig:cmflab_zoom}. Plume Regions of Interest (ROIs, red pixels) consist of those connected components in the mask that are less than 10 px from any component that overlaps an expert provided plume instance. We exclude any plume Regions of Interest (ROIs) and their corresponding plume instances whose plume ROI area is less than 16 px and whose maximum enhancement is less than 1000 \ppmm. This filtering step excludes very small, localized candidates whose enhancements typically cannot be confidently identified as plumes by human experts without additional context (e.g., high-resolution base map images, repeat measurements of known source locations, etc), and sometimes correspond to mislabeled or erroneous plume instances. The remaining non-plume ROIs (yellow pixels) represent high concentration pixels with enhancements above the BGE threshold. Since we cannot trivially assign these to the background class, we prioritize collecting background samples from these regions to train and validate our models. 

\subsection{Airborne \methane CMF and Plume Label Quality Control} \label{sub:angqc}

Because the CMF-guided \methane plume labels we use for the airborne imaging campaigns are generated via image processing functions, manual quality control is necessary to make sure the resulting plume labels are informative. We also must exclude scenes containing ambiguous enhancements from scenewise validation, along with any extracted CMF tiles containing ambiguous enhancements from tilewise training. Our quality control procedures involve three sequential stages, defined as follows: 

\begin{itemize}
\item Level 0: Triage: applies automated data triage routines to flag scenes with known and  easily detectable data quality issues to prioritize follow-up manual review. 
\item Level 1: Tilewise QC: selects plume tiles with unambiguous pixelwise plume labels and removes background tiles with ambiguous enhancements for tilewise training. 
\item Level 2: Scenewise QC: selects only high quality scenes with unambiguous pixelwise labels from those accepted by tilewise quality control. 
\end{itemize} 

{\bf Triage} applies predefined data triage routines designed to flag known, easily detectable data quality issues (e.g., high BGE scenes, columnwise CMF artifacts, orthoprocesing issues) to prioritize follow-up manual QC efforts, but performs no filtering. Both {\bf Tilewise QC} and {\bf Scenewise QC} apply the criteria defined below to accept or reject CMF image data according to pixelwise plume labels, but differ in scope and objectives. Scenewise QC applies the strongest filtering criteria we consider, requiring that entire scenes---including the pixelwise labels of all plume instances in each scene---be unambiguous. This ensures that scenewise validation metrics capture detection performance on high-quality, exhaustively labeled scenes that only contain plumes that human experts can confidently identify. In contrast, Tilewise QC is more permissive, since only the data and labels within the extent of each sampled tile must be unambiguous. This allows tiles with satisfactory pixelwise labels sampled from high-quality regions of any scene---including those rejected by scenewise QC---to be included in tilewise optimization. 

\parsub{CMF \& Plume Label QC Criteria}: We apply the following criteria to accept or reject each CMF image we inspect according to the quality of their observed enhancements and corresponding pixelwise plume labels. All plume instances  in the CMF must be exhaustively labeled and each plume instance must be identifiable as a real GHG plume with high confidence based solely on visual inspection of the CMF alone. While prior plume identification efforts allow external data (e.g., high resolution base map images or GIS databases) to confirm observed CMF enhancements represent plumes, we only allow the RGB quicklooks captured by the imaging spectrometer in our QC process to ensure that the CMF products alone are sufficient for detection. Images containing {\em any} ambiguous enhancements that cannot be identified as plumes or false enhancements with sufficient confidence must be rejected. Small, high-concentration plumes ($<100$px area with maximum enhancement $>1000$\ppmm) and weak diffuse plumes (large plumes with maximum enhancements near 500\ppmm) may only be included if they are obviously distinct from surrounding background enhancements. Pixelwise plume labels must be roughly aligned with CMF pixels with distinctly higher concentration levels than their surrounding lower concentration background pixels. When identifiable false enhancements are present, they must be spatially disjoint from pixelwise plume labels. CMF images that satisfy all of the above criteria are accepted by default, and those that satisfy the above criteria for the majority of plume instances are accepted by user discretion for experimental use in tilewise optimization or scenewise validation experiments, as appropriate. All others are excluded from our experiments. 

\subsection{EMIT CMF \& Plume Label Quality Control}
 \label{sub:emitqc}

The EMIT mission uses a formalized set of plume identification, pixelwise labeling and multi-user review procedures detailed in the EMIT GHG Algorithms ATBD, sections 4.2-4.3. These procedures ensure the CMF scenes and constituent manually labeled plumes released to the public-facing VISIONS portal satisfy mission-specific QC criteria. All EMIT scenes and plumes in the EMIT VISIONS data set satisfy these QC criteria, and consequently meet or exceed the Scenewise QC criteria applied to airborne CMF images. 

\subsection{Interpreting Quality Controlled Products}

While no quality control effort is perfect, we can safely assume that all CMF scenes approved via scenewise quality control (along with their constituent tiles) and all tiles approved via tilewise quality control satisfy the following criteria:
	
(a) all plumes are identifiable via inspection of the associated CMF images alone, and the pixelwise labels associated with each plume satisfy domain expert expectations 
(b) all unlabeled enhancements are identifiable as background or false enhancements via  visual inspection of the CMF image alone.
	
We note that regions of unlabeled ``background'' pixels in our quality controlled products will contain common false enhancements from confuser materials, albedo variation, measurement or processing artifacts. As we discussed in Section~\ref{sec:background}, and show in Figure~\ref{fig:bge_false} , these issues typically produce enhancements with spatial characteristics that are easily distinguishable from real GHG plumes. However, false enhancements with small spatial footprints and those that intersect (or are adjacent to) real GHG plume enhancements are not readily identifiable, and are consequently excluded  in our quality controlled data.

\section{Machine Learning Procedures and Models} \label{apx:ml}

\begin{table}[htb]
 \centering \includegraphics[width=1.0\linewidth,clip,keepaspectratio]{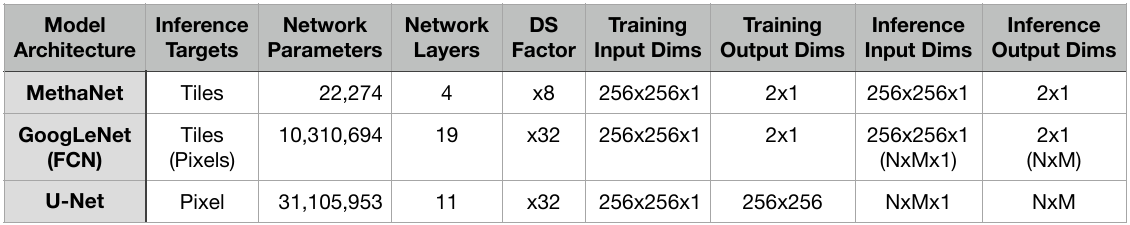} 
 \caption{ML Model Architecture Summary}
 \label{tab:model_table}
\end{table}

Table~\ref{tab:model_table} column definitions:
\begin{itemize}\addtolength{\itemsep}{-0.5\baselineskip}
\item Model Architecture: model identifier
\item Inference Targets: model predicts tilewise (classification) or pixelwise (segmentation) labels. 
\item Network Parameters: total \# of model parameters
\item Network Layers: total \# of model layers (note: GoogLeNet has 10 layers + 9 'inception blocks' where each block is a subnetwork with 4x2 layers)
\item DS Factor: internal downsampling factor after applying all pooling layers
\item Training Input Dims: Dimensions of training image tiles 
\item Training Output Dims: Dimensions of training image labels \& output predictions (2x1=tilewise classification, 256x256=pixelwise segmentation)
\item Inference Input Dims: Dimensions of input images during inference (256x256x1=tilewise classification, NxMx1=pixelwise segmentation) I
\item Inference Output Dims: Dimensions of output predictions during inference (2x1=tilewise classification, NxM=pixelwise segmentation)
\end{itemize}

Table~\ref{tab:model_table} gives a summary of the model architectures we consider in this work. All models were trained on $256 \times 256$ px tiles extracted from CMF flightlines using the sampling methodology described in Section \ref{sec:methods}. Excluding negative CMF enhancements by clipping negative concentrations to zero is standard practice in plume identification, as they represent retrievals that are not physically realizable. We also clip concentrations to a 4000 ppm-m upper bound, which captures the 95th percentile range of the labeled plume enhancements observed in the \ANG multicampaign dataset. Clipping input values to an upper bound prevents extremely large outliers from destabilizing gradients during model training. A preliminary assessment in \cite{mancoridis_adaptation_2023} also found that the [0,4000] \ppmm range is sufficient to capture the 95th percentile of labeled plume enhancements in 327 plume complexes observed by EMIT. As such, we apply the same [0,4000]\ppmm clipping range to both \ANG and EMIT CMF tiles, although future follow-up is necessary to ensure the practical utility of this range for the global observations captured by EMIT. 

All models are initialized with random weights and trained only using the tile images and corresponding labels associated with each experiment. Although not included in this work, we performed a basic assessment of models ``pre-trained'' on three-channel RGB image databases such as ImageNet \citep{deng2009imagenet} or COCO \citep{lin2014coco}, then fine-tuned with RGB-converted CMF images. We found that the models trained from scratch using single channel CMF image data consistently outperformed the pre-trained/fine-tuned RGB models on several single-campaign AVIRIS-NG classification and segmentation tasks, and did not investigate further.

During training, the training dataset is augmented by randomly applying axis-aligned rotations $\in \{0,90,180,270\}^\circ$ followed by identity, transpose, vertical flip, or horizontal flip operations. Applying these transformations prevents the model from learning orientation-specific characteristics of the training samples, while preserving the geometric characteristics of the observed \methane enhancements. Axis-aligned transformations do not require resampling that could create aliasing artifacts, and do not introduce edges of empty pixels at the boundaries of rotated or rescaled tiles.

We train each model for a maximum of 200 epochs using the Adam optimizer \citep{kingma_adam_2014} with fixed learning rate of 0.003. According to extensive analysis of training and test loss curves and complementary pixelwise and tilewise metrics computed per-epoch, 200 epochs is sufficient for our models to converge without detrimental overfitting in the test losses or metrics on both large and small-scale experiments, and both training and test set metrics tend to plateau near 100-150 epochs. As such, while additional training epochs may yield improved generalization performance for some of the experiments we present in this work, we expect that the improvements will be marginal relative to the reported results. 

At inference time, each model outputs a plume salience score $\in [0.0,1.0]$ for each valid pixel in the input CMF image. Because these salience scores are not calibrated (i.e., they are not aligned with the $P(\text{plume}|\text{CMF})$ probability distribution) we cannot directly compare the numerical values of salience scores produced by distinct models. Instead, we compute a salience threshold for each model that maximizes the pixelwise F1-score on the training tiles. By applying this threshold on the pixelwise salience maps, we convert them to pixelwise binary masks of plume class predictions that can be easily compared between methods.
 
% - Tiles: 256x256x1 CMF tiles
% - Data normalization norm\_unit=clip(tile/4000,0,1)
% - Data augmentation
% - Optimizer: ADAM details (lr=0.003)

\subsection{Tilewise-Trained Model} \label{sub:tilewise}

% - CNN: 256x256 tile, 1024 nodes in last conv layer → 1024 x 256//F  x 256//F intermediate (F=8 → 1024 x 32 x 32, F=32 → 1024 x 8 x 8) → Pool to 1024x1x1 → FC maps 1024x1x1 to 1x1 class probability
% - FCN: Shift+Stitch applied to R x C image for $F \times F$ offsets (i,j), i $\in$ [0,F], j $\in$ [0,F] $\ra$ Ri x Cj pad256 image at shift (i,j) $\ra$ 1024 x Ri//F x Ci//F intermediate → Pool to 1024 x F x F  → Ri//F x Ci//F salience 
% - Replace fully connected layers with 1x1 convolutions
% - Replace top-level pooling layer with
%onverting Tilewise CNNs to FCNs for Plume Segmentation

The tilewise-trained model is initially a Convolutional Neural Network (CNN) image classifier only trained with tilewise labels that describe whether or not a methane plume is contained within each tile. However, by applying a modified version of the ``shift-and-stitch'' procedure described by \cite{long_fully_2015}, we convert this CNN classification model into a Fully Convolutional Network (FCN) image segmentation model that can generate pixelwise predictions provided any $N \times M$ px input image. 

We begin by training a GoogLeNet CNN model \citep{szegedy_going_2015} to predict whether a tile contains a methane plume. Given a $256 \times 256$ px image tile, the model produces a $C=2$ dimensional output vector of the softmax probabilities for the ``plume'' and ``not plume'' classes. While this work focuses on GoogLeNet, the same methodology is applicable to other CNN architectures, but must be adjusted according to their respective downsampling factors. GoogLeNet applies five pooling operations that each downsamples by a factor of 2, resulting in a collective downsampling factor of $F = 2^5 = 32$, convolving from $1 \times 256 \times 256$ to $1024 \times 8 \times 8$. At the end of the architecture, an average pooling layer further reduces the representation to a 1024-dimension vector, which is then mapped to class probabilities by a final fully connected layer and a softmax function.

%We start with an existing  CNN image classifier pre-trained to predict a 2-dimensional output vector that gives the softmax probability that our $D \times D = 256 \times 256$ px image tile contains a plume. We focus on the original GoogLeNet CNN architecture here, but the same procedure is applicable to similar architectures adjusted according to their respective downsampling factor. In the case of GoogLeNet, the model applies a total of five pooling operations that each downsample the outputs of learned convolutional filters by a factor of 2. These operations collectively convolve our $256 \times 256$ input tile to a pooled representation downsampled by factor $F = 2^5 = 32$, yielding an internal representation with dimensions $1024 \times (256//F) \times (256//F)$ = $1024 \times 8 \times 8$. An average pooling layer then reduces the pooled representation to a 1024-dimension vector, which is then mapped to a class probability by applying the softmax function to the C-dimensional output of a final fully connected layer. Here, $C=2$ is the number of classes. 

Converting this CNN architecture into an FCN architecture involves replacing the average pooling layer and onwards. We first keep the $1024 \times 8 \times 8$ representation, removing average pooling. Then, we replace the final fully-connected layer with a $1\times 1$ convolutional layer, also copying over the learned parameters. Instead of applying the fully-connected layer to a pooled representation of the input image, we now apply it to each of the $8 \times 8$ downsampled representation of the original $256 \times 256$ image, resulting in a coarse saliency map of output dimensions $2 \times 8 \times 8$. For an input image of a different size, for example $N \times M$, the output dimensions become $2 \times N // F \times M // F$, where $//$ indicates integer division.

%We first fix the kernel size of the average pooling layer to size 8x8, which maintains the $1024 \times 8 \times 8$ internal representation matching the downsampling factor of our trained model. Then, we replace the final fully-connected layer with a $1 \times 1$ convolutional layer, also copying over the learned parameters. Instead of applying the fully-connected layer to an pooled representation of our $N \times M$ input image, it is now applied to each of the $8 \times 8$ pooled pixels associated with each $256 \times 256$ contiguous block of our $N \times M$ input image, resulting in a coarse class prediction map with output dimensions $2 \times (N//F) \times (M//F)$. 

``Shift-and-stitch'' allows us to generate dense (pixelwise) predictions from these coarse outputs by shifting the inputs and interlacing the resulting output prediction maps. Due to the intermediate max-pooling layers, we can assign each coarse prediction to its $F \times F$ px receptive field. Then, by ``shifting'' the input horizontally and vertically by F pixels ($F^2-1$ total shifts) then ``stitching'' or interlacing the output accordingly, we produce dense classification map at the same resolution as the input image. The end result is identical to applying the model via sliding window to the set of $256x256$ contiguous tiles centered on each pixel in the image. However, due to hardware-based acceleration, this approach generates $N \times M$ native resolution outputs between two to three orders of magnitude faster than the brute force sliding window approach. 

% We follow this procedure only to convert our classifier model into a segmentation model without any retraining. We rely on a more modern architecture to also train a segmentation model from scratch, as described in Section \ref{sec:methods}.

%- CNN: 256x256 tile → 1024 x 256//F  x 256//F intermediate (F=8 → 1024 x 32 x 32) → Pool to 1024x1x1 → FC maps 1024x1x1 to 1x1 class probability
%- FCN: Shift+Stitch with (i x j)  $\in$ [0,F] $\ra$ Ri x Cj pad256 image at shift (i,j) $\ra$ 1024 x Ri//F x Ci//F intermediate → Pool to 1024 x F x F  → Ri//F x Ci//F salience  for offset (i,j)
%- Replace fully connected layers with 1x1 convolutions
%- Replace top-level pooling layer with

The \GNET architecture is a 22-layer, 4M parameter image classification model that utilizes blockwise layers of "Inception modules" which learn spatial features at varying scales \citep{szegedy_going_2015}. We trained our \GNET classifier from scratch, discarding the ImageNet-based pre-trained weights and 1024 class softmax layer, replacing it with a 2 class softmax layer which gives the probability that a given $D \times D$ tile contains a plume or not. We optimized the \GNET model for tilewise plume classification by converting the $D \times D$ plume label images ${\bf y}$ for each tile to scalar plume labels $y=\max({\bf y})$. Plume tiles with $y=1$ contain labeled plume pixels, while background tiles labeled $y=0$ lack any labeled plume pixels. We optimize the model using the binary cross entropy loss, defined in Equation \ref{eq:bce}.  

\begin{align} 
\text{L}_{bce}(p,y,w_+) &=-[ w_+ y \log(p) + (1-y) \log(1-p) ] \label{eq:bce}\\
w_{cls}&= n_{neg}\ /\ n_{pos} = n_{bg}/n_{plume} \label{eq:w_cls}
\end{align}

Here, $p$ is the scalar prediction produced by the image classification model that gives the probability of a given tile containing a plume.  We apply a positive class weight derived from the ratio of background versus plume tiles present in the training set (Eq. \ref{eq:w_cls}). This weight is crucial to account for the severe class imbalance between the relatively small quantity of identified plumes in each dataset versus the large and diverse set of background tiles sampled to cover each scene considered in our experiments. Without it, our training procedures were very slow to converge, as the rare discriminative updates including positive samples tended to be drowned out by the majority of weight updates based on purely negative examples that have little impact on improving discrimination between plume and background classes. 

For purposes of comparison, we also trained and evaluated the capabilities of the lightweight \METHANET model, which consists of 6 layers and 1.2M parameters, as proposed by \cite{jongaramrungruang_methanet_2022} using the same procedures as described above. 

\subsection{Pixelwise-Trained Model} \label{sub:pixelwise}
The pixelwise model we consider is a common architecture for image segmentation known as U-Net \citep{ronneberger_u-net_2015}. Similar to the FCN, the U-Net architecture leverages convolution and pooling layers to learn discriminative spatial features at varying spatial scales, and also uses $1 \times 1$ convolutions to permit inference on input images with arbitrary dimensions. However, unlike the FCN, U-Net uses de-convolution layers with skip connections between paired downsampling and upsampling layers to learn features directly from pixelwise labels that preserve fine spatial structures. The specific U-Net model we consider uses five convolutional blocks that double from 32 to 1024 channels, which allows the model to better learn spatial features for methane plumes that vary significantly in size. We use the Focal Loss  \citep{lin_focal_2017}, defined in Equation \ref{eq:focal}, to train the pixelwise model. The Focal Loss is a modified version of the BCE loss which focuses learning on difficult or ambiguous samples by penalizing mispredictions over correct predictions, which tends to increase optimization efficiency and prediction accuracy on imbalanced or rare classes. 

\begin{align} 
\text{L}_{focal}({\bf p},{\bf y},w_+) &= \frac{1}{D^2}\sum_{ij=1}^{D^2} \text{L}_{f}({\bf p}_{ij},{\bf y}_{ij},w_+) \label{eq:focal}\\
\text{L}_{f}(p,y,w_+) &= \alpha_f (1-e^{-p_+})^{\gamma_f}\ \text{L}_{bce}(p,y,w_+) \nonumber\\
p_+&=yp+(1-y)(1-p) \nonumber\\
\alpha_{f}&=0.25,\ \gamma_{f}=2,\ w_{seg}=1.25 \nonumber
\end{align}

Here, ${\bf y}$ is the $D \times D$ plume label image and ${\bf p}$ is the $D \times D$ image of pixelwise predictions generated by the U-net model for a given training/test tile. We assign a slightly higher weight to positive pixels by scaling positive pixel losses by a factor of $w_{seg}=1.25$. However, because the focal loss is equipped to handle imbalanced classes more effectively than the BCE, we did not observe a significant difference in segmentation results using positive loss weights $w_{seg} \in [1.0, 1.3]$. Larger values of $w_{seg}$ tended to negatively impact prediction accuracy. We did not investigate why larger positive pixel weights reduced pixelwise prediction accuracy, but we suspect that doing so forces the optimizer to over-emphasize generating accurate predictions on potentially ambiguous plume labels. 

\subsection{Multitask Model} \label{sub:multitask}

Our multitask model provides an effective compromise between the plume segmentation and instance detection capabilities of the pixelwise and the tilewise models, respectively. The model uses the same U-net architecture as the pixelwise model, but is trained using the loss function defined below which simultaneously accounts for pixelwise segmentation and tilewise classification objectives. 

\begin{equation} \label{eq:mtl}
\begin{split}
\text{L}_{MT}({\bf p},{\bf y},\alpha_{seg},w_{seg},w_{cls}) = \alpha_{seg}\ \text{L}_{focal}({\bf p},{\bf y},w_{seg})\\ + (1-\alpha_{seg})\ \text{L}_{bce}(\max({\bf p}),\max({\bf y}),w_{cls}) \\
\alpha_{seg}=0.5
\end{split}
\end{equation}

The multitask loss $\text{L}_{MT}$ combines the pixelwise focal loss $\text{L}_{focal}$ used to train the pixelwise model with the tilewise classification loss $\text{L}_{bce}$ used to optimize the tilewise classifier backend of our FCN model described in Section~\ref{sub:tilewise}. We optimize the tilewise classification loss between $\max({\bf p})$ and $\max({\bf y})$, where ${\bf y}$ and  ${\bf p}$ are the $D \times D$ pixelwise label and prediction images for a given tile as defined in \ref{sub:pixelwise}. The class label for a given tile $\max({\bf y})=1$ if the tile contains any labeled plume pixels, and is otherwise 0. This is similar to the multitask loss used by the Mask R-CNN model \cite{he_mask_2017}, which combines classification, bounding box and pixelwise losses to optimize a pixelwise model for object detection and instance segmentation objectives. Rather than optimizing separate classification and segmentation objectives as Mask R-CNN, we derive tilewise class predictions from pixelwise predictions by interpreting the maximum value of the pixelwise predictions $\max({\bf p})$ for a given tile as the predicted probability the tile contains a plume. So, optimizing our model to minimize the sum of the pixelwise focal loss (Equation~\ref{eq:focal}) and the tilewise loss $\text{L}_{bce}(\max({\bf y}),\max({\bf p}))$ drives the model to generate predictions $\in {\bf p} \rightarrow 0$ {\em for all pixels in background tiles containing no plumes} (i.e., $\max({\bf y})=0$). Consequently, training the model with the above multitask loss allows it to learn the pixelwise spatial characteristics of plume enhancements but also drives it to make more conservative predictions than the pixelwise model. 

\subsection{Measuring Instance-level Plume Detection Accuracy} \label{sub:instance}
While evaluating pixelwise segmentation metrics is standard practice and generally straightforward, measuring instance-level plume detection accuracy requires grouping pixelwise plume predictions into ``detection ROIs'' that each represent distinct detection candidates, and similarly, grouping pixelwise plume labels into ``label ROIs'' -- each representing a single plume candidate.

\begin{figure*}
 \centering
 \includegraphics[width=1.0\linewidth,clip,keepaspectratio]{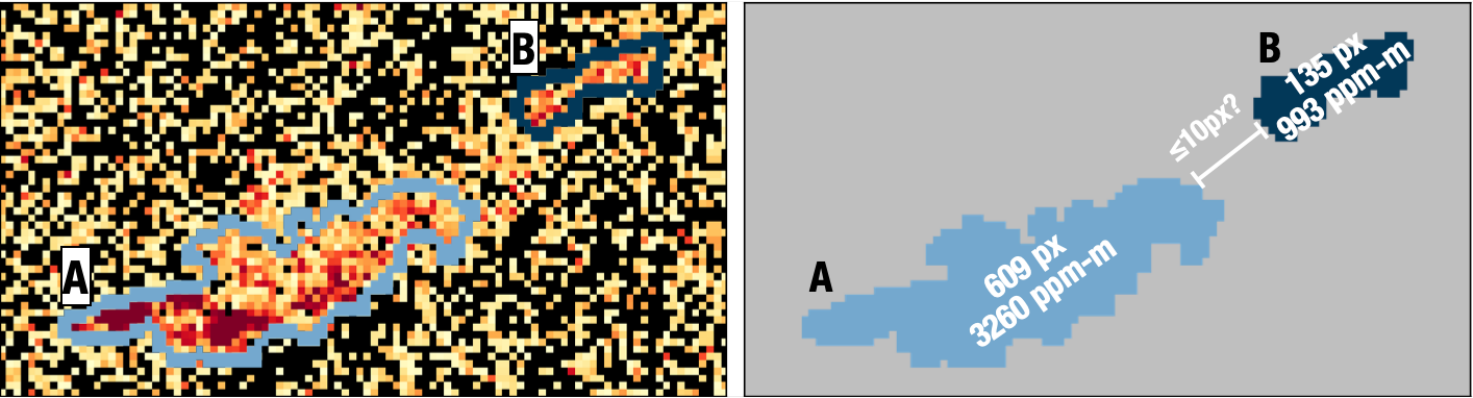} 
 \caption{\textbf{Left:} example `disconnected' plume consisting of two adjacent detection ROIs representing the plume body (A) and plume tail (B). \textbf{Right:} corresponding detection mask for each ROI. White text gives the pixel area and maximum enhancement per ROI. If A and B are within 10px of one another, we merge both ROIs into a single (multi-component) detection ROI. If not, A and B are considered distinct instance detections, and ROI B is flagged as ambiguous since its max enhancement $< 1000$ \ppmm. }
 \label{fig:instance_rois}
\end{figure*}

We use the conventions described below to group the connected components of a pixelwise plume label image and its corresponding pixelwise plume prediction mask into label and detection ROIs, respectively. A visualization of these conventions is shown in Figure \ref{fig:instance_rois}.

\begin{itemize}\addtolength{\itemsep}{-0.5\baselineskip}
\item Each label ROI consists of one or more adjacent connected components of the pixelwise labels in a given scene representing a single manually-identified plume candidate. 
\item Each detection ROI consists of the adjacent connected components of the pixelwise plume prediction mask occurring within 10 px of one another.
\item  Any (label or detection) ROI is ``ambiguous'' if it was flagged ambiguous during manual quality control, its area is smaller than $(10-1)^2=81$ px, or if its maximum enhancement is less than 1000 \ppmm. 
\end{itemize}

Common practice in object detection tasks is to require that the Intersection over Union (IoU) scores between label ROIs and detection ROIs exceed a fixed threshold $\tau_{\text{IoU}}$ to assign labeled ROIs as correct predictions (TP) or as mispredictions (FN or FP, for label or detection ROIs respectively). The IoU score is defined as,
\begin{equation}
\text{IoU}(L,D) = \text{area}(L \cap D) / \text{area}(L \cup D) 
\end{equation}

where $L$ and $D$ are the 2D pixel coordinates of label ROI $L$ and detection ROI $D$, respectively.

Given the challenges of the plume detection task (driven by the ambiguity of plume boundary labels, the variety of plume areas produced by emitters from distinct source sectors, the variability in plume versus background contrast, and spatial resolution across flightlines andcampaigns), we do {\em not} filter ROIs with respect to their IoU scores ($\tau_{IoU}=0$), instead using simple ROI overlap to evaluate instance-level accuracy. In other words, instance-level mispredictions consist of the set of detection ROIs that are spatially disjoint from all label ROIs (FP), and the set of label ROIs that are spatially disjoint from all detection ROIs (FN). The remaining set of label ROIs that spatially overlap at least one detection ROIs (i.e., IoU $> 0$) counted as correct instance predictions (TP). We handle many-to-one mappings between adjacent label and detection ROIs as follows: we count multiple adjacent detection ROIs overlapping a single label ROI as a single TP, while we count each adjacent label ROI that overlaps a single detection ROI as a single TP. 

Measuring both instance detection and pixelwise segmentation metrics accounts for pathological issues that may occur with unfiltered instance detections ($\tau_{IoU}=0$). For instance, a single detection ROIs covering an entire scene would produce high instance level accuracy since all label ROIs would be counted as TP, and the single detection would not be counted as a FP since it was not spatially disjoint from the labeled ROI in the scene. However, such cases will also yield low pixelwise accuracy---a very high false positive rate in the aforementioned case. 

\section{Case Studies} \label{apx:studies}

\subsection{Repeat Measurements and Persistent Enhancements} \label{apx:bias_persist}

Airborne GHG monitoring campaigns commonly collect repeated measurements of the same site to characterize the persistent emission sources. Because the locations of GHG emission sources and the characteristics of background substrate rarely change within the timespan of a typical airborne GHG imaging campaign, repeat measurements of the same site often contain enhancements with similar spatial characteristics that persist across distinct observations.

\begin{figure*}[htb]
 \centering
 \includegraphics[width=1.0\linewidth,clip,keepaspectratio]{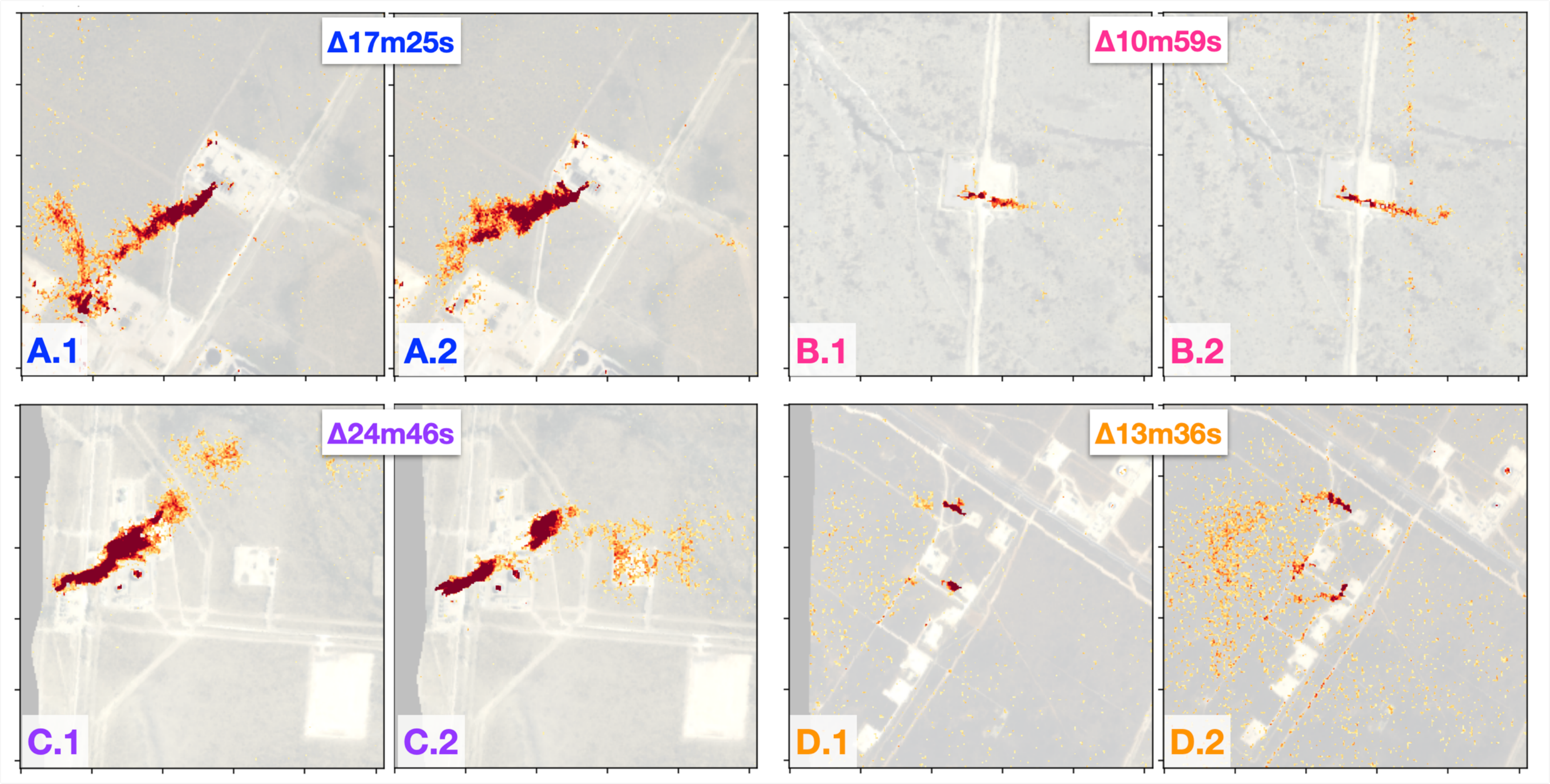} 
 \caption{Examples of spatially correlated plume enhancements from point sources in pairs of repeat overflights of four sites observed in the 2019 \ANG Permian campaign. Colored text indicates distinct sites and gives the time delta between each pair of observations.}
 \label{fig:plumecor}
\end{figure*}

Figure~\ref{fig:plumecor} shows several example plumes observed in repeat measurements of persistent emissions sources. While plume morphology varies by observation, high-concentration pixels typically occur near the location of the emission source. The neighboring background enhancements associated with substrate materials in the vicinity of the observed plumes are also spatially correlated across repeat measurements (e.g., bright rooftops and roads).

\begin{figure*}[htb]
 \centering
 \includegraphics[width=1.0\linewidth,clip,keepaspectratio]{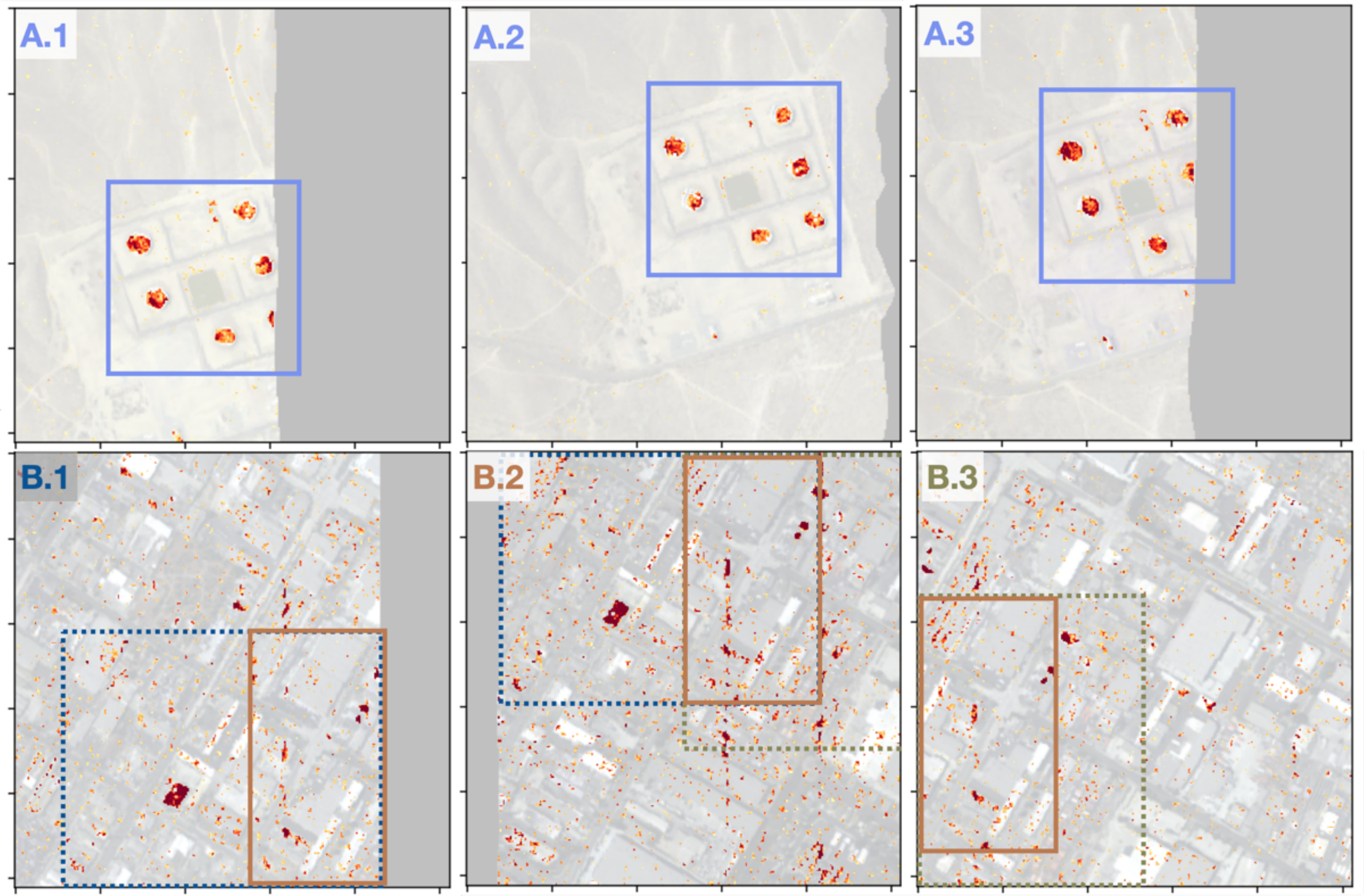} 
 \caption{Examples of persistent false enhancements observed in three repeat observations (columns) \ANG overflights of two sites (rows). Observations at site A (top row) contain false enhancements from gas storage tanks observed in the 2019 \ANG Permian campaign. Observations from site B (bottom row) contain false enhancements from bright rooftop materials from the 2020 \ANG COVID campaign along with reocurring columnwise artifacts. Blue boxes show the region of common overlap shared among all three observations of each site, while cyan and magenta boxes show the overlapping regions common to observations (B.1,B.2) and observations (B.2,B.3), respectively.}
 \label{fig:falsecor}
\end{figure*}

Figure~\ref{fig:falsecor} shows repeat observations of two sites observed in the \ANG Permian and COVID campaigns, each containing spatially correlated persistent false enhancements from surface confuser materials (e.g., gas storage tanks and bright rooftop materials) and image artifacts (reoccurring columnwise artifacts). 

\begin{figure*}
 \centering
 \includegraphics[width=0.8\linewidth,clip,keepaspectratio]{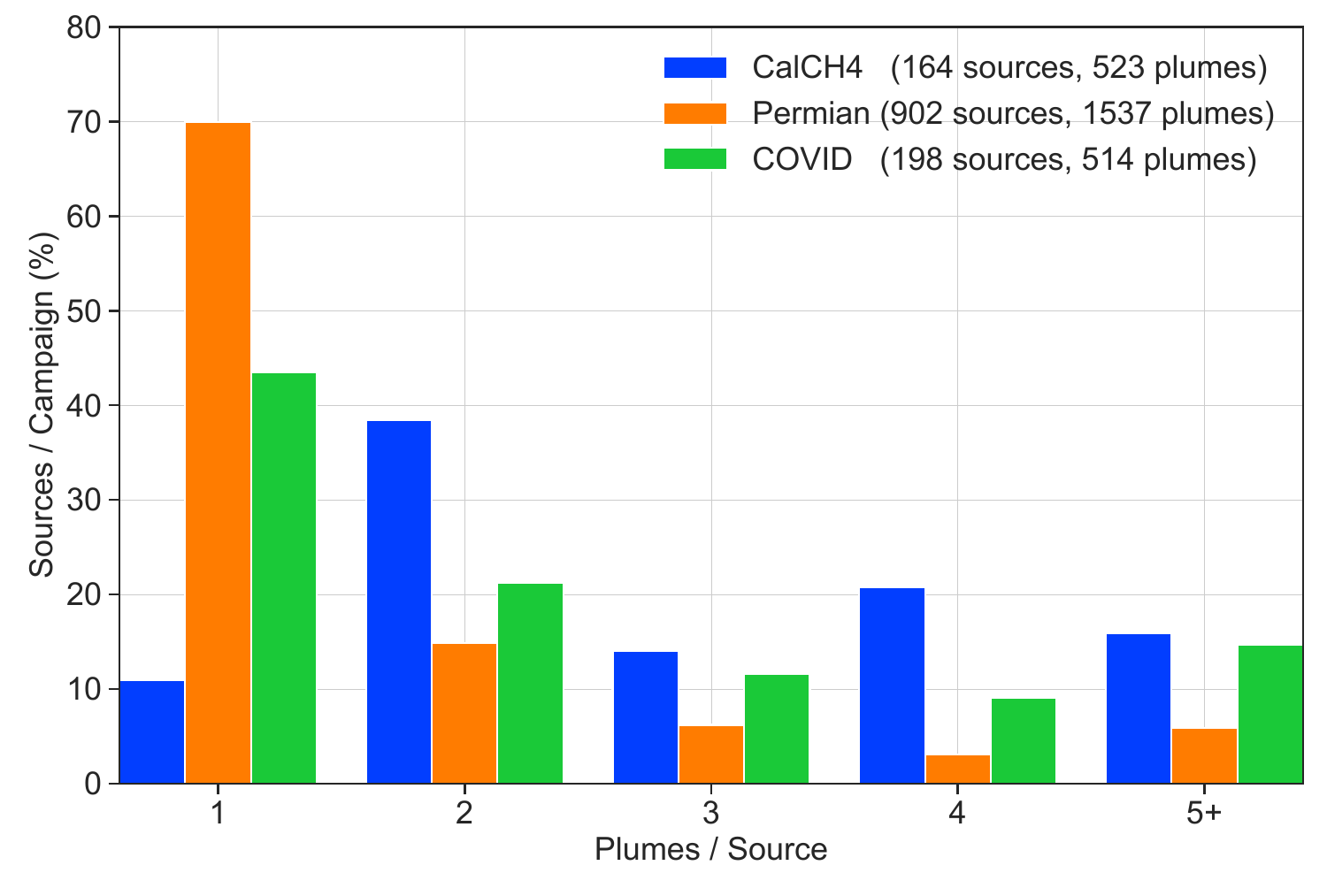} 
 \caption{Percentage of sources per campaign (y-axis) versus the number of plumes associated with repeat measurements of each source (x-axis) in 2018-2020 \ANG campaigns. Sources represented by a single plume are statistically independent, while plumes from repeat measurements are autocorrelated with other plumes representing the same source.}
 \label{fig:angrevisit}
\end{figure*}

The majority of plume candidates identified in the airborne GHG imaging campaigns thus far represent plumes observed in repeat measurements of a subset of sources targeted in each campaign. 

Figure~\ref{fig:angrevisit} gives the number of plume candidates representing one or more observations of the same emission source for \ANG campaigns from 2018-2020. Overall, 69\% ($(523-164)/523$) of the \CALMETHANE plume candidates, 41\% (635) of the Permian plume candidates, and 62\% (316) of the COVID plume candidates represent one or more repeat measurements of distinct emission sources. Only 18 of the 164 sources (11\%) observed in the 2018 \CALMETHANE campaign were measured once, which means only 18 of the 523 plumes (3\%) are independent samples representing unique sources, while the remaining 505 candidates (97\%) represent the remaining 146 sources (89\%) with at least two plumes observed for each source. Notably, a third (174/523) of the 2018 \CALMETHANE plume candidates represent five or more repeat measurements of only 16\% (26) of the sources observed in that campaign. In contrast, 631 of the 902 Permian sources (70\%) were measured once. The corresponding 631 plume candidates (41\%) constitute the independent observations of unique sources from that campaign, while the remaining 906 plume candidates (59\%) represent plumes observed in repeat measurements of the remaining 271 sources (30\%). 

\begin{figure*}
 \centering \includegraphics[width=0.8\linewidth,clip,keepaspectratio]{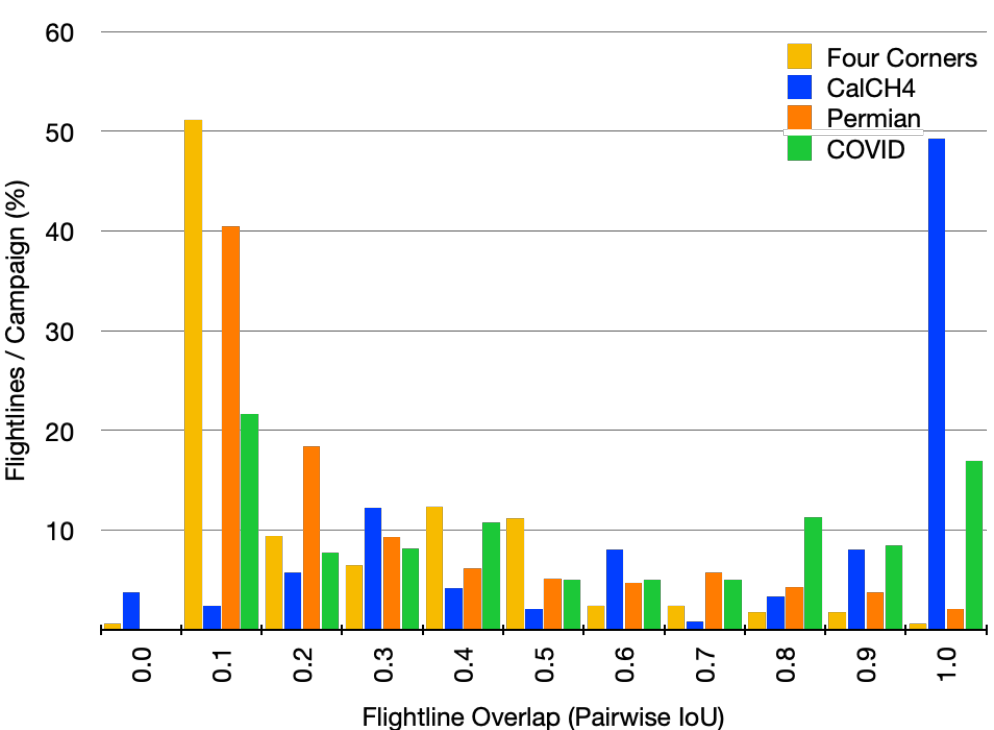} 
 \caption{Distribution of pairwise flightline IoU scores for selected \ANG campaigns. An IoU score of 0.0 indicates the bounding boxes of two  flightlines do not overlap, while an IoU score of 1.0 indicates two flightlines share an identical bounding box.}
 \label{fig:angoverlap}
\end{figure*}

The proportion of overlapping scenes captured in each campaign and the overall extents of their overlapping regions may lead to biased evaluation of plume detection results. Here, background samples are our primary concern, for two reasons. First, overlapping background tiles are not independent, and background tiles sampled from highly overlapping flightlines will inevitably overlap; especially so in this work since our sampling procedure aims to cover as many non-plume pixels in each scene as possible. Second, since overlapping tiles often contain enhancements with common spatial patterns, training a plume detector with background samples from campaigns consisting of many highly overlapping flightlines increases the risk of overfitting, because those commonalities may be misconstrued as discriminative features of the background class.

Figure~\ref{fig:angoverlap} shows the distribution of pairwise flightline overlap from the 2018-2020 \ANG campaigns, measured by calculating the IoU scores of flightlines' bounding boxes. The largest proportion of overlapping flightlines occur in the 2018 \CALMETHANE campaign, with nearly 50\% of the flightlines sharing nearly identical bounding boxes (IoU $\in [0.9,1.0]$). In contrast, the flightlines from the \FOURC campaign overlap the least, with roughly half of the flightlines reaching IoU scores $\in [0.1,0.2]$.

\subsection{Impact of Spatiotemporal Bias} \label{apx:bias_stratify}
Spatial stratification is crucial to ensure accuracy metrics provide realistic estimates of the expected performance of a model when applied to data similar to the data supplied during training. By splitting data into spatially disjoint training and test sets, spatiotemporal sample stratification prevents data leakage that occurs when spatially-informed models memorize site-specific persistent enhancements captured in repeat overflights of the same area.

\begin{table}
 \centering \includegraphics[width=1.0\linewidth,clip,keepaspectratio]{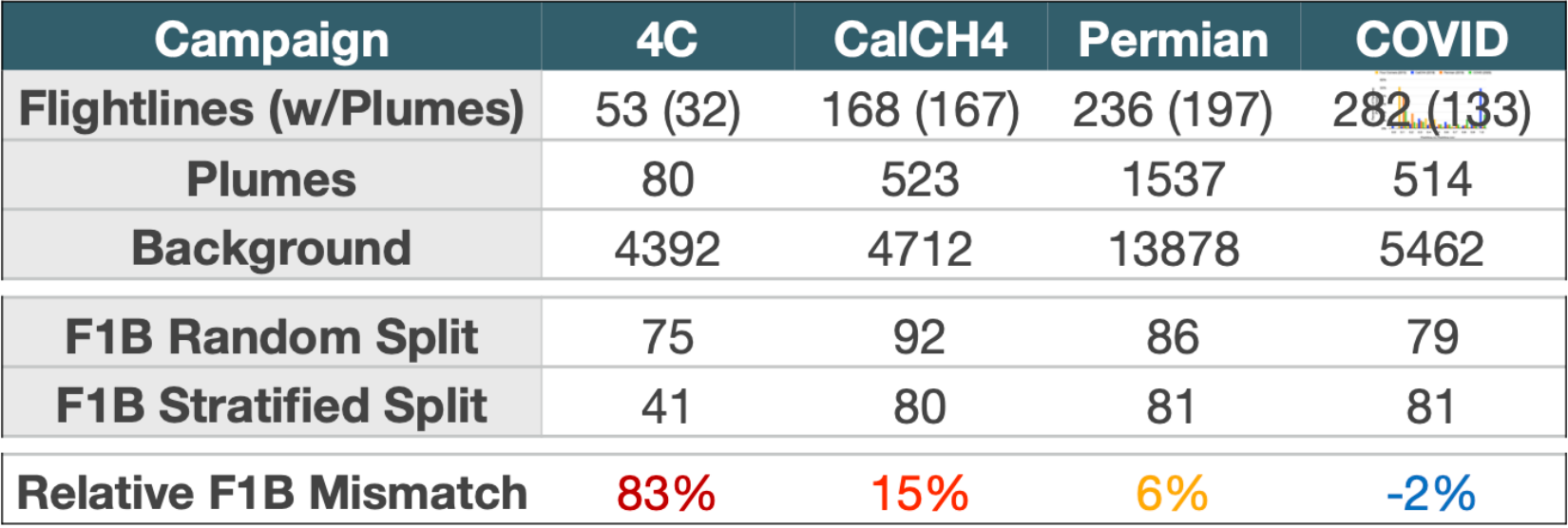} 
 \caption{Tilewise test \FB using \GNET model trained/tested with random vs. spatially stratified train/test partitions.}
 \label{tab:stratify_random}
\end{table}

To assess the impact of spatiotemporal bias on plume detecion results, we compared the test \FB scores produced by single-campaign \GNET models trained \& tested using tiles partitioned using the scenewise spatial stratification approach described in Section~3 versus tiles partitioned via stratified random sampling. We stress that same set of tiles and campaign were used in both experiments; only the training and test partitions differ. In the random split case, we used class-stratified sampling to ensure the number of plume and background tiles in the training \& test sets roughly matched the proportion of plume and background tiles in the stratified sampling case. We use a 75/25\% train/test split for each campaign.

Table~\ref{tab:stratify_random} gives the test \FB scores of the tilewise \GNET model with respect to randomly-partitioned training \& test tiles versus spatially-stratified training \& test tiles. The impact of spatiotemporal bias is most observable in the difference between random vs. spatially stratified prediction accuracies when the training set is small (i.e., \FOURC campaign) and/or contains many spatially overlapping samples from a comparably small number of sites (i.e., \CALMETHANE campaign).

In the Four Corners dataset, in which only 32 of the 53 flightlines contain a total of 80 plumes, the test \FB-score of the model trained and tested on the random split (0.75) is highly inflated compared to the test \FB-score of the model trained and tested on a stratified split (0.41). The performance on the random split dataset suggests the model is highly overfit, while performance on the stratified split dataset suggests that the 4C data does not contain enough discriminative information to adequately capture distinctions between plume and background classes. In the \CALMETHANE dataset, over half of the flightlines are repeat observations with high spatial correlation that significantly overlap (IoU scores $\in [0.9,1.0]$. The model performs fairly well with stratification (\FB=80\%), which suggests the data from the campaign is of adequate sample size and diversity to train a plume detector. However, without stratification, we observe an optimistic 12\% increase in \FB score (=92\%) since nearly half of the flightlines in the training set overlap test set by more than 90\%, allowing the model to ``cheat'' on the autocorrelated tiles in test set. However, the effect is not as dramatic in this case as with the \FOURC campaign, likely due to the significant increase in plume samples (80 vs. 523).
% \begin{itemize}\addtolength{\itemsep}{-0.5\baselineskip}
% \item Four Corners (small sample size) only 32 of the 53 flightlines in the campaign containing a total of 80 plumes, the random model is highly overfit, while the stratified model suggests that the 4C data does not contain enough discriminative information to adequately capture distinctions between plume vs. background classes.
% \item \CALMETHANE (low sample diversity):  over half of the flightlines in this campaign have IoU scores $\in$ [0.9,1.0], and these repeat observations are highly (spatially) correlated. The model performs fairly well with stratification (\FB=80\%), which suggests the data from the campaign is of adequate sample size and diversity to train a plume detector. However, without stratification, we observe an optimistic 12\% increase in \FB score (=92\%) since nearly half of the flightlines in the training set overlap test set by more than 90\%, allowing the model to ``cheat'' on the autocorrelated tiles in test set. However, the effect is not as dramatic in this case as with the \FOURC campaign, likely due to the significant increase in plume samples (80 vs. 523).
% \end{itemize}

\subsection{Impact of Regional Bias} \label{apx:bias_region}
Regional spatiotemporal bias occurs when a model learns geospatial characteristics from observations of a fixed region (e.g., data captured in a single airborne campaign) that are not informative for other similar regions. Controlling for regional bias is difficult when train/test data are from a single region, so single campaign datasets are generally inadequate to measure generalization performance. 
\begin{table*}
 \centering
 \includegraphics[width=1.0\linewidth,clip,keepaspectratio]{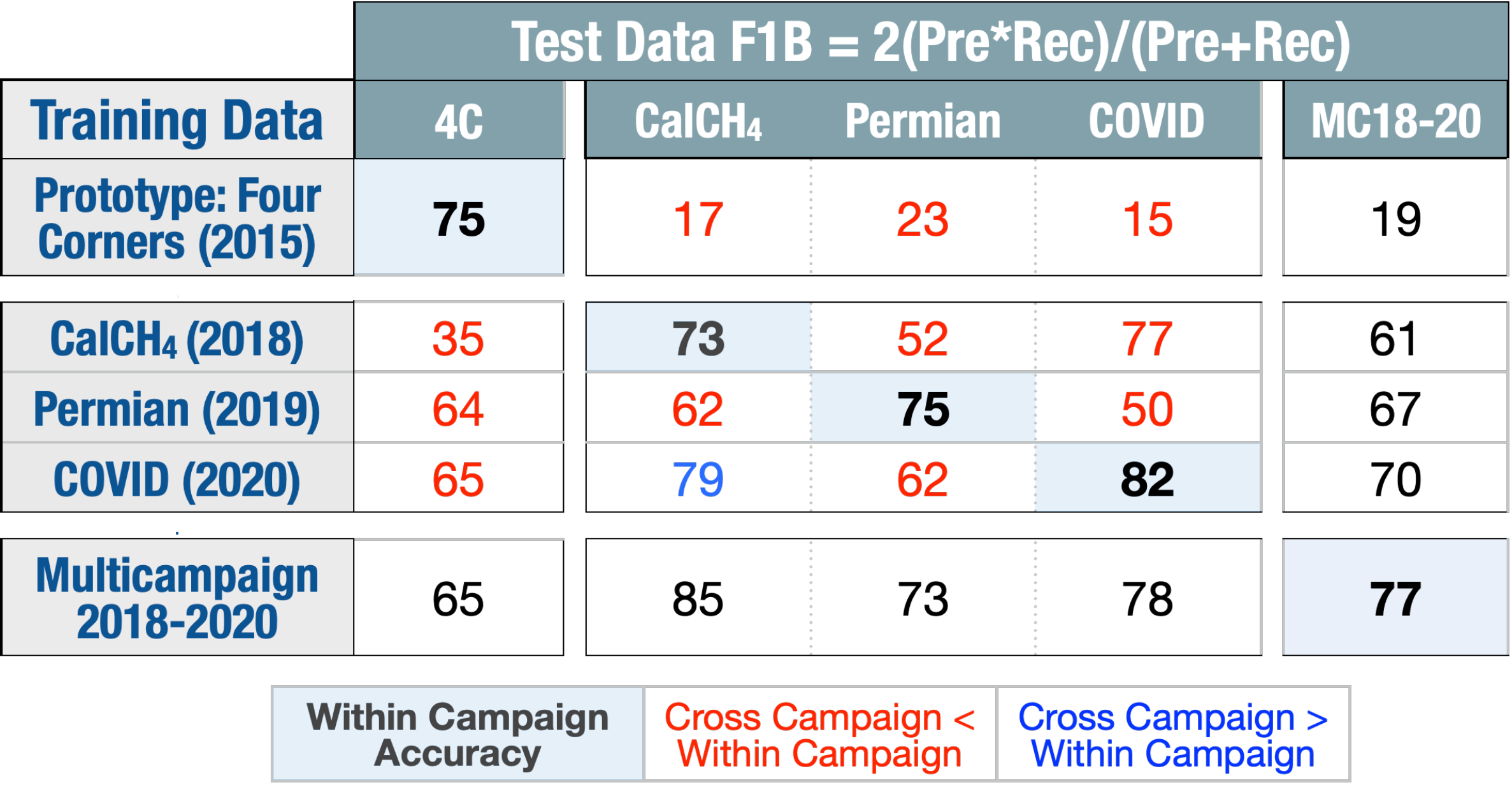} 
 \caption{Test \FB of single campaign models vs. multicampaign models applied to “within campaign” test samples from the same campaign(s) versus “cross campaign” test samples from campaign(s) outside of training set.}
 \label{tab:singlevsmulti}
\end{table*}

Table~\ref{tab:singlevsmulti} gives the test \FB scores of ``single campaign'' models trained on the Four Corners, \CALMETHANE, Permian, and COVID campaign data sets versus the multicampaign model trained on the latter three campaigns. With the exception of the Four Corners model, spatially stratified samples selected using the procedure described in Section~3 were used to train each model. The last row gives the test scores achieved by the multicampaign model applied to data from each campaign, and the last column gives the scores of the single campaign models applied to the multicampaign test set. Otherwise, values on the diagonal give the ``intra-campaign'' scores where each model was trained and tested on data from the same campaign, off-diagonal values give the ``inter-campaign'' scores where each model was tested on data outside its training set.

\subsection{Regional Bias \& Plume Identifiability in the GAO Penn Campaign} \label{apx:gao_penn}
\begin{figure*}
 \centering
 \includegraphics[width=1.0\linewidth,clip,keepaspectratio]{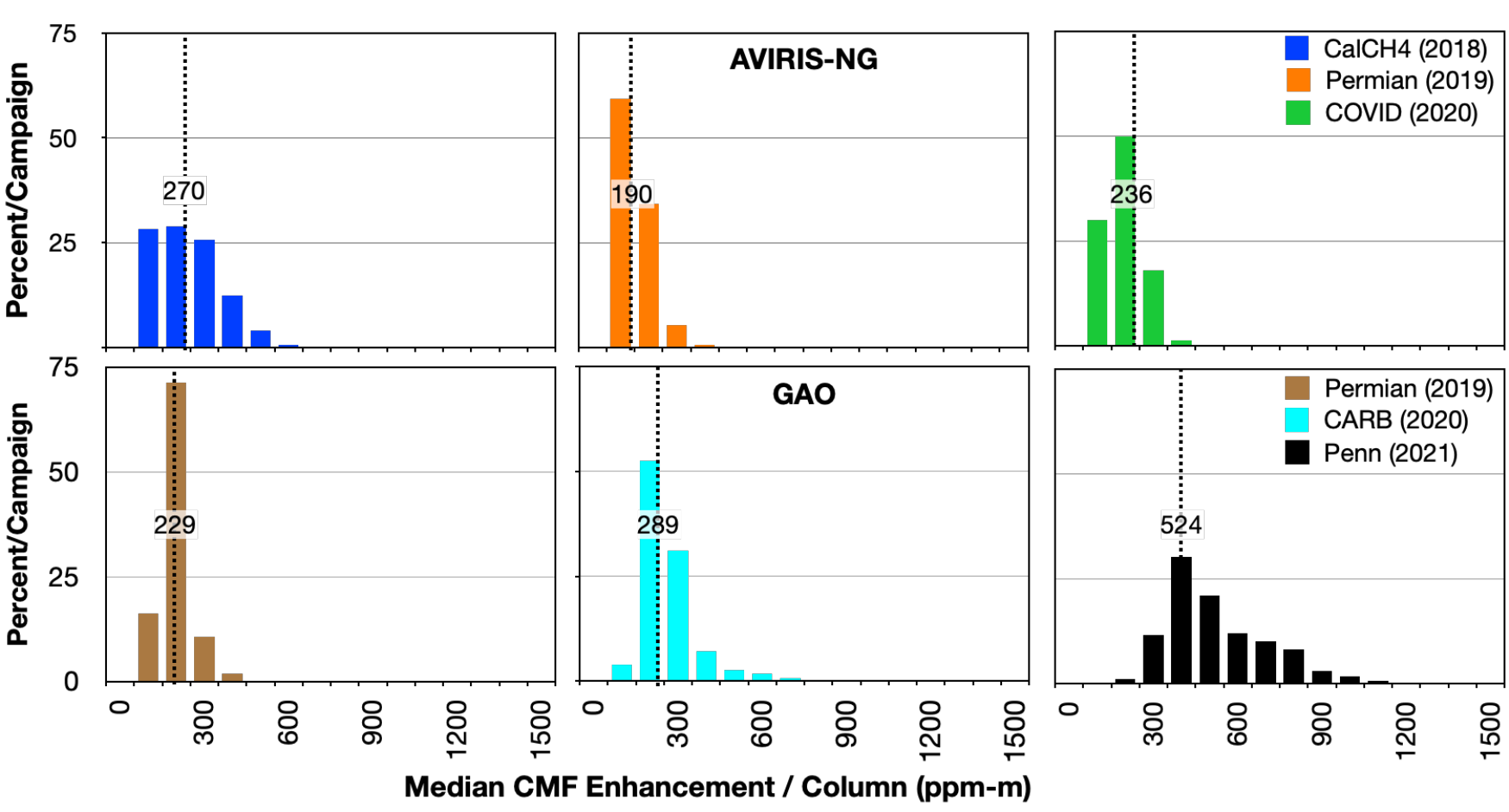} 
 \caption{Distribution of median CMF \methane column enhancements / flightline / campaign.}
 \label{fig:bgedist}
\end{figure*}

The GAO Penn campaign collected the first airborne \methane imaging spectrometer products targeting emission sources associated with coal mining infrastructure in the densely forested, humid continental climate of southern and eastern Pennsylvania. As shown in Figure \ref{fig:bgedist}, typical CMF background enhancement levels in these dark, heavily vegetated regions are much higher, ranging between 500-700 \ppmm on average per-scene, than the typical background enhancements in the 200-300 \ppmm range observed in prior airborne \methane imaging campaigns targeting emission sources in the desert, semi-arid, or Mediterranean climate zones of the southwestern United States. 

\begin{figure*}
 \centering \includegraphics[width=0.9\linewidth,clip,keepaspectratio]{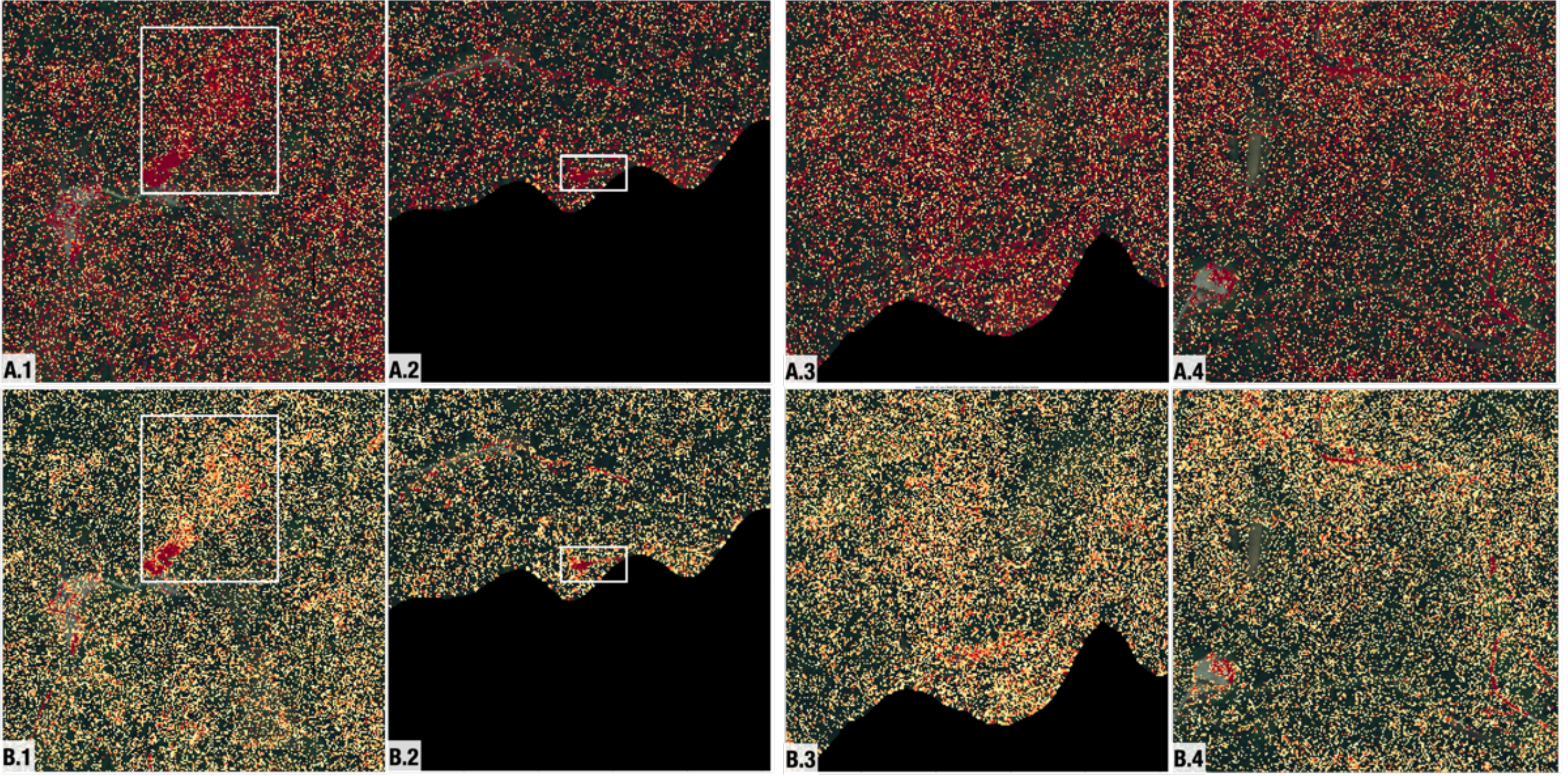} 
\caption{Example tiles from the 2021 \GAO Pennsylvania imaging campaign clipped to the [500,1500]\ppmm range (top row) versus the [500,4000]\ppmm range. Plumes (columns 1 and 2) are visibly distinct when viewed in the [500,4000]\ppmm range, but are difficult to visually identify using the [500,1500]\ppmm range generally sufficient to identify plumes in CMF images captured in other airborne campaigns. Background tiles (columns 3 and 4) are often falsely detected as plumes due to their high background enhancement levels, but lack observable structure when viewed with broad dynamic range. White boxes indicate the approximate bounding regions of \methane plumes produced by emission sources from coal mining infrastructure.}
 \label{fig:penn_tiles}
\end{figure*}

The uncharacteristically high background enhancement levels observed in the GAO Penn campaign are due to the presence of numerous large false enhancements that collectively cover the majority of pixels in most scenes from the campaign. The presence of these false enhancements explained why we observed the highest FDR of all the airborne campaigns we consider in this work on the GAO Penn data, as these highly concentrated "background" regions  are often misclassified as plumes. It also revealed that scientifically valid, but nonstandard plume identification procedures were applied to identify plumes in these scenes. Specifically, because domain experts could not effectively identify plumes generated by real \methane sources in many of the GAO Penn scenes, experts first located areas where candidate infrastructure elements were visible in fine-resolution basemap images, and then used those candidate source locations to guide plume identification using the associated CMF products that were clipped to a enhancement range broader than the range typically sufficient in other campaigns. Figure~\ref{fig:penn_tiles} show characteristic plume and background tiles from the GAO Penn campaign clipped to the standard [500,1500]\ppmm range sufficient to identify plumes in prior airborne campaigns, versus the same tiles clipped to the nonstandard [500,4000]\ppmm range where the GAO Penn plumes are easier to visually identify. While these nonstandard procedures are scientifically valid and perfectly acceptable, they illustrate human capabilities to dynamically account for statistical issues that a ML-driven plume detector cannot. 

\subsection{Impact of Sampling Bias} \label{apx:bias_sample}
Sampling bias occurs when either the training or test set are not representative samples of the joint data and class distribution. With respect to GHG plume detection, a common form of sampling bias occurs when the set of background tiles does not capture the diversity of enhancements present in a given scene. When the background class tiles used to train and validate a plume detector are undersampled, two issues may occur:

\begin{enumerate}[I]
\item A model trained on a biased sample will generalize poorly on a representative sample of test data. 
\item Accuracy metrics computed on a biased sample of test background tiles are similarly biased, and typically overestimate performance.
\end{enumerate}

\begin{figure*}
 \centering
 \includegraphics[width=1.0\linewidth,clip,keepaspectratio]{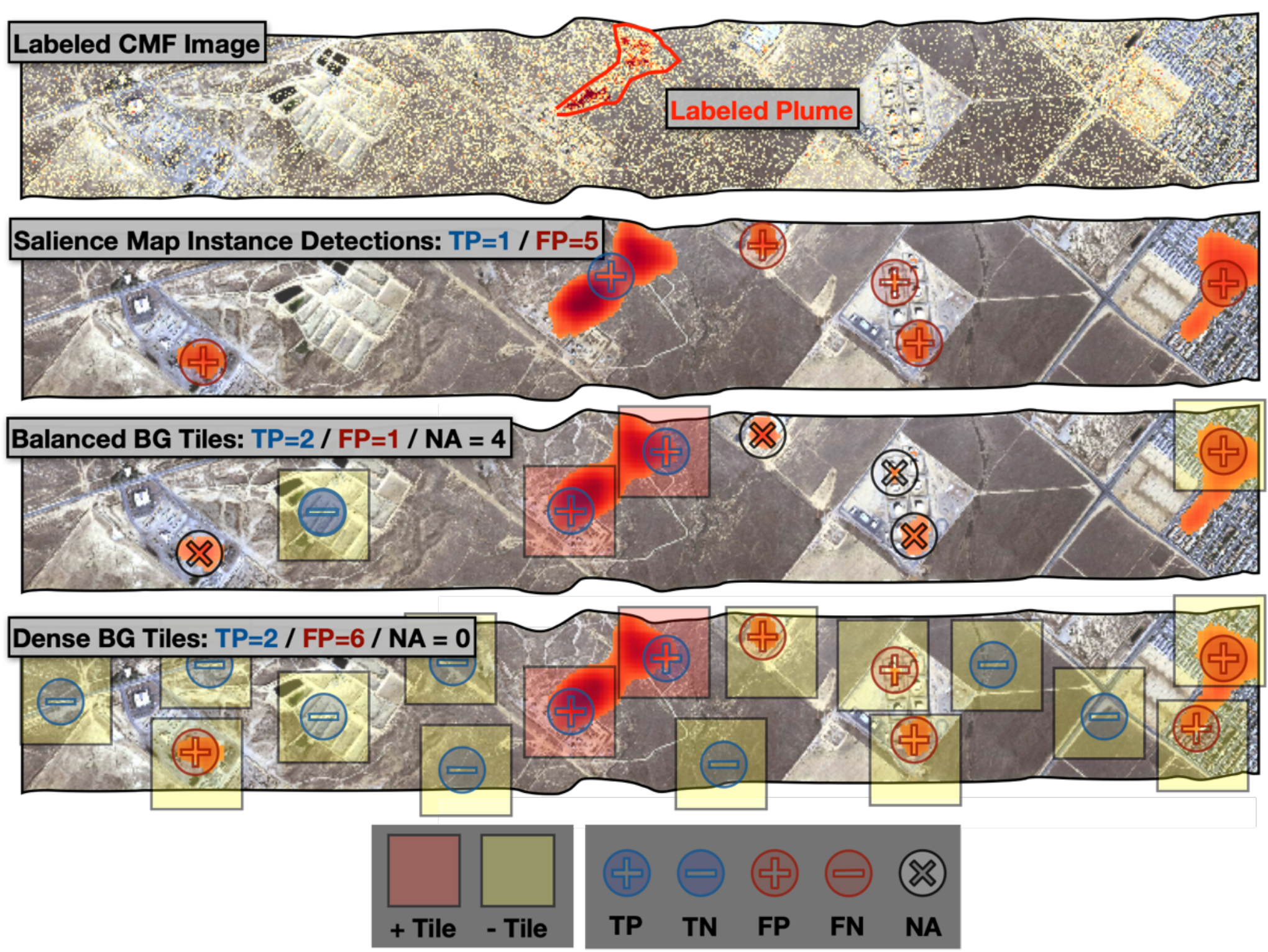} 
 \caption{An illustration of the impact of sampling bias on model performance metrics. \textbf{First row:} A single labeled plume is present in this example scene. \textbf{Second row:} A salience map output from a plume detection model shows one true positive instance detection and five false positive detections due to false enhancements. \textbf{Third row:} In the balanced sampling case, the same number of plume tiles and background tiles are sampled. Because only two background tiles are sampled, many false positive detections are not captured in the accuracy metrics. \textbf{Last row:} In the dense sampling case, as many background tiles as possible are sampled from the scene. Every false positive detection is captured, and the accuracy metrics more accurately reflect realistic, operational performance.}
 \label{fig:instance_tilebias}
\end{figure*}

Figure~\ref{fig:instance_tilebias} illustrates issue II in detail. In the case where the same number of plume tiles and background tiles are sampled, several false predictions are not captured, leading to an optimistic performance. Dense background tile sampling is critical for accurate performance of model performance.

\begin{figure}
 \centering
 \includegraphics[width=1.0\linewidth,clip,keepaspectratio]{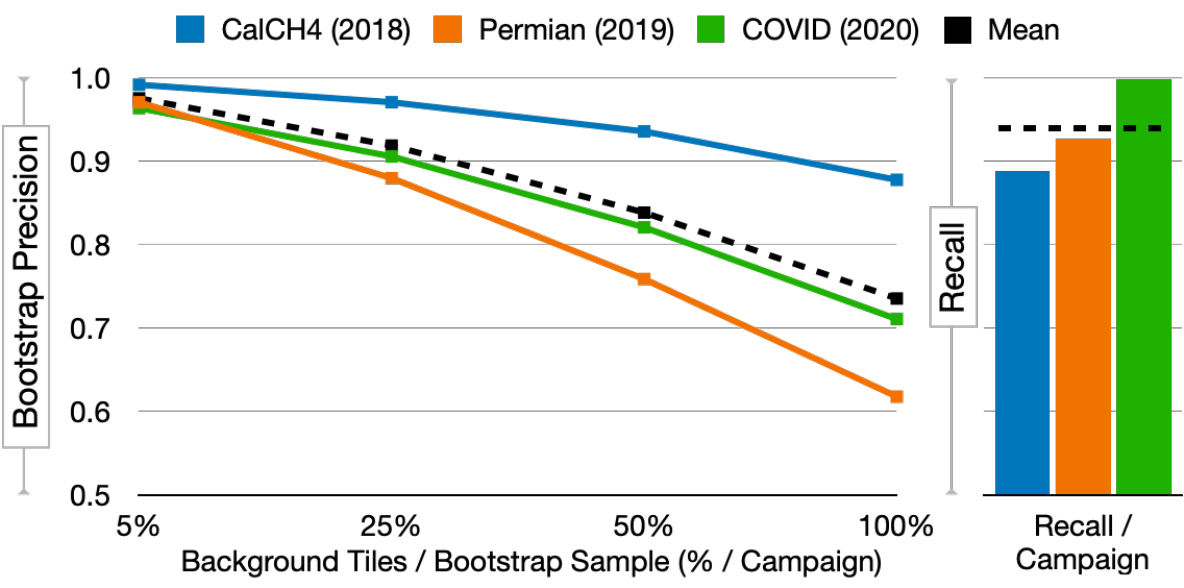} 
 \caption{Left plot: Average precision scores of the \ANG multicampaign model predictions with respect to five bootstrap samples of sizes ranging of 5\% to 100\% of the background test tiles in the \ANG multicampaign data set. Right plot: Corresponding recall scores / campaign. Recall scores are constant as plume tiles are held fixed in each experiment.}
 \label{fig:bootstrap_pre_rec}
\end{figure}

\subsection{Impact of Train/Test Distribution Mismatch} \label{apx:bias_distrib}
In this section, we consider the problem of distribution mismatch, and show that models trained using only \LES-simulated plumes generalize poorly to real plumes observed by the \ANG spectrometer. Our results suggest that the \LES plumes do not adequately capture the diversity of real plumes observed in observational data.

\begin{figure*}
 \centering
 \includegraphics[width=1.0\linewidth,clip,keepaspectratio]{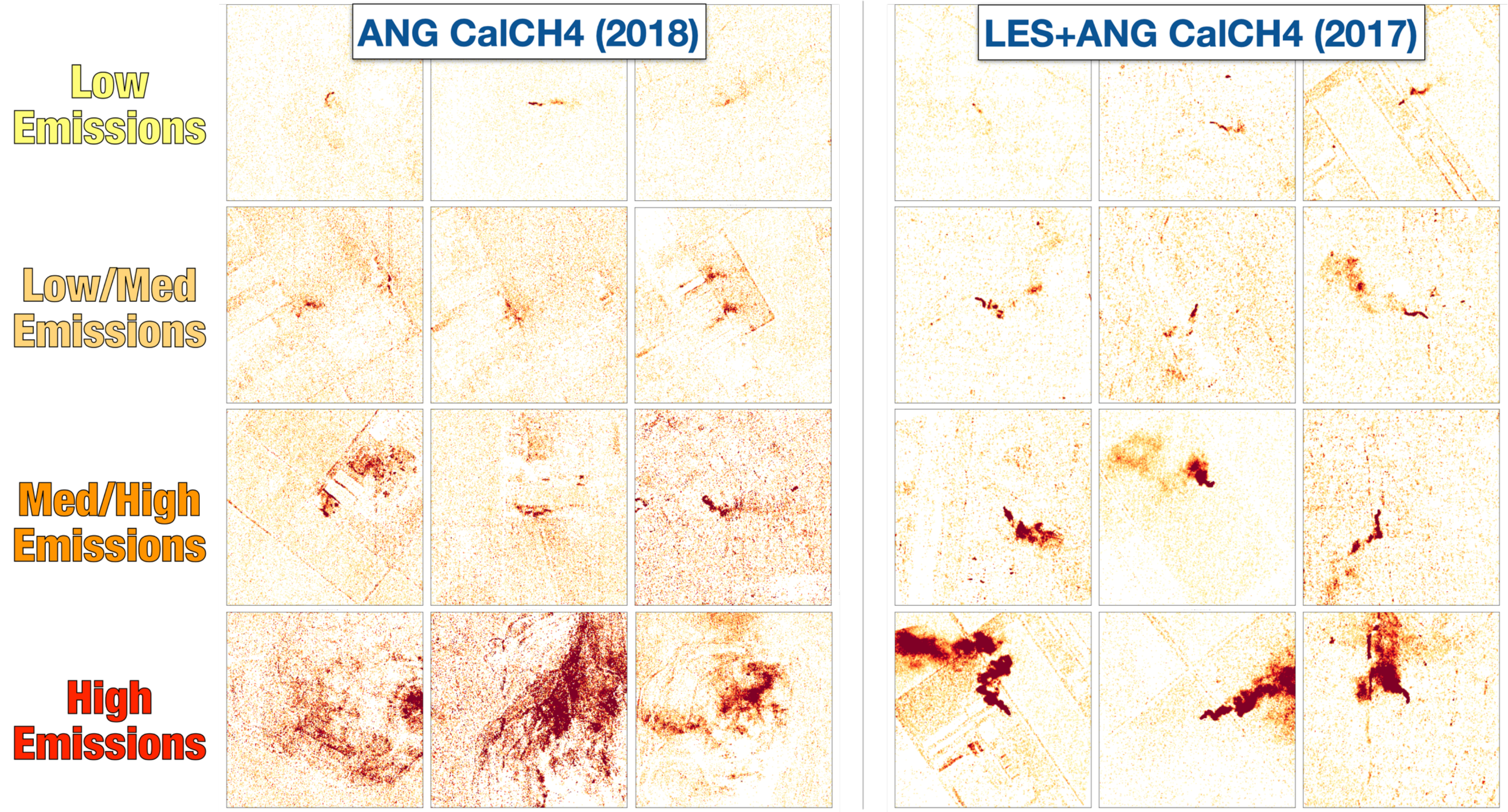} 
 \caption{Example plumes from \CALMETHANE (2018) campaign (left) versus \LES plumes inserted into \CMF background tiles from \CALMETHANE (2017) campaign from \cite{jongaramrungruang_methanet_2022}.}
 \label{fig:ang_vs_les_plumes}
\end{figure*}

\begin{figure*}
 \centering \includegraphics[width=1.0\linewidth,clip,keepaspectratio]{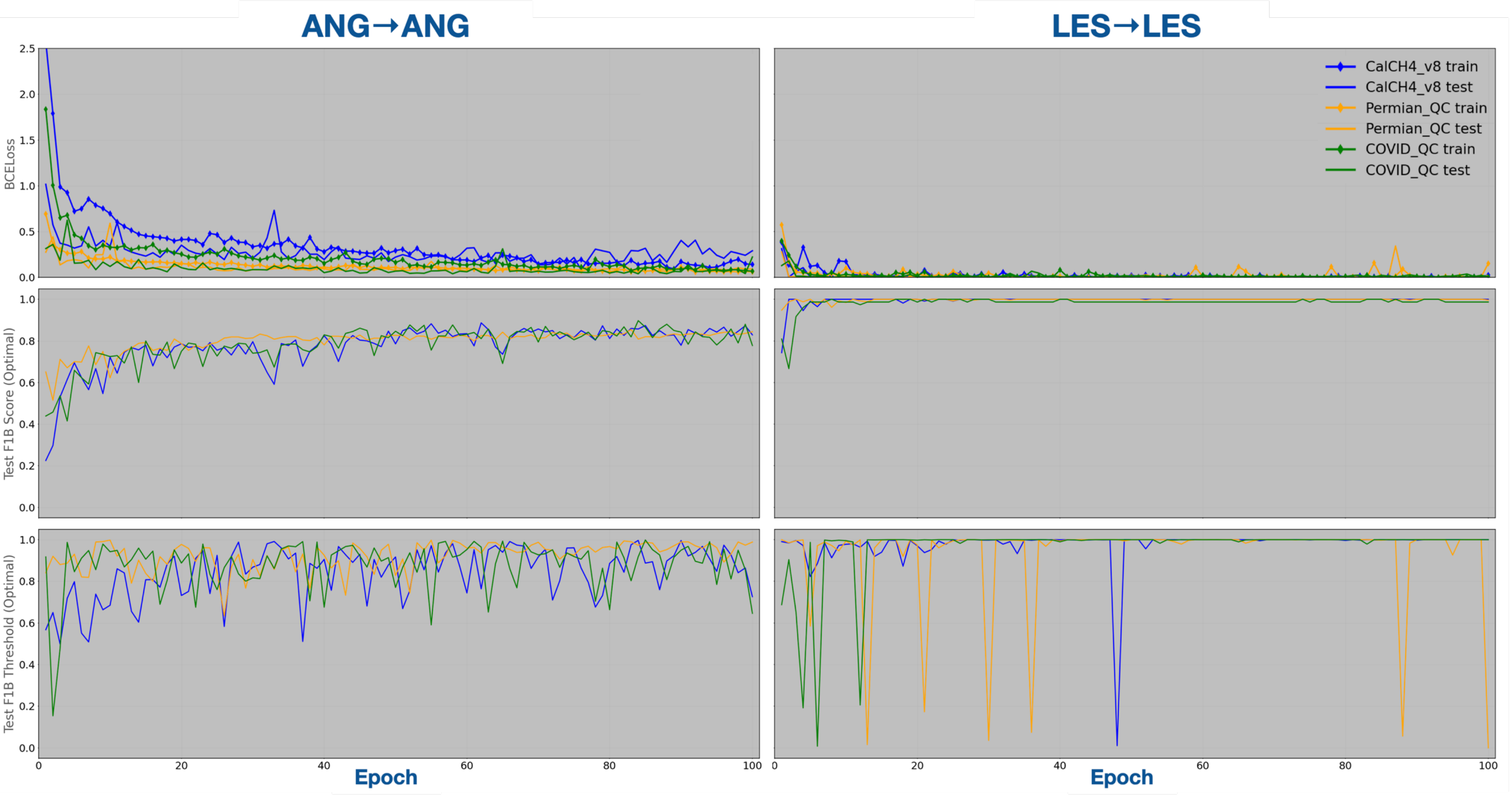} 
 \caption{Tilewise plume classification results with train/test data from the same domain. Left column: \ANG training and testing data. Right column: LES training and testing data. Top row: Training loss/epoch. Middle row: Optimal test \FB score/epoch. Bottom row: Salience threshold maximizing test \FB/epoch. }
 \label{fig:ang_vs_les_intra}
\end{figure*}

\begin{figure*}
 \centering \includegraphics[width=1.0\linewidth,clip,keepaspectratio]{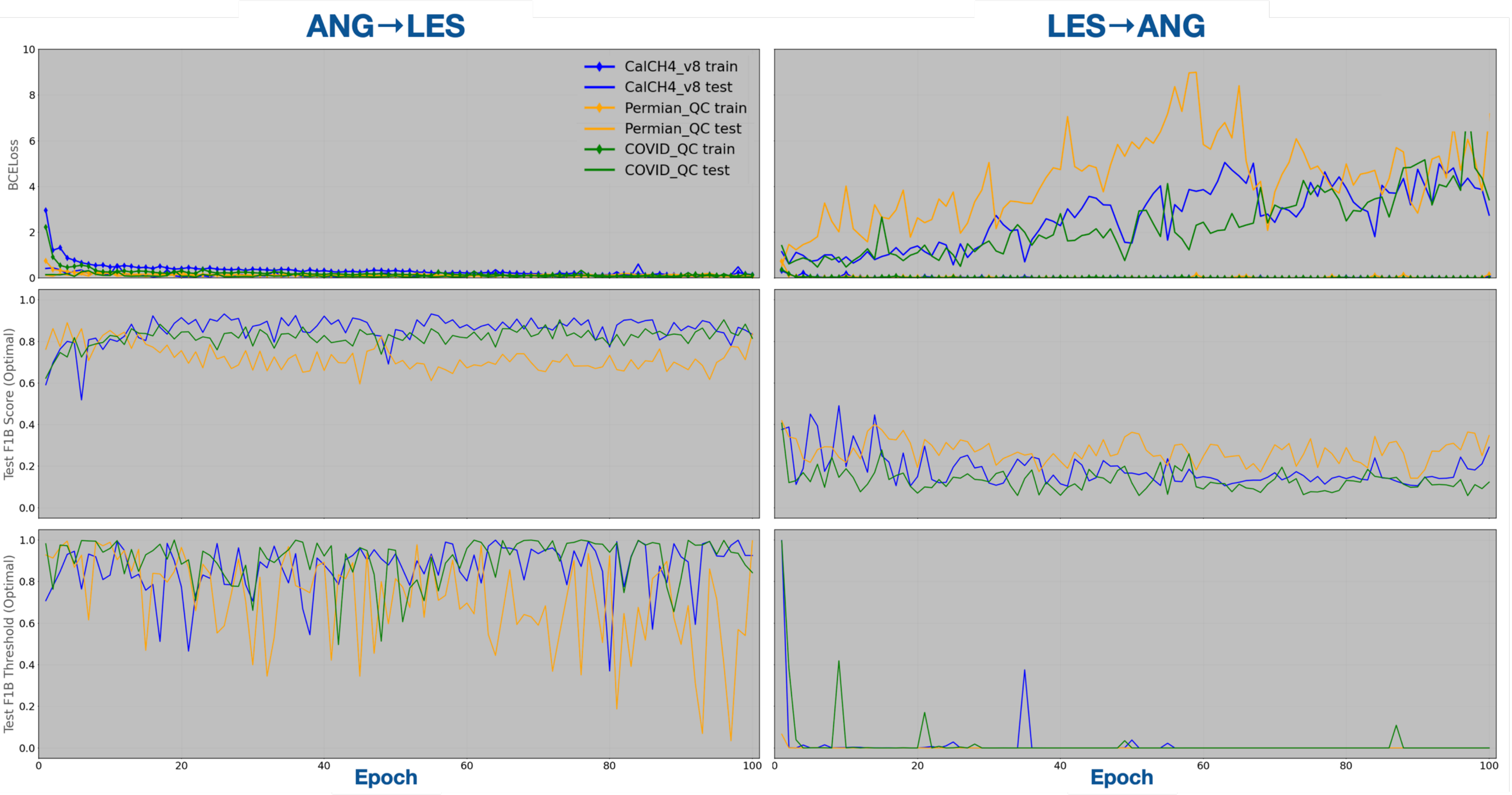} 
 \caption{Tilewise plume classification results with train/test data from {\em different} domains. Left column: \ANG training and \LES testing data. Right column: \LES training and ANG testing data. Top row: Training loss/epoch. Middle row: Optimal test \FB score/epoch. Bottom row: Salience threshold maximizing test \FB/epoch.}
 \label{fig:ang_vs_les_inter}
\end{figure*}

This case study summarizes our results using \LES-generated plumes to train plume detectors for the \ANG, \CALMETHANE, Permian, and COVID campaigns. We used the same \LES generated plumes injected into ANG CMF background described in \cite{jongaramrungruang_methanet_2022}. For each campaign, we created a new dataset where the background tiles are unchanged, but the observed \ANG training and test plume tiles were replaced with synthetic WRF-LES tiles. We used the following process to select \LES tiles that roughly align with the distribution of plume emission rates/wind speeds observed in the above \ANG campaigns. First, we computed the (\nth{5},\nth{95}) percentile ranges of the observed flux rates and wind speeds associated with all plumes observed in the \ANG \CALMETHANE, Permian, and COVID campaigns, adding a $\pm$ 10\% buffer to each range to permit sampling some out-of-distribution \LES tiles. We used these \ANG derived bounds to select \LES plume tiles from the \cite{jongaramrungruang_methanet_2022} data set via stratified random sampling with emission rates in 10 equiprobable flux bins $\in [36.11, 1860.97]$ \kgh and wind speeds in 5 equiprobable bins $\in [1.08, 5.44]$ \mps. Finally, we partition the 2651 \LES plume tiles sampled using the aforementioned criteria into 75/25\% training/test sets. 

We generated a new \LES dataset for each campaign by replacing the \ANG plume tiles observed in that campaign with the \LES plume tiles described above.  We then trained single campaign \GNET and \METHANET CNN models using the \LES plume data sets, and measured the accuracy of their resulting predictions on the existing \ANG plume test sets versus the new test sets containing the \LES plumes. All experiments involving \ANG plumes are based on fixed training and test partitions defined for each campaign, while all experiments involving \LES plumes use the same set of training and test plume samples in each experiment, but use the same set of training/test background samples used in the \ANG plume experiments for each campaign. All models and optimization parameters were initialized with fixed parameters per architecture (i.e., \GNET weights and state parameters initialized with numerically identical values at the starting epoch in both \ANG and \LES experiments for each campaign).

Table~\ref{tab:capacity}
The ``intra-domain'' loss curves where each model was trained and tested on data from the same domain (i.e., \ANG train/test plumes, \LES train/test plumes, respectively) are shown in Figure~\ref{fig:ang_vs_les_intra}. The ``inter-domain'' loss curves where each model was tested on data from one domain and tested on data from the other domain (i.e., \ANG trained model applied to \LES plumes) are shown in Figure~\ref{fig:ang_vs_les_inter}. Models trained using \ANG plumes perform well on both \ANG and \LES plumes. On the other hand, despite the significant increase in the number of plume samples that occurs when we replace the \ANG plumes / campaign with the multicampaign set of \LES plumes, models trained with \LES plumes only perform well when applied to other \LES plumes. The rapid convergence combined with the high test accuracy with \LES train/test data indicates \LES plumes are well-separated from \ANG background. However, the fact that the \ANG test loss diverges with the \LES trained model suggests that the \LES plumes do not fully capture the diversity of the observed \ANG plumes to permit robust detection.

\begin{table*}
 \centering \includegraphics[width=1.0\linewidth,clip,keepaspectratio]{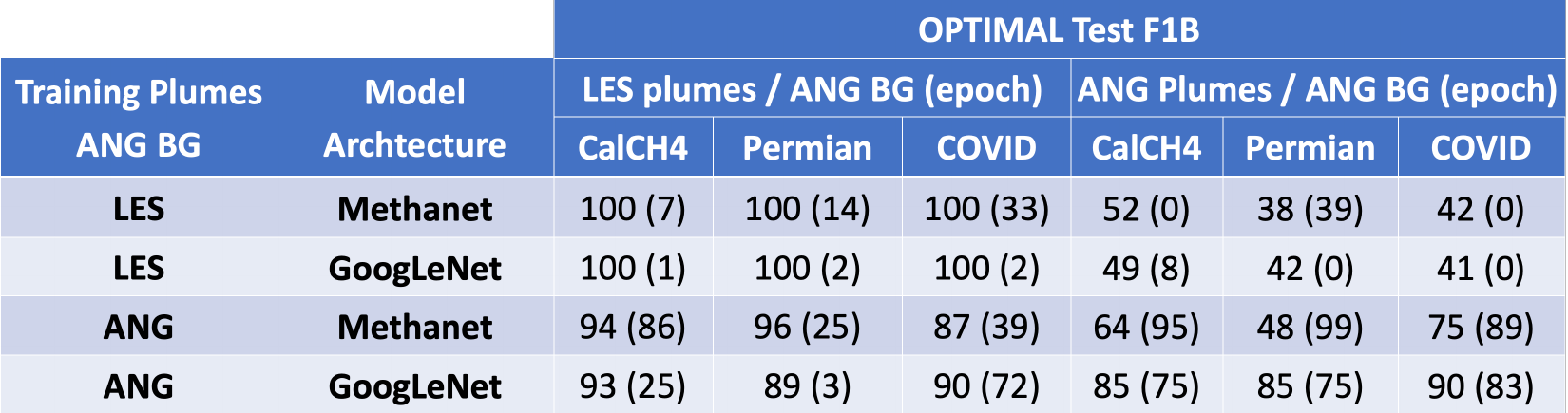} 
\caption{Summary of \METHANET \& \GNET tilewise plume classification results using \ANG versus \LES training \& test tiles.}
 \label{tab:capacity}
\end{table*}

%using only plumes generated by \LES , 
%Augmenting training data consisting of manually identified plumes with \LES-injected plumes can potentially improve generalization performance (see e.g., \cite{rao_lesch4_2022}), while 

%\cite{theiler_transductive_2014}, 

\subsection{Impact of Internal Downsampling: \GNET versus MethaNet} \label{apx:model_capacity}

\begin{figure*}
 \centering \includegraphics[width=1.0\linewidth,clip,keepaspectratio]{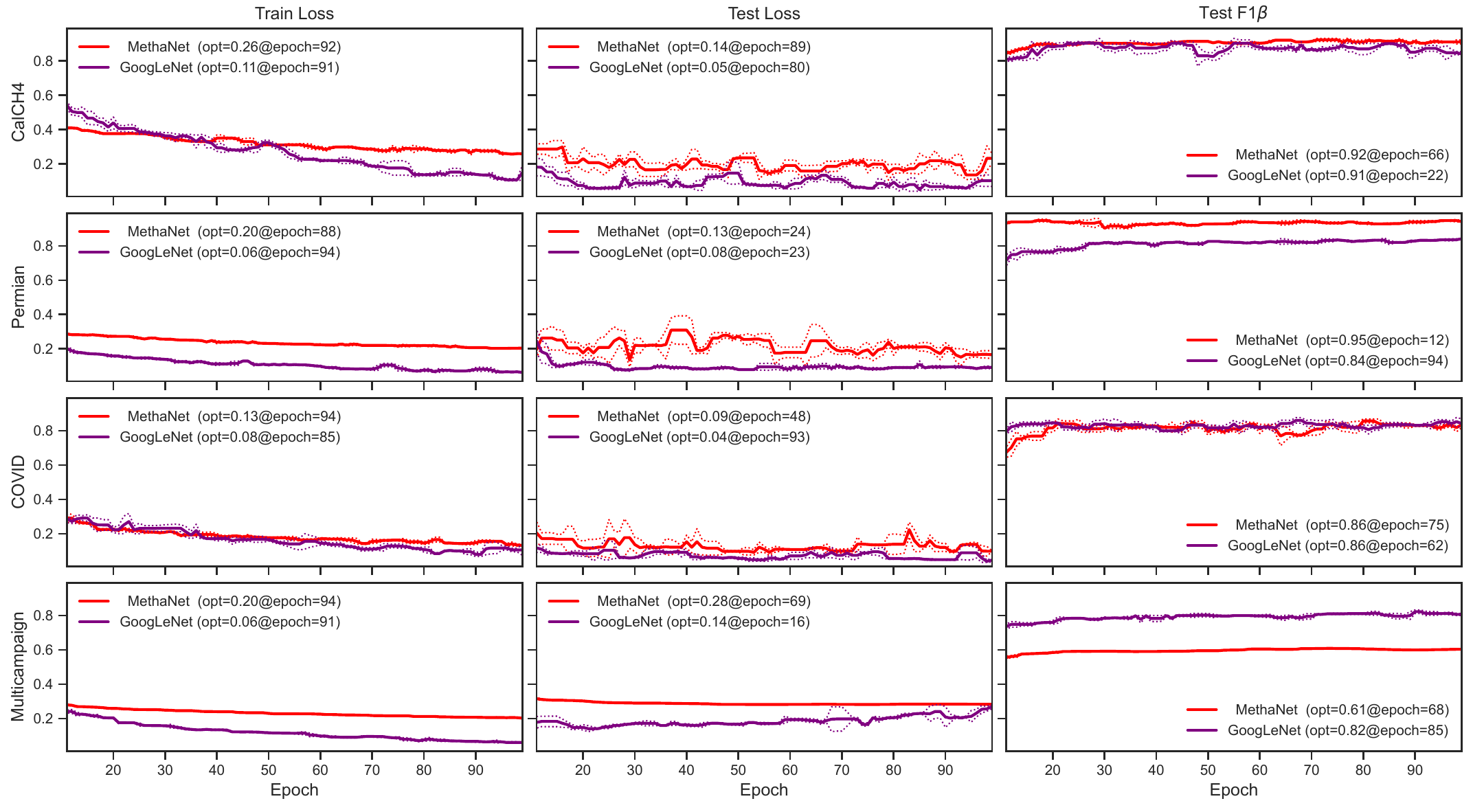} 
\caption{Comparing the capacity of \METHANET (red) vs. \GNET (purple) models for multicampaign plume detection. Left column: Training loss curves. Middle: Test loss curves. Right: test \FB per-epoch. Top three rows: Single Campaign \CALMETHANE vs. Permian vs. COVID results. Bottom row: MultiCampaign results.}
 \label{fig:capacity}
\end{figure*}
ML models must have sufficient representational capacity. I.E., they require a sufficient quantity of parameters to learn discriminative patterns with respect to each class and retain those patterns internally. A common means to increase the capacity of a CNN-based model is to increase the depth of the network by stacking convolution \& pooling layers. Doing so enables these ``deep'' models to learn spatial features at varying scales and permits some degree of translation invariance in the input data. However, the repeated pooling operations applied in deep networks may reduce their ability to capture small-scale features that may be learnable with comparable shallow networks with fewer layers.   

While \GNET performs well in both single-campaign and multicampaign settings, \METHANET only performs well in the single-campaign case, and plateaus quickly in the multicampaign setting. Notably, both train and test loss flatten while accuracy plateaus for \METHANET. This indicates the model has not overfit to the training set, but rather, the model \textit{cannot} overfit to the training set since it is underparametrized. However, \METHANET significantly outperforms \GNET on the Permian campaign data, achieving test \FB near 0.95 vs. GoogleNet's 0.84 optimum.

This result is likely due to two factors: First, Sources in Permian Basin are all Oil \& NG sources, with surface infrastructure types typified by small, high concentration plumes. The CMF images captured in the \ANG Permian campaign are also distinguished by their relatively coarse spatial resolution (GSD ~7.5\msq) versus the \CALMETHANE and COVID airborne campaigns used for training the model (GSD ~3.1\msq).Overall \GNET internal downsampling factor (DSF) is a factor of $4 \times$ larger than \METHANET. The \GNET architecture is several layers with 13.4M parameters, uses 3 pooling layers each with a downsampling factor of two, with a total internal downsampling factor of $2^3 = 8$. \GNET generalizes well for more diverse plumes (varying morphologies,  concentration), at the cost of sacrificing performance on very small plumes. In contrast, the \METHANET architecture consists of only 4 convolutional layers and 1.2M parameters, with a total internal downsampling factor of 2. This allows this architecture avoid losing very small plumes during downsampling, resulting in the improved performance.

\bibliographystyle{elsarticle-num-names-nourl} 
\bibliography{opsghgml}